\documentclass[a4paper,fleqn]{cas-sc}

\usepackage[numbers]{natbib}
\usepackage{hyperref}
\usepackage{array}
\usepackage{float}
\usepackage{makecell}

\newcommand{\new}[1]{{\color{red}#1}}
\renewcommand{\new}[1]{{#1}}
\newcommand{\blue}[1]{{\color{blue}#1}}
\renewcommand{\blue}[1]{{#1}}

\begin{document}

\shorttitle{}

\shortauthors{J. P. Vieira et~al.}

\title [mode=title]{Towards a more realistic evaluation of machine learning models for bearing fault diagnosis}

\author[inst1]{João Paulo Vieira}[orcid=0009-0002-0971-1610]
\credit{Methodology, Software, Formal analysis, Investigation,  Visualization, Writing - original draft}
\author[inst2]{Victor Afonso Bauler}[orcid=0009-0001-4754-9610]
\credit{Conceptualization, Methodology, Data Curation, Software, Writing - original draft}
\author[inst1]{Rodrigo Kobashikawa Rosa}[orcid=0009-0008-9325-3600]
\credit{Conceptualization, Data curation, Visualization, Software, Writing - original draft}
\author[inst1]{Danilo Silva}[orcid=0000-0001-6290-7968]
\cormark[1]
\credit{Conceptualization, Methodology, Project administration, Resources, Supervision, Writing - review \& editing}
\affiliation[inst1]{organization={Department of Electrical and Electronic Engineering, Federal University of Santa Catarina},
            city={Florianópolis},
            country={Brazil}}
\affiliation[inst2]{organization={Department of Mechanical Engineering, Federal University of Santa Catarina},
            city={Florianópolis},
            country={Brazil}}
\cortext[cor1]{Corresponding author}

\begin{abstract}
Reliable detection of bearing faults is essential for maintaining the safety and operational efficiency of rotating machinery. While recent advances in machine learning (ML), particularly deep learning, have shown strong performance in controlled settings, many studies fail to generalize to real-world applications due to methodological flaws, most notably data leakage. This paper investigates the issue of data leakage in vibration-based bearing fault diagnosis and its impact on model evaluation. We demonstrate that common dataset partitioning strategies, such as segment-wise and condition-wise splits, introduce spurious correlations that inflate performance metrics. To address this, we propose a rigorous, leakage-free evaluation methodology centered on bearing-wise data partitioning, ensuring no overlap between the physical components used for training and testing. Additionally, we reformulate the classification task as a multi-label problem, enabling the detection of co-occurring fault types and the use of prevalence-independent metrics \new{based on the ROC curve}. Beyond preventing leakage, we also examine the effect of dataset diversity on generalization, showing that the number of unique training bearings is a decisive factor for achieving robust performance. We evaluate our methodology on \new{four} widely adopted datasets: \blue{Case Western Reserve University} (CWRU), Paderborn University (PU), University of Ottawa (UORED-VAFCLS) and \blue{Hanoi University of Science and Technology (HUST bearing)}. This study highlights the importance of leakage-aware evaluation protocols and provides practical guidelines for dataset partitioning, model selection, and validation, fostering the development of more trustworthy ML systems for industrial fault diagnosis applications.
\end{abstract}



\maketitle

\section{Introduction}
\label{sec:introduction}

The field of fault diagnosis for rotating machinery, particularly rolling bearings, has seen increased attention due to its critical role in various industries and the demand for efficient operations \cite{lei2020applications}. Early and accurate detection of bearing failures can significantly reduce unexpected machine downtime and improve maintenance schedules, avoiding financial losses and safety risks. Machine Learning (ML) approaches, including deep learning architectures, coupled with wireless sensor technologies, have enabled health monitoring and failure prediction at scale \new{\cite{hoang_survey_2019, barai_bearing_2022, soomro_insights_2024}}.

Despite the significant advancements offered by ML, its application requires a careful methodology to ensure models generalize reliably to real-world scenarios. A critical methodological pitfall in ML-based science is data leakage \cite{Kapoor2024}. Data leakage is defined as a spurious relationship between independent variables and the target variable, arising from flaws in data collection, sampling, or preprocessing \cite{kapoorleakage2023}. Such an artifact, not present in the true data distribution, typically leads to overoptimistic estimates of model performance. This phenomenon is a major source of error in ML applications, often causing published models to fail when deployed in practical settings, impacting at least 294 papers across 17 scientific fields \cite{kapoorleakage2023}. For instance, in medicine, improper handling of patient data can lead to leakage if samples from the same patient are used in both training and test sets. Cases have been reported where models included features that were effectively proxies for the outcome, such as the use of anti-hypertensive drugs to predict hypertension or antibiotics to predict sepsis, leading to artificially inflated performance.

Our observations indicate that data leakage remains a prevalent issue in the field of bearing fault diagnosis. Numerous studies fail to partition datasets correctly, resulting in information leakage and, consequently, over-optimistic performance estimates that do not hold in real-world scenarios. For instance, studies that assign waveform recordings from the same bearing to both training and test partitions have consistently reported inflated performance. Early work by \cite{Rauber_2021, varejao_similarity_2025} highlighted this ``similarity bias'' in machine learning research utilizing vibration data, revealing that nearly all reviewed studies (published between 2008 and 2020), including 40 out of 41 using the widely adopted Case Western Reserve University (CWRU) dataset, employed experimental designs susceptible to this bias. These findings are corroborated by \cite{wheat_impact_2024}, who examined 55 papers published between 2020 and 2024 in the bearing fault diagnosis domain, finding that only six employed rigorous data-splitting methodologies. In addition to these previous studies, the present work contributes an investigation of 18 papers published in 2025, identifying that data leakage persists in the majority of works in the field, leading to overestimated results (Section \ref{sec:prevalence}).

To achieve a more realistic evaluation, \new{the train-test split must mirror the intended deployment scenario. Considering a deployment scenario in which bearings may be replaced during maintenance,} this paper advocates for bearing-wise splitting, which consists of ensuring that all data originating from the same bearing is assigned exclusively to a single (train or test) partition. This strategy is necessary to prevent data leakage, since otherwise a model may learn bearing-specific artifacts that are spuriously correlated with the target variable, leading to overly optimistic results. In other words, the model may simply memorize the bearing identity and its associated fault label, rather than learning robust fault signatures that generalize to unseen bearings.

Establishing that conventional (non-bearing-wise) splitting strategies indeed lead to unrealistic results---and thus should be avoided in future work---requires one to show that using a bearing-wise split causes a significant performance gap in an otherwise identical setup. This path was taken by \cite{wheat_impact_2024,hendriks2022towards,matania-leakage,abburi2023closer}, which have observed accuracy dropping from near 100\% to around 40\%-60\% depending on the specific setup. However, \cite{hendriks2022towards,matania-leakage} proposed splitting data by fault size, which does not necessarily corresponds to a bearing-wise split (see Section~\ref{sec:relatedworks}), while \cite{hendriks2022towards,abburi2023closer} could not entirely eliminate leakage with respect to the healthy class in the CWRU dataset, as this dataset contains a single healthy bearing configuration.

Additionally, none of these works controlled an important confounding factor in their experimental design: the number of training bearings. Specifically, when naively changing from a traditional split (where all available bearings are used for training) to a bearing-wise split (where the bearings used for testing are not included in the training set), the number of bearings seen during training is reduced. As is well-known in machine learning literature and corroborated by our experiments with bearing data, the diversity of training data (not just the raw number of samples) is an important driver of model performance. Thus, it is conceivable that the aforementioned performance gap could arise simply due to a reduced training diversity. To convincingly show that this is not the case, in this paper we perform controlled experiments where the exact trained model is kept fixed and only the test dataset is changed based on the splitting strategy. To the best of our knowledge, this is the first paper to present such experiments, through which we hope to convince readers that using an appropriate splitting procedure is strictly necessary to produce valid results.

An alternative approach to completely eliminating data leakage is conducting inter-testbench experiments, typically framed within a cross-domain or domain generalization context. In this formulation, the testbenches used for training and testing are completely isolated; however, this introduces the significant challenge of domain generalization. We suspect that the bearing fault diagnosis literature has largely treated intra-domain classification as a solved problem, prompting a shift toward cross-domain research. Research by \cite{chen_deep_2023} indicates that many studies define a domain as a combination of operating conditions (e.g., load, rotation speed, torque) rather than distinct physical bearings. This perspective has led to various data splitting methodologies within single datasets, where data is partitioned according to these conditions. A limited number of studies have addressed the inter-testbench scenario by isolating datasets, such as \cite{matania_zero-fault-shot_2025}, and \cite{zheng_deep_2021}. Although these approaches effectively eliminate data leakage, domain generalization remains a complex challenge, as target datasets often exhibit disparate feature distributions, necessitating specialized techniques to extract domain-invariant features. The present work, however, focuses on a simpler yet fundamentally critical problem: training and testing within a single testbench.

This paper aims to further advance the reliable development and deployment of ML models for bearing fault diagnosis. We propose a novel methodology that rigorously addresses data leakage and class imbalance in an intra-dataset scenario, particularly for datasets with limited healthy bearing data. Our approach formulates the problem as a binary multi-label classification for each sensor location (e.g., drive end and fan end), enabling the detection of the presence or absence of each fault type (e.g., inner, outer, ball). This formulation specifically addresses the disadvantages of multiclass accuracy, which serves as a poor proxy for real-world performance. Because accuracy treats all misclassifications as equal, it fails to distinguish between ``false alarm'' (False Positive) and a ``missed detection'' (False Negative), the latter being significantly more costly in industrial maintenance. Furthermore, in the presence of class imbalance, a model can achieve high accuracy by simply predicting the majority class, effectively ``hiding'' its inability to identify rare but critical fault states. Another issue caused by the multiclass formulation is the inability of detecting co-occurring faults. While it is possible to create classes for combined faults, it is often unpractical and most public datasets contain few examples of those cases. By treating faults as independent binary problems, we can accommodate co-occurring defects and utilize faulty signals from one label as true negatives for others---a significant advantage when healthy data is scarce, such as in the CWRU dataset. In this formulation, we can also utilize prevalence-independent metrics such as the Area Under the Receiver Operating Characteristic curve (AUROC). This allows for a more realistic representation of real-world conditions where decision thresholds must be precisely tuned to prioritize sensitivity or specificity based on specific safety and operational requirements.

In summary, our contributions include:
\begin{itemize}
    \item Designing experiments that isolate data leakage from other confounding factors using both synthetic and real data.
    By fixing the trained model and comparing a test set containing signals from bearings seen during training (leaked) against another test set composed entirely of unseen bearings, we show that performance degradation is directly attributable to data leakage rather than  reduced training data diversity.
    \item Providing a systematic methodology for creating and using vibration-based datasets in ML experiments that strives to prevent data leakage and suggesting an accompanying hyperparameter tuning process that adheres to the same bias minimization principle. 
    \item Proposing a multilabel problem formulation, which enables a more precise evaluation by using prevalence-independent metrics such as the ROC curve and the AUROC and provides  a more realistic representation of real-world conditions where multiple fault types could coexist.
    \item Applying the proposed methodology on widely-used vibration datasets, such as CWRU, Paderborn University (PU), University of Ottawa (UORED-VAFCLS) \new{and HUST bearing} datasets. Our results reveal the significant impact of bearing diversity on model generalization and demonstrate that the optimal choice between deep and shallow learning models is highly dataset-dependent.

\end{itemize}

With these contributions, this work aims to foster the development of more robust and trustworthy ML models for bearing fault diagnosis, ensuring their performance more closely reflects their capabilities in real-world industrial settings. The source code for this paper can be found at \href{https://github.com/gama-ufsc/bearing-data-leakage}{github.com/gama-ufsc/bearing-data-leakage}.

\section{Background}
\label{sec:background}

\subsection{Basic concepts on supervised learning}

Supervised machine learning is a paradigm centered on learning a mapping from inputs to outputs based on a set of labeled examples. The fundamental goal is to approximate an unknown underlying function that dictates the relationship between the observed data and their corresponding labels.

Mathematically, we consider an input space $\mathcal{X}$ and an output space $\mathcal{Y}$. The relationship between them is governed by a true, but unknown, joint probability distribution $P(X, Y)$, where $X \in \mathcal{X}$ and $Y \in \mathcal{Y}$. We are not given access to $P(X, Y)$ directly. Instead, we are provided with a finite set of observations, known as the \textbf{training set}, $D_{train} = \{(x_i, y_i)\}_{i=1}^{N}$, where each \textbf{sample} $(x_i, y_i)$ is assumed to be an independent and identically distributed (i.i.d.) draw from $P(X, Y)$.

Each input $x_i$ is typically a \textbf{feature vector}, $x_i \in \mathbb{R}^d$, representing a set of $d$ measurable properties of the phenomenon being observed. The output $y_i$ is the corresponding label or target value. The task is to select a \textbf{model} from a hypothesis space $\mathcal{H}$, which is a family of functions $f: \mathcal{X} \to \mathcal{Y}$. A specific model is defined by a set of parameters, $\theta$, denoted as $f_{\theta}$.

The learning process consists of finding the optimal parameters $\theta^*$ that enable the model to make accurate predictions. To achieve this, we first define a \textbf{loss function}, $\mathcal{L}(f_{\theta}(x), y)$, which quantifies the penalty or error for predicting $f_{\theta}(x)$ when the true label is $y$. The ultimate objective is to minimize the \textit{true risk} or \textit{expected loss} over the entire data distribution
\begin{equation}
R(f_{\theta}) = \mathbb{E}_{(x,y) \sim P(X,Y)}[\mathcal{L}(f_{\theta}(x), y)].
\end{equation}
Since $P(X, Y)$ is unknown, the true risk cannot be calculated directly. Therefore, we approximate it using the \textit{empirical risk} on the training set
\begin{equation}
R_{emp}(f_{\theta}) = \frac{1}{N} \sum_{i=1}^{N} \mathcal{L}(f_{\theta}(x_i), y_i).
\end{equation}
The training process then becomes an optimization problem focused on finding the parameters $\theta^*$ that \textbf{minimize this empirical risk}
\begin{equation}
\theta^* = \arg\min_{\theta} R_{emp}(f_{\theta}).
\end{equation}

The model's parameters, $\theta^*$, are optimized exclusively on the training set, $D_{train}$. Since the model is ultimately a function of the training set, the model's performance on that same data is not a reliable indicator of its actual predictive power; it is an inherently biased and optimistic measure. What truly matters is the model's \textbf{generalization performance}---how well it performs on new, unseen data from the same underlying distribution, $P(X, Y)$. To perform this evaluation, we use a disjoint \textbf{test set}, $D_{test}$, which is a collection of i.i.d.\ samples from $P(X, Y)$ kept completely separate during the entire training and model selection process.

However, in addition to the learnable parameters $\theta$, most models are also characterized by a set of \textbf{hyperparameters}, $\lambda$, which are not optimized during training but rather define the model's architecture or the learning algorithm's behavior (e.g., learning rate, regularization strength). The process of selecting the optimal configuration of these hyperparameters is known as \textbf{model selection}. Using the test set, $D_{test}$, to guide this selection process is methodologically unsound, as it would mean that information from the test set has leaked into the model configuration, violating the i.i.d.\ assumption. Consequently, $D_{test}$ would no longer provide an unbiased estimate of the final model's generalization performance.

To address this, the dataset $D$ is typically partitioned into three disjoint subsets: a \textbf{training set} ($D_{train}$), a \textbf{validation set} ($D_{val}$), and a \textbf{test set} ($D_{test}$). The hyperparameter optimization process proceeds as follows: for each candidate hyperparameter configuration $\lambda$, a model is trained on $D_{train}$ to find the optimal parameters $\theta^*(\lambda)$. The performance of this trained model is then evaluated on the validation set, yielding a \textbf{validation risk}. The hyperparameter configuration $\lambda^*$ that results in the lowest validation risk is selected as the optimal one:
\begin{equation}
\lambda^* = \arg\min_{\lambda} \left( \frac{1}{|D_{val}|} \sum_{(x,y) \in D_{val}} \mathcal{L}(f_{\theta^*(\lambda)}(x), y) \right).
\end{equation}

Once the optimal hyperparameters $\lambda^*$ have been identified, the final model is trained and its generalization performance is reported based on a single, final evaluation on the held-out test set, $D_{test}$. The performance on this held-out data gives us an unbiased estimate of the true risk, $R(f_{\theta^*(\lambda^*)})$. In essence, it is our best proxy for how the model will behave in the wild. This practice is crucial for diagnosing \textbf{overfitting}, a common pitfall where a model appears highly accurate on the data it was trained on but fails to generalize to new examples.

While the assumption that data samples are independent and identically distributed is fundamental to supervised learning, it is frequently violated in real-world machine learning applications, potentially leading to overly optimistic performance estimates if not properly addressed \cite{Roberts_2017}. A common scenario where this assumption breaks down is when the underlying data generation process naturally produces groups of dependent samples, such as multiple medical records from the same patient \cite{kapoorleakage2023}. Addressing these violations is critical for achieving robust and reliable model performance, preventing these overly optimistic estimates. The overarching strategy involves group cross-validation (CV) \cite{sklearn}\footnote{An implementation of cross-validation applied to grouped data is available on \url{https://scikit-learn.org/stable/modules/cross_validation.html\#cross-validation-iterators-for-grouped-data}.}, where data is split in a manner that respects these inherent correlation structures rather than through purely random partitioning, ensuring that training and evaluation data remain genuinely independent.

\subsection{Evaluation metrics}
\label{sec:evaluation-metrics}

\newcommand{\indicator}{\mathbf{1}}

Consider now a classification model $f: \mathcal{X} \to \mathcal{Y}$, where $\mathcal{Y} = \{1,\ldots,K\}$ is a finite set and $K$ is the number of classes. Let $X \in \mathcal{X}$ and $Y \in \mathcal{Y}$ be random variables whose joint distribution is given by $P(X,Y)$. The primary objective of a classification model is to correctly map inputs to their respective categories, an ability that can be evaluated using quantitative performance metrics. The most widely used metric, \textbf{accuracy}, measures the proportion of correctly classified instances and is defined as
\begin{equation}
\text{Acc}(f) = P[f(X) = Y].
\end{equation}
However, this metric is fundamentally limited by its sensitivity to class prevalence \cite{hogan2023on}. In problems with significant class imbalance, accuracy becomes misleading, as a model can achieve a deceptively high score by simply predicting the majority class. To address this, metrics more robust to class imbalance are required, such as the \textbf{balanced accuracy}, defined as the arithmetic mean of the per-class recall:
\begin{equation}
\text{Bacc}(f) = \frac{1}{K} \sum_{k=1}^K P[f(X) = k | Y = k].
\end{equation}

For binary classification, where $\mathcal{Y} = \{0,1\}$ and $K=2$, the balanced accuracy can be expressed as
\begin{equation}
\text{Bacc}(f) = \frac{\text{TNR} + \text{TPR}}{2} = 1 - \frac{\text{FPR} + \text{FNR}}{2}
\end{equation}
where
\begin{align}
\text{TNR} &= P[f(X) = 0 | Y = 0] \\
\text{TPR} &= P[f(X) = 1 | Y = 1] \\
\text{FPR} &= P[f(X) = 1 | Y = 0] = 1 - \text{TNR} \\
\text{FNR} &= P[f(X) = 0 | Y = 1] = 1 - \text{TPR}
\end{align}
which stand for True Negative Rate, True Positive Rate, False Positive Rate and False Negative Rate, respectively. Note that, in this case, accuracy can be equivalently expressed as
\begin{equation}
\text{Acc}(f) = (1-p) \cdot \text{TNR} + p \cdot\text{TPR} = 1 - (1-p) \cdot \text{FPR} - p \cdot\text{FNR}, \qquad p = P[Y=1]
\end{equation}
highlighting its dependence on the prevalence $p$. For instance, a classifier $f(x) = 0$ that always predicts the negative class achieves accuracy $1 - p$ (while $\text{FPR} = 0$ and $\text{FNR} = 1$). Ideally, we would like both error metrics FPR and FNR to be close to zero.

Often, in binary classification, we have access to a family of classifiers $f_\tau$ parameterized by a threshold $\tau$, specifically, $f_\tau(x) = \indicator[s(x) \geq \tau]$, where $s: \mathcal{X} \to \mathbb{R}$ is a scoring function and $\indicator(\cdot)$ denotes the indicator function. In this case, we are not limited to a specific pair (FPR, FNR) of error metrics; instead, by sweeping over $\tau$, we can arbitrarily choose a desired operating point. The tradeoff between the achievable (FPR, FNR)---or, more commonly, between (FPR, TPR)---for all possible $\tau$ is known as the \textbf{Receiver Operating Characteristic} (ROC) curve \cite{fawcett2006}.
 
As the ROC curve represents classifier behavior across a continuum of decision thresholds, it is often convenient to summarize this information into a single scalar value. The Area Under the ROC Curve (AUROC) provides such a summary by computing the integral of the TPR $\times$ FPR curve over the entire $[0,1]$ range. In particular, the AUROC of a perfect classifier equals 1, while that of a random classifier equals 0.5. Because it is computed over all operating points, the AUROC does not depend on the selection of a specific decision threshold. Furthermore, as it is derived from conditional rates rather than absolute class frequencies, the AUROC is invariant to class prevalence.

It is straightforward to extend these definitions to multi-label classification.
In binary multi-label classification with $L$ labels, one deals with $L$ independent binary classifiers. Each classifier yields its own score function and corresponding ROC curve, resulting in a set of label-specific AUROC values. To obtain a single representative measure of performance across all $L$ labels during model development, these AUROC values can be aggregated via macro-averaging. The resulting Macro AUROC is defined as the arithmetic mean of the individual AUROC scores:
\begin{equation}
\text{Macro AUROC} = \frac{1}{L} \cdot \sum _{i=1}^{L} \text{AUROC}_i .
\end{equation}
This aggregation assigns equal weight to each label and provides a concise summary of overall model behavior, while the individual ROC curves retain their relevance for label-specific analysis and threshold selection.

\section{Data Leakage}
\label{sec:dataleakage}

\subsection{Data splitting and Data Leakage}


Data leakage is a major source of error in machine learning applications \cite{datamining}, and often the reason why published models fail to generalize to real-world data \cite{pitfalls}. As defined by \cite{kapoorleakage2023}, it refers to spurious correlations between input and target variables arising from flaws in data collection, sampling, or preprocessing. These artificial relationships, absent in the true data distribution, typically yield overly optimistic performance estimates during development but poor generalization to unseen data. 

As mentioned in Section \ref{sec:background}, there are many methodologic flaws that violate the i.i.d.\ assumption, which ultimately results in data leakage. This may happen through model or feature selection before data partitioning, during preprocessing, improper handling of test data and splitting methods. In sequence, examples for the mentioned cases will be discussed. 

A common flaw that introduces data leakage during preprocessing is computing statistics such as means or ranges for scaling, or imputing missing values, using the entire dataset instead of restricting calculations to the training portion. Similarly, performing model or feature selection before data partitioning allows information from the test set to influence model configuration \cite{pitfalls}. To prevent these issues, all preprocessing and feature engineering steps must be fitted exclusively on training data.

Another source of data leakage comes from improper handling of test data. Using the same test set to evaluate multiple models can inadvertently inform model selection, leading to overfitting. Additionally, applying data augmentation before dividing the dataset can cause augmented information from the test set to seep into the training data, compromising the model's ability to generalize. 

In time-series applications, random partitioning without preserving temporal order can allow future information to leak into the training process, producing inflated performance metrics. Even in non-temporal datasets, experimental designs may introduce dependencies or duplicate samples that, if split incorrectly, create information overlap between train and test sets. 

Key to avoiding these pitfalls is strict maintenance of train-test separation throughout the entire pipeline. This means ensuring that no information from the test set influences preprocessing, feature selection, hyperparameter tuning, or model training \cite{Lones_2024}. In practice, this involves grouping dependent samples (e.g., patient-level records or bearing-level measurements) within the same fold, and using specialized cross-validation schemes for structured data---such as blocked CV for time-series \cite{Roberts_2017}---to prevent “look-ahead bias” and maintain a valid evaluation protocol.

\subsubsection{Data Leakage in Bearing Fault Diagnosis}

\label{sec:bearing-leakage}

In the evaluation of bearing fault diagnosis models, data leakage represents a significant pitfall that can lead to overly optimistic performance estimates. We identify and categorize two common but flawed intra-bearing splitting protocols that lead to various forms of leakage: segmentation-level and bearing-level leakage.

\textbf{Segmentation-Level Leakage} occurs when non-overlapping segments from the same time-series signal (e.g., from a continuous experimental run) are split between training and test sets. Although the segments do not overlap in time, the underlying signal is temporally and physically coherent, allowing the model to learn time-specific or signal-specific artifacts rather than generalizable fault features. In bearing fault diagnosis, this form of leakage is especially common and is discussed in detail in \cite{varejao_similarity_2025}.

\textbf{Bearing-Level Leakage} arises when data from the same physical bearing is distributed across both training and test sets. This may occur through common random splitting procedures or, more specifically, through approaches such as \textit{condition-wise} and \textit{repetition-wise} splits, defined as follows.

A \textit{condition-wise} split happens when signals with different machine conditions are divided between training and testing. A condition is usually represented by the combination of machine configurations, such as the load and rotation speed, for instance. Alternatively, a condition may be defined in terms of the fault severity level of a given bearing, as in the UORED-VAFCLS dataset, where two distinct severity levels are provided for each bearing. This type of data splitting is commonly used in bearing fault diagnosis, as it is generally understood that a realistic evaluation methodology requires diverse machine configurations to be separated between training and testing. While some may argue that such variation justifies this split, it is conceivable that some intrinsic signature of the bearing remains present in both sets, allowing the model to exploit identity-specific features rather than learning robust patterns.

An even more severe form of data leakage arises from \textit{repetition-wise} splits (also referred to as run-to-run in \cite{wheat_impact_2024}), where signals acquired under the same configuration are divided between training and testing. Since signals captured under identical setups typically share similar characteristics, this setup encourages models to memorize signal-specific patterns rather than learn features that generalize to fault diagnosis. In Section \ref{sec:dataleakage-pu}, we show that in the PU dataset, all splits with bearing-level leakage produce overoptimistic results, with repetition- and segment-wise splits reaching almost 100\% accuracy.

All these cases result in evaluations that are fundamentally closer to measuring memorization (e.g. the training performance) rather than generalization. To preclude such pitfalls, we advocate for a strict \textit{bearing-wise split}, in which the physical bearings used for training and testing are mutually exclusive. This approach ensures the model is assessed on its ability to generalize to unseen components, preventing it from leveraging identity-specific artifacts\footnote{If samples from the same bearing but assigned different labels (for example, when a healthy bearing is subjected to accelerated lifetime testing and later develops a fault) are split between training and test sets, this configuration does not characterize bearing-level leakage, because the resulting correlations do not inherently bias the model toward the correct class. Rather, it constitutes a more challenging evaluation scenario, as it assesses whether the model has learned meaningful feature representations rather than relying on bearing-specific signatures.}. Ultimately, this promotes the development of models that identify universal fault characteristics, a goal best supported by datasets that include multiple physical bearings per fault category. It is important to note that even with bearing-wise splits, leakage may still occur due to spurious correlations learned from poor data collection practices. For example, if all fault signals in a dataset contain additional interference absent from healthy signals, the model might rely on that interference for classification, thus enabling memorization.

\subsection{A Toy Example}
\label{sec:toyexample}

To demonstrate the performance inflation caused by a non-bearing-wise data split, we constructed a synthetic binary fault detection experiment, in which the task is to classify each sample as either faulty or healthy. We simulate a feature space for a dataset of $B=48$ distinct bearings, which could be obtained, for example, as the output of a deep learning model's feature extractor. This simulated space contains two feature types: 3 fault-predictive features and 48 bearing-identity features, resulting in 51 features. Each identity feature is unique to a specific bearing, representing a spurious correlation unrelated to the fault condition. All features were generated by adding zero-mean Gaussian noise with unit variance to a constant base value: $a_f = 1.5$ for fault-predictive features and $a_b = 8$ for bearing-identity features. All experiments were designed to maintain a fixed number of samples per bearing, using a base value of 40 samples each. The dataset comprised 24 healthy bearings and 24 faulty bearings, ensuring a consistently balanced class distribution.

First, we established a theoretical performance ceiling for the classifier, as shown in Appendix \ref{app:a}. In summary, computing the mean of the 3 fault predictive features and applying a threshold of $a_f/2$ gives the maximum achievable accuracy of $90.30\%$. We then evaluated models with two distinct test set configurations: the first consisted of samples generated from the same bearings used in the training set, while the second comprised samples generated from different bearings.

\begin{figure*}
    \centering
    \includegraphics[width=0.8\linewidth]{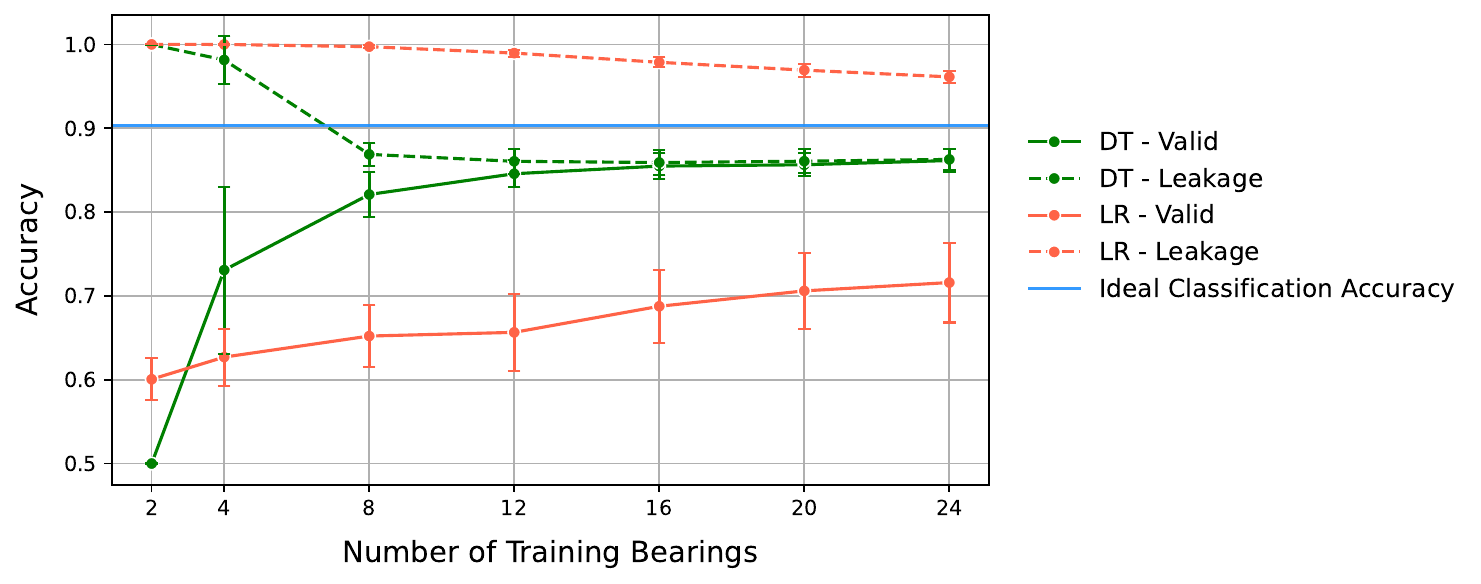}
    \caption{Comparison of \new{a low-capacity} Decision Tree (DT) and Logistic Regression (LR) accuracy across varying numbers of training bearings, evaluated under two conditions: a leakage-free test (Valid) set and a test set with data leakage (Leakage). \new{The higher-capacity LR appears to achieve higher performance under the data-leakage evaluation, even surpassing the theoretical optimal, which is an illusion caused by the fact that this evaluation is effectively a proxy for training performance.}
}
    \label{fig:toy-problem}
\end{figure*}

\new{Using a Logistic Regression classifier and a low-capacity Decision Tree classifier (with maximum depth 3)} to assess the impact of model capacity, our results in Figure~\ref{fig:toy-problem} show that test sets containing samples from bearings included in the training set yield overly optimistic performance metrics for both classifiers, even exceeding the theoretical maximum classification accuracy. \new{This behavior arises from the distinct inductive biases of the two models in this synthetic setting. Since each bearing is defined by its own identity feature, Logistic Regression can exploit its high-dimensional linear decision boundary to effectively memorize a large number of bearing-specific patterns. In contrast, the Decision Tree is constrained to splitting on one feature at a time, which limits its ability to memorize many bearings within a depth of 3. As the number of training bearings increases, the tree is therefore encouraged to rely on the weaker but genuinely fault-predictive features, which generalize across bearings.} The results obtained under data leakage misleadingly suggest that Logistic Regression is the superior model; however, under the valid test set, the Decision Tree actually demonstrates better performance when 4 or more training bearings are used. This highlights that results influenced by data leakage are unreliable for informed decision-making. Additionally, increasing the number of bearings in training improves performance on the leakage-free test set, while it also mitigates the impact of leakage on both models, suggesting it learns to prioritize the correct predictive features. These findings underscore the critical importance of a rigorous validation methodology to prevent misinterpretation of a model's true generalization capabilities.

\subsection{Prevalence of data leakage in bearing fault diagnosis}
\label{sec:prevalence}

As mentioned in the introduction, data leakage is widespread in the literature on bearing fault diagnosis, with previous works \cite{Rauber_2021, wheat_impact_2024} reporting data splitting issues in over 90\% of published papers. To complement and update these findings, we conducted an investigation of papers published in 2025 using a similar methodology. First, we searched the Mechanical Systems and Signal Processing journal database using the queries ``bearing fault diagnosis'' and ``machine learning'', which yielded 195 published papers from January 2025 up to December 2025. We then randomly selected 10 papers that focus on the problem of training and testing within a single dataset under a supervised machine learning context. Second, we extended our search to include papers from other journals within the same scope, compiling an additional 8 papers. Finally, we carefully analyzed the experimental methodology in each of these papers to understand how the data split was performed.

Table \ref{tab:papers-leakage} lists these 18 papers, detailing whether they specified the train-test partitioning and, if so, how it was performed. We found that 8 of the 18 papers used a random splitting strategy, while 9 detailed a condition-wise split. The most common condition considered for partitioning was the load, although other papers considered rotation speed or level of noise. The remaining paper did not mention any partitioning methodology, which suggests that the authors may not have devoted sufficient attention to a detail critical for preventing leakage. Our conclusion is that despite an increase in the number of articles that detail their train-test partitioning methodology, data leakage remains a prevalent issue.

\begin{table}
\caption{Sample of papers published in 2025 that propose or apply machine learning techniques on bearing fault diagnosis datasets. }

\centering
\resizebox{0.8\textwidth}{!}{%

\begin{tabular}{ccccc}
\textbf{Paper} &
  \textbf{Split type} &
  \textbf{Journal/Conference} &
  \textbf{Datasets Used} &
  \textbf{Results} \\
  \hline \cite{leakage1}
 & Random         & Other & Other                  & \textgreater 98.8\%  \\ \cite{leakage2}
 & Random         & MSSP  & Other                  & \textgreater 99.75\% \\ \cite{leakage3}
 & Random         & MSSP  & CWRU, MFPT + Others    & 100\%                \\ \cite{leakage4}
 & Condition-wise & MSSP  & Others                 & 100\%                \\ \cite{leakage5}
 & Condition-wise & MSSP  & UORED-VAFCLS, XJTU-SQV & 100\%                \\ \cite{leakage6}
 & Condition-wise & MSSP  & CWRU, MFPT, PU         & 100\%                \\ \cite{leakage7}
 & Condition-wise & MSSP  & UORED-VAFCLS, HIT      & 100\%                \\ \cite{leakage8}
 & Condition-wise & MSSP  & PU, JNU, HIT           & \textgreater 99.2\%  \\ \cite{leakage9}
 & Condition-wise & Other & CWRU                   & \textgreater 99.3\%  \\ \cite{leakage10}
 & Random         & Other & CWRU                   & \textgreater 99.9\%  \\ \cite{leakage11}
 & Random         & Other & CWRU, XJTU-SY          & \textgreater 99.5\%  \\ \cite{leakage12}
 & Condition-wise & Other & CWRU                   & \textgreater 98.5\%  \\ \cite{leakage13}
 & Not detailed  & MSSP  & CWRU + Other           & \textgreater 94\%    \\ \cite{leakage14} 
 & Random         & Other & Other                  & \textgreater 99.3\%  \\ \cite{leakage15}
 & Condition-wise & Other & CWRU + Other           & \textgreater 99.7\%  \\ \cite{leakage16}
 &
  \begin{tabular}[c]{@{}c@{}}Condition-wise \end{tabular} &
  MSSP &
  UORED-VAFCLS + Other &
  \textgreater 99.9\% \\ \cite{leakage17}
 & Random         & MSSP  & Other                  & \textgreater 98\%    \\ \cite{leakage18}
 & Random         & Other & CWRU, PU               & \textgreater 97\%   
\end{tabular}%
}
\label{tab:papers-leakage}
\end{table}


\subsection{Related works}
\label{sec:relatedworks}

Several studies in the literature have highlighted the issue of data leakage in bearing fault diagnosis, including \cite{varejao_similarity_2025}, \cite{wheat_impact_2024}, \cite{hendriks2022towards}, \cite{matania-leakage} and \cite{abburi2023closer}. Three of these works propose new data-splitting strategies that primarily avoid segmentation-level leakage, but still allow bearing-wise leakage to persist, with the exception of \cite{wheat_impact_2024} and \cite{matania-leakage}.

Hendriks et al. \cite{hendriks2022towards} proposed a fault-size splitting strategy for the CWRU dataset, considering it a better approach to obtain domain shift in bearing fault diagnosis. 
Since in the CWRU dataset each combination of fault size and type (i.e. inner, outer, ball) corresponds to a different bearing, splitting by fault size indirectly produces as a bearing-wise partition.
However, signals from the single healthy configuration (consisting of a healthy bearing in the fan end and another healthy bearing in the drive end) were included in both training and test sets, \new{as their separation was based on load conditions. For example, healthy signals acquired under 1 HP were assigned to the training set, while the remaining load conditions (2 HP and 3 HP) were reserved for testing. This amounts to a condition-wise split of the healthy signals, resulting in data leakage.}


Abburi et al. \cite{abburi2023closer} also proposed an alternative split strategy, now directly aimed at reducing bearing-level information leakage on the CWRU dataset. The authors employed traditional machine learning models in a multiclass classification setting with three fault types (inner race, outer race, ball) and the healthy condition. Their results showed that the bearing-wise split consistently led to worse performance across all metrics (accuracy, precision, recall, and macro F1-score) when compared to a random split that included bearing information leakage. However, they have also split the signals from the healthy configuration \new{similarly to \cite{hendriks2022towards}}, leading to inflated model performance, notably on the binary fault detection (fault versus no fault) problem.

Matania et al. \cite{matania-leakage} highlighted common types of data leakage that can occur when using bearing fault diagnosis datasets, proposing a fault-size guided splitting strategy similar to \cite{hendriks2022towards}, which they applied to the CWRU and PU datasets. Since this work primarily focused on diagnosis after a fault had been detected, it excluded healthy samples, thereby eliminating data leakage when utilizing CWRU. The paper claims that the correct way to avoid data leakage is to separate the data based on fault sizes. While this approach is suitable for CWRU and PU, where each bearing has a single fault size, it would not be applicable to datasets where each bearing is considered under multiple fault sizes (corresponding to the evolution of a fault over time), such as the UORED-VAFCLS. This limitation is clearly demonstrated in Section \ref{sec:ott-leakage}, where performance gains were observed in the bearing-level leakage experiment.

A comprehensive investigation into dataset biases and evaluation protocols was conducted by \cite{varejao_similarity_2025}, who proposed evaluation methodologies for widely used datasets such as CWRU, PU, MFPT, IMS, and UOC. 
Their approach focused on mitigating segmentation-level leakage, advocating for condition-wise splits across all datasets under a multiclass framework, with F1-macro reported as an auxiliary metric. However, the results remained overoptimistic due to residual bearing-level leakage, which was not addressed in their work.

Another notable contribution is the study by \cite{wheat_impact_2024}, which proposed a bearing-wise split under a multiclass classification setting, referred to as a “part-to-part” approach. Using the KAt and CMTH datasets, the authors demonstrated a high correlation between signals originating from the same bearing---violating the i.i.d. assumption---and showed the substantial drop in accuracy when shifting from a condition-wise to a bearing-wise split, revealing the effects of a bearing-level leakage. Although this paper propose a similar analysis to ours, their experiments contain confounding factors that make it difficult to understand the origin of the decrease in model performance. In their work, data splitting strategies such as ``run-to-run'', ``day-to-day'' and ``part-to-part'' were proposed. Each strategy corresponds to completely different training set compositions, making it difficult to identify which one is more diverse. One could argue that the decrease in performance when changing to the ``part-to-part'' split is associated with the model being less diverse, rather than uniquely due to data leakage. Our paper focuses on solving this issue by proposing controlled data leakage experiments, where the training set remains fixed while varying the test sets.

It is worth mentioning that a few studies using the PU dataset have apparently adopted strict bearing-wise splits and reported strong results, such as \cite{wang2021} and \cite{wang2023}. This approach was also employed in the original PU paper \cite{lessmeier2016condition}, which reported a 98.3\% multiclass accuracy. Although these works provide clear instructions on their proposed splitting strategies, we were unable to reproduce the same results in our own experiments.


\section{Methodology}
\label{sec:methodology}
In this section, we first introduce our general methodology, which is exemplified using a hypothetical generic dataset; then,
we apply and specialize this methodology to three public bearing diagnosis datasets: University of Ottawa (UORED-VAFCLS), Paderborn University (PU), and Case Western Reserve University (CWRU). \new{Appendix \ref{app:c} further demonstrates the application of the proposed methodology to the HUST bearing dataset.} Finally, we describe the features, deep learning architectures, and data augmentation techniques used in our experiments.

\subsection{General Methodology}

Our general methodology consists of three parts: the bearing-wise data splitting strategy; our problem formulation as multi-label binary classification with ROC-based evaluation metrics; and a hyperparameter tuning and model evaluation protocol chosen to minimize bias.

\subsubsection{Data Splitting}

\begin{figure}
\centering
\begin{tabular}{l}
    \includegraphics[width=8cm]{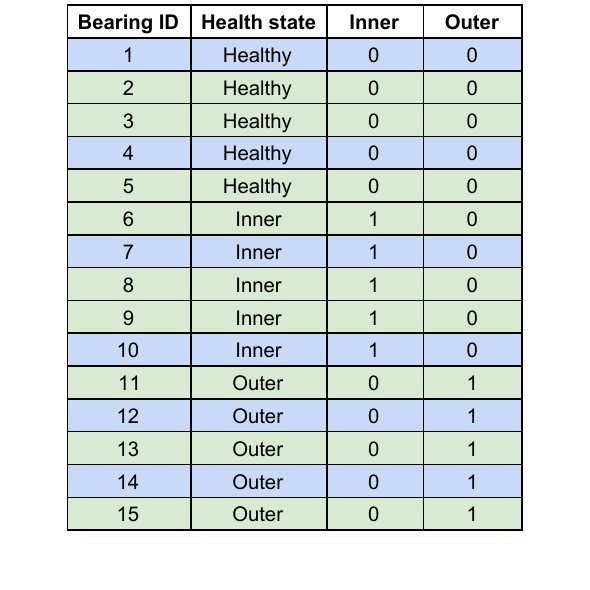} \\ 
\end{tabular}
\caption{Exemplary bearing-level data partitioning for the generic dataset. The training set (green) and test set (blue) are disjoint at the bearing level, with a 3:2 allocation of bearings per health state.}
\label{tab:gen-bearings}
\end{figure}

Central to our methodology is a strict, bearing-wise data partitioning strategy designed to prevent data leakage and ensure a valid assessment of model generalization. To illustrate this principle, we define a generic dataset construct, specified in Figure \ref{tab:gen-bearings}. This dataset comprises $B=15$ unique physical bearings, each characterized by one of three health states: healthy, inner race fault, or outer race fault.

\begin{figure}
\centering
\begin{tabular}{l}
    \includegraphics[width=12cm]{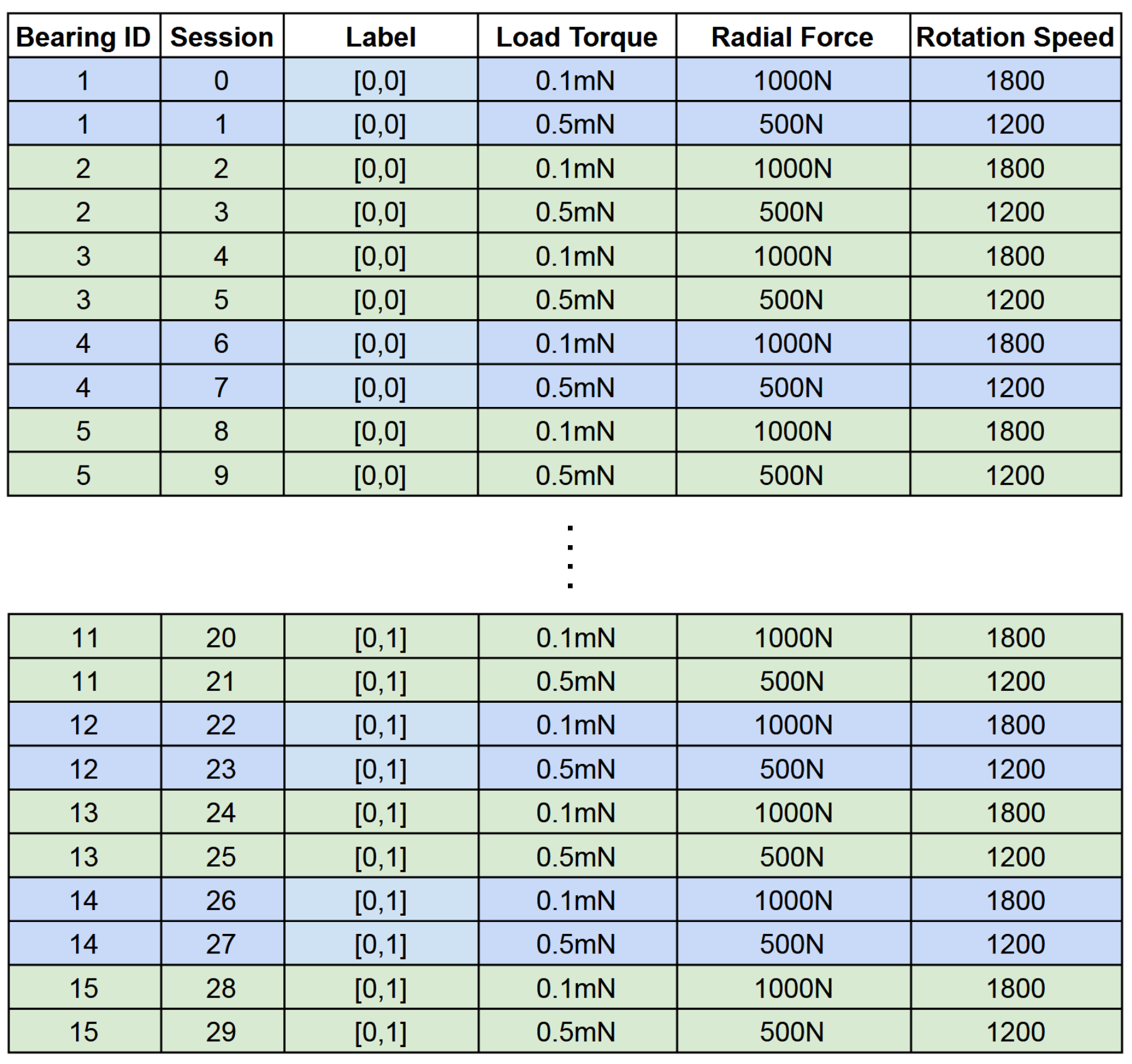} \\ 
\end{tabular}
\caption{Specification of the generic bearing fault dataset, comprising 15 unique bearings, two fault modes (inner, outer), and two distinct acquisition configurations per bearing.}
\label{tab:gen-table}
\end{figure}

Under our multi-label framework, these states are represented by binary vectors, where a healthy bearing is encoded as [0,0], an inner race fault as [1,0], and an outer race fault as [0,1]. The partitioning of this dataset, which is detailed in Figure \ref{tab:gen-table}, adheres to a 3:2 train-to-test ratio applied at the bearing level. Specifically, for each health state, three distinct bearings are allocated to the training set, while the remaining two are reserved for the test set. This ensures that no data from a single physical bearing appears in both the training and test partitions, thereby creating a realistic scenario for evaluating performance on unseen components.

\subsubsection{Problem formulation and evaluation metric}
\label{sec:problem-formulation}
 
The fundamental objective in bearing health monitoring is fault detection, the ability to reliably distinguish between normal and faulty operating conditions. While the secondary objective of diagnosis (identifying the specific fault mode) is critical, it relies heavily on the robustness of this initial detection. The predominant problem formulation in literature combines detection with diagnosis using a multiclass framework, in which the healthy condition is treated as one of the classes, alongside fault types. Under this formulation, accuracy is widely used as the evaluation metric, although it presents significant limitations in this context. Standard multiclass accuracy is an inadequate measure of detection quality due to the inherent class imbalance in typical fault diagnosis datasets, which contain a disproportionate number of faulty samples compared to healthy ones. Consequently, relying on a multiclass formulation can be misleading, as a model may achieve a high score despite classifying all healthy samples as one or more fault classes, failing to reflect performance in real-world scenarios where the healthy class is more prevalent.

One approach to mitigating the limitations of a multiclass framework involves decoupling the tasks of detection and diagnosis into a two-stage process. In the first stage, fault detection is treated as a standalone binary classification problem. To account for the characteristic class imbalance of these datasets, one might employ a prevalence-independent metric, as discussed in Section~\ref{sec:evaluation-metrics}. In the second stage, a multiclass framework can be used for diagnosis, although it introduces significant practical drawbacks. The traditional multiclass formulation assumes that fault modes are mutually exclusive (only one can exist) and exhaustive (all possibilities are covered), which precludes the detection of co-occurring faults and complicates the handling of novel defect types. While it is possible to create classes that account for co-occurring faults, it is often unpractical and limited, specially due to the lack of samples in public datasets that correspond to these cases. 
Furthermore, accuracy is often a misleading metric because it lacks the nuance to reflect varying importance levels. In contexts where one fault class is significantly more damaging than others, accuracy fails to penalize critical misses more heavily than minor ones. This makes it difficult to optimize a model for scenarios where certain errors are far more costly than others.

To address these limitations and provide a simpler structure that combines detection and diagnosis, we advocate for a multi-label framework that treats each fault type (excluding the healthy state, e.g., Inner, Outer, Ball) as an independent binary classification problem. 
In other words, the model outputs a yes/no classification for each fault type, with the healthy state being understood as the case where no fault type is present.\footnote{Naturally, this interpretation assumes that all possible fault types are included as classifier outputs. If additional unmodeled faults may occur, then the case of no fault detected should not be interpreted as a healthy state, but simply as absence of known faults.} Under this framework, conventional fault detection amounts to simply verifying if \textit{any} of the specific detectors gives a positive output, while fault diagnosis amounts to retrieving \textit{which} specific detectors give a positive output.
This approach enables a nuanced evaluation through the ROC curve, providing the ability to independently select a decision threshold for each classifier. In particular, operating points can be adjusted to meet application-specific requirements, such as prioritizing a high TPR for more critical faults. It also provides transparent, fault-specific insight into classifier behavior, making the reliability of each detector explicit.  An additional benefit is that the multi-label approach allows for the use of faulty signals of one label (such as Inner) as true negatives for all other fault types (e.g. Outer and Ball), which is an advantage over the multiclass formulation, specially in cases where healthy signals may be scarce, such as the CWRU dataset. 

While the ROC curve provides a detailed characterization of classifier behavior across all possible decision thresholds and is highly recommended for final evaluation, its interpretation can be impractical during model development, where concise indicators are typically preferred. The AUROC summarizes the information conveyed by a ROC curve into a single scalar value, as described in Section~\ref{sec:evaluation-metrics}. In a multi-label framework, where each fault type produces its own ROC curve, we therefore adopt the Macro AUROC as the primary metric for model development, providing an overall view of performance across all classes for hyperparameter tuning and model comparison.

\subsubsection{Hyperparameter Tuning and Model evaluation}

For hyperparameter optimization (also called model selection) and performance evaluation, we adopt the Double Cross-Validation Method (CVM-CV), described in \cite{tsamardinos2015performance}. This protocol consists of applying the cross-validation method for hyperparameter optimization (CVM) and then reevaluating it on different train-test splits (CV) for final performance estimation for the single, selected, best model. Note that using only CVM is well-known to overestimate performance since it returns the maximum performance achieved across several hyperparameter configurations. This bias can be reduced by reevaluating only the selected hyperparameter configuration on different train-test splits.

Our implementation follows a two-stage process:
\begin{itemize}
    \item  \textbf{Hyperparameter Optimization (CVM):} An inner cross-validation loop is employed exclusively for identifying the optimal hyperparameter set for a given model. To accommodate the specific structures of the public datasets, this stage was adapted: for the PU and OU datasets, a 5-run random train-test split was used, while for the more constrained CWRU dataset, a 3-fold partitioning was applied. The hyperparameter configuration yielding the highest average performance in this inner loop was selected for the next stage.
    \item \textbf{Performance Estimation (CV):} The model, using the selected hyperparameters, is retrained and evaluated across 100 distinct, randomly generated train-test splits. Although some test data in this stage may have been seen during hyperparameter tuning, the large number of disjoint splits dilutes the influence of any specific instance. The final reported performance is the Macro AUROC over these 100 runs, providing a more stable and representative estimate of the model's generalization ability.
\end{itemize}

\subsection{Specific details for each dataset}
\label{sec:datasets}

\subsubsection{University of Ottawa (UORED-VAFCLS)}
\label{subsec:uored-methodology}

The University of Ottawa Rolling-element Dataset -- Vibration and Acoustic Faults under Constant Load and Speed conditions (UORED-VAFCLS) \cite{ottawa} provides a contemporary benchmark for fault diagnosis methodologies, offering multi-modal data streams including acoustic, vibration, and temperature signals. The dataset encompasses four distinct fault modes: inner race, outer race, ball, and cage. For each fault category, five unique physical bearings were tested, resulting in a total of 20 distinct components. All acquisitions were conducted under a single, fixed operating condition (500N load, 1750 RPM) with signals recorded for 10 seconds at a 42 kHz sampling rate.

A notable characteristic of the UORED-VAFCLS dataset is its hierarchical structure: each of the 20 physical bearings was recorded across three progressive health states: 1) normal operation, 2) weak fault severity, and 3) strong fault severity. This design results in a total of 60 discrete time-series signals. The composition of the dataset is summarized in Table \ref{tab:ott-bearings}.
\begin{table}
\centering
\caption{Bearing-level structure of the University of Ottawa (UORED-VAFCLS) dataset, detailing the 20 unique bearings across four fault categories.}
\resizebox{\textwidth}{!}{%
\begin{tabular}{cclcclcclcc}
\textbf{Bearing ID} & \textbf{Health state} &  & \textbf{Bearing ID} & \textbf{Health state} &  & \textbf{Bearing ID} & \textbf{Health state} &  & \textbf{Bearing ID} & \textbf{Health state} \\ \cline{1-2} \cline{4-5} \cline{7-8} \cline{10-11} 
1 & Healthy &  & 6 & Healthy &  & 11 & Healthy &  & 16 & Healthy \\
1 & Inner-1 &  & 6 & Outer-1 &  & 11 & Ball-1 &  & 16 & Cage-1 \\
1 & Inner-2 &  & 6 & Outer-2 &  & 11 & Ball-2 &  & 16 & Cage-2 \\
2 & Healthy &  & 7 & Healthy &  & 12 & Healthy &  & 17 & Healthy \\
2 & Inner-1 &  & 7 & Outer-1 &  & 12 & Ball-1 &  & 17 & Cage-1 \\
2 & Inner-2 &  & 7 & Outer-2 &  & 12 & Ball-2 &  & 17 & Cage-2 \\
3 & Healthy &  & 8 & Healthy &  & 13 & Healthy &  & 18 & Healthy \\
3 & Inner-1 &  & 8 & Outer-1 &  & 13 & Ball-1 &  & 18 & Cage-1 \\
3 & Inner-2 &  & 8 & Outer-2 &  & 13 & Ball-2 &  & 18 & Cage-2 \\
4 & Healthy &  & 9 & Healthy &  & 14 & Healthy &  & 19 & Healthy \\
4 & Inner-1 &  & 9 & Outer-1 &  & 14 & Ball-1 &  & 19 & Cage-1 \\
4 & Inner-2 &  & 9 & Outer-2 &  & 14 & Ball-2 &  & 19 & Cage-2 \\
5 & Healthy &  & 10 & Healthy &  & 15 & Healthy &  & 20 & Healthy \\
5 & Inner-1 &  & 10 & Outer-1 &  & 15 & Ball-1 &  & 20 & Cage-1 \\
5 & Inner-2 &  & 10 & Outer-2 &  & 15 & Ball-2 &  & 20 & Cage-2
\end{tabular}%
}

\label{tab:ott-bearings}
\end{table}

This multi-state-per-bearing structure does not compromise the integrity of our proposed methodology. On the contrary, it underscores the necessity of bearing-level partitioning. By assigning all signals from a single physical component exclusively to either the training or the test set, we rigorously prevent data leakage and ensure a valid assessment of the model's ability to generalize to entirely unseen hardware.

To implement our CVM-CV protocol, we systematically partitioned the dataset. For each of the four fault modes, the five available bearings were split into a 3:2 train-test ratio. The number of unique ways to select three of the five bearings for the training set is equal to 10. As the selection for each fault mode is independent, the total combinatorial space of unique splits is 10$^4$. From this space, we instantiated 105 distinct splits for our experiment.

\begin{figure*}
    \centering
    \includegraphics[width=\linewidth]{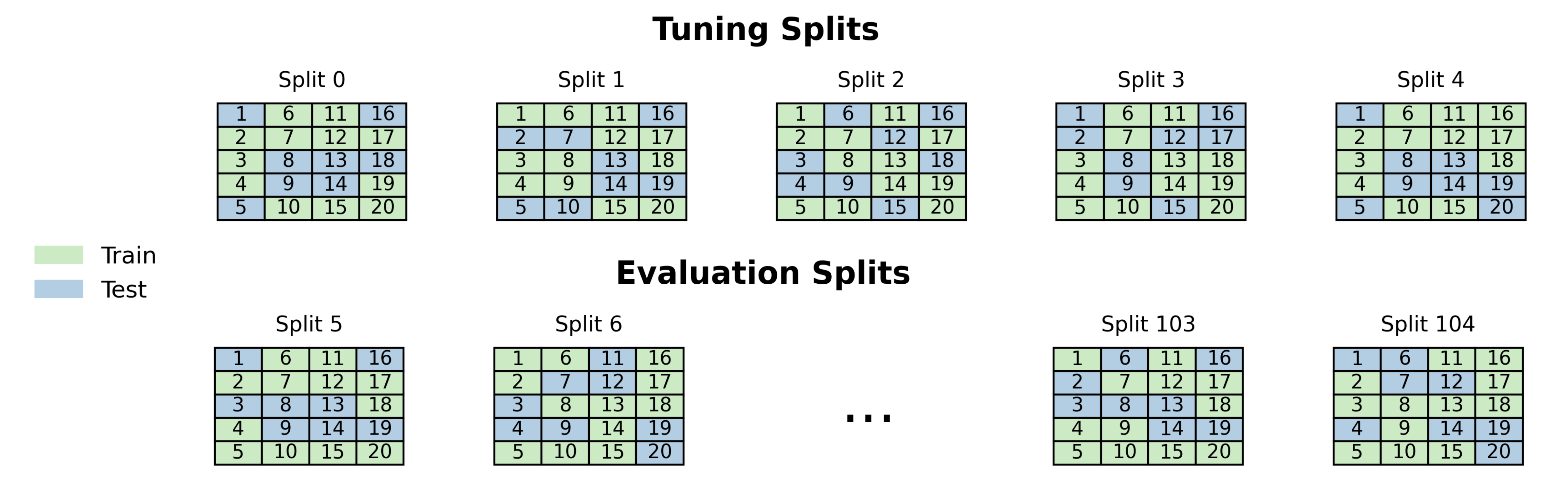}
    \caption{Schematic of the Double Cross-Validation (CVM-CV) protocol applied to the UORED-VAFCLS dataset. A distinct set of 5 bearing-level splits is used for hyperparameter tuning, while a separate set of 100 splits is used for final performance evaluation.
}
    \label{fig:ott-splits}
\end{figure*}

As illustrated in Figure \ref{fig:ott-splits}, these splits were strictly segregated:
\begin{itemize}
    \item \textbf{Hyperparameter Tuning (CVM):} The first 5 unique splits were used exclusively for the inner cross-validation loop to perform model selection.
    \item \textbf{Performance Estimation (CV):} The subsequent 100 disjoint splits were reserved for the outer loop to evaluate the performance of the selected model.
\end{itemize}

This two-tiered approach guarantees that the data combinations used for performance estimation were entirely unseen during the hyperparameter tuning process, thereby yielding a more robust measure of model generalization.

\subsubsection{Paderborn University (PU) Dataset}

The Paderborn University (PU) bearing dataset \cite{lessmeier2016condition} represents a complex and widely-used benchmark, distinguished by its inclusion of bearings from multiple manufacturers and two distinct fault origination paradigms: artificial damage and natural degradation from accelerated lifetime tests. The dataset contains 6 healthy bearings, 12 with artificially induced faults (inner/outer race), and 14 with faults developed during operation (inner race, outer race, and combined inner/outer race). Data was captured under four discrete operating conditions, varying rotational speed, load torque, and radial force, with vibration signals recorded at a 64 kHz sampling rate.

Crucially, prior work by  \cite{lessmeier2016condition} demonstrated that a significant domain shift exists between artificially damaged and naturally degraded bearings, leading to poor generalization when training on the former and testing on the latter. To circumvent this issue and ensure the practical relevance of our findings, our investigation exclusively utilizes the subset of bearings with naturally occurring faults from accelerated lifetime tests, along with the healthy reference bearings. Our analysis incorporates components K006 and KI17, thereby expanding upon the set used in the original benchmark study, which are shown in Table \ref{tab:pd-bearings}. Lastly, to limit computational overhead, all measurements in this dataset were resampled to 42 kHz.
Although bearings with combined faults are in principle compatible with our multi-label methodology, their limited representation in the dataset hinders robust learning and evaluation. In fact, only three such bearings are available (IDs KB23, KB24, and KB27). Therefore, we exclude these samples from both training and testing to prevent biased or unreliable conclusions.

\begin{table}
\centering
\caption{IDs of all healthy and naturally damaged bearings from the PU dataset utilized in this study. The selection expands upon the original benchmark set from \cite{lessmeier2016condition} (highlighted in yellow) by incorporating additional available components (highlighted in green).}
\begin{tabular}{c|c|c}
\textbf{Healthy} &
  \textbf{\begin{tabular}[c]{@{}c@{}}Outer ring \\ damage\end{tabular}} &
  \textbf{\begin{tabular}[c]{@{}c@{}}Inner ring \\ damage\end{tabular}} \\
  \hline
\rowcolor[HTML]{FFF2CC} 
K001                         & KA04                         & KI04                          \\
\rowcolor[HTML]{FFF2CC} 
K002                         & KA15                         & KI14                          \\
\rowcolor[HTML]{FFF2CC} 
K003                         & KA16                         & KI16                         \\
\cellcolor[HTML]{FFF2CC}K004 & \cellcolor[HTML]{FFF2CC}KA22 & \cellcolor[HTML]{FFF2CC}KI18    \\
\cellcolor[HTML]{FFF2CC}K005 & \cellcolor[HTML]{FFF2CC}KA30 & \cellcolor[HTML]{FFF2CC}KI21     \\
\cellcolor[HTML]{D9EAD3}K006 & -                            & \cellcolor[HTML]{D9EAD3}KI17                           
\end{tabular}

\label{tab:pd-bearings}
\end{table}

\begin{figure}
    \centering
    \includegraphics[width=\linewidth]{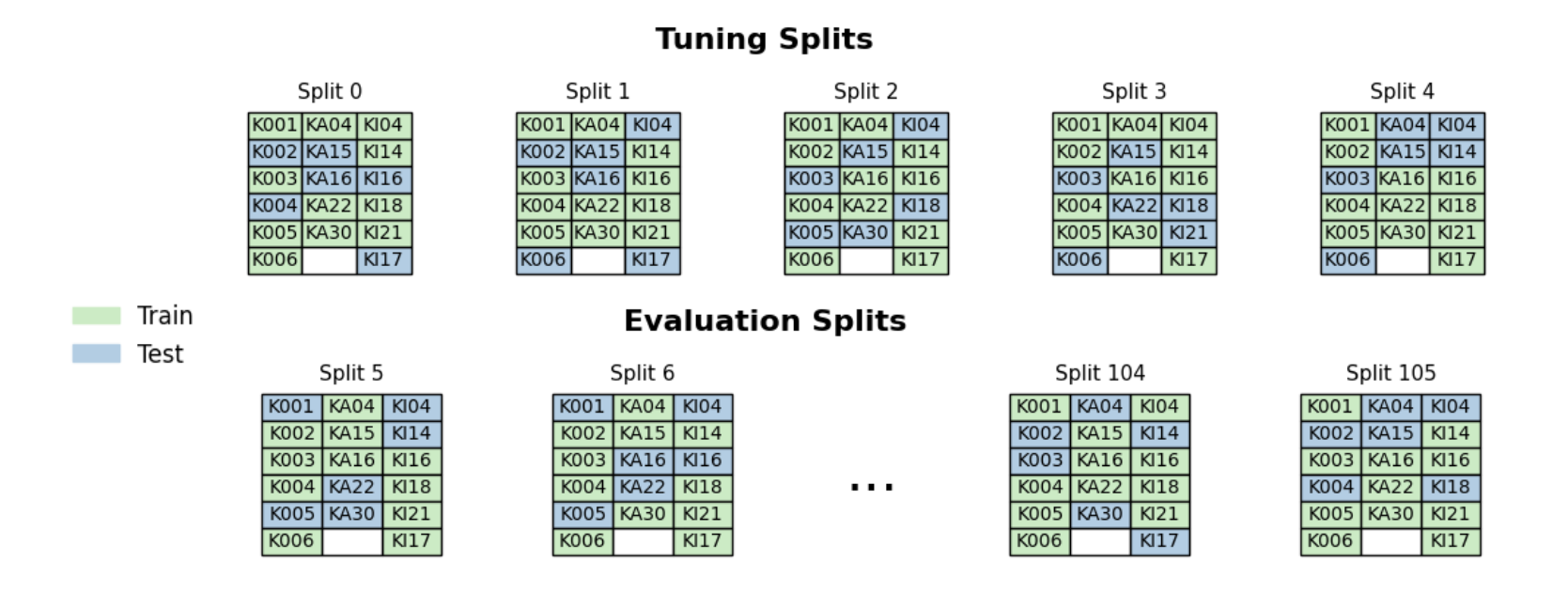}
    \caption{Schematic of the Double Cross-Validation (CVM-CV) protocol applied to the PU dataset.
}
    \label{fig:pd-splits}
\end{figure}

Our partitioning strategy was tailored to the specific composition of this curated subset. To accommodate the varying number of available bearings along the three classes, a differentiated partitioning scheme was adopted for the CVM-CV protocol. For the healthy and inner race fault categories, each of which contains six bearings, a 4:2 train-test split was implemented. For the outer race fault category, with five available bearings, a 3:2 split was used. As depicted in Figure \ref{fig:pd-splits}, the splits designated for hyperparameter tuning and final performance evaluation were kept entirely separate to maintain the integrity of the validation process.

\subsubsection{CWRU bearing fault dataset}
\label{sec:cwru-dataset}
The Case Western Reserve University (CWRU) bearing fault dataset is a collection of experiments that involved a single pair of healthy bearings and several artificially created faulty bearings \cite{smith2015rolling}. The faults were created through electro-discharge machining, introducing point faults with diameters of 7, 14, 21, and 28 mils in the inner race, outer race, and rolling element separately. For the outer race faults, the experiments considered faults located at three different positions relative to the load zone. The healthy and faulty bearings were reinstalled at both the drive end (DE) and fan end (FE) locations (where each configuration comprises either two healthy bearings or one healthy bearing and one faulty bearing), and data were collected synchronously, with one accelerometer at each location. For each configuration, experiments were made using four operational motor load conditions ranging from 0 (no load) to 3 horsepower (HP). In most cases, the experiments used a 12 kHz sampling rate, while some used 48 kHz. All the experiments consist of signals that are approximately 10 seconds long.

\begin{figure}
    \centering
    \includegraphics[width=0.8\linewidth]{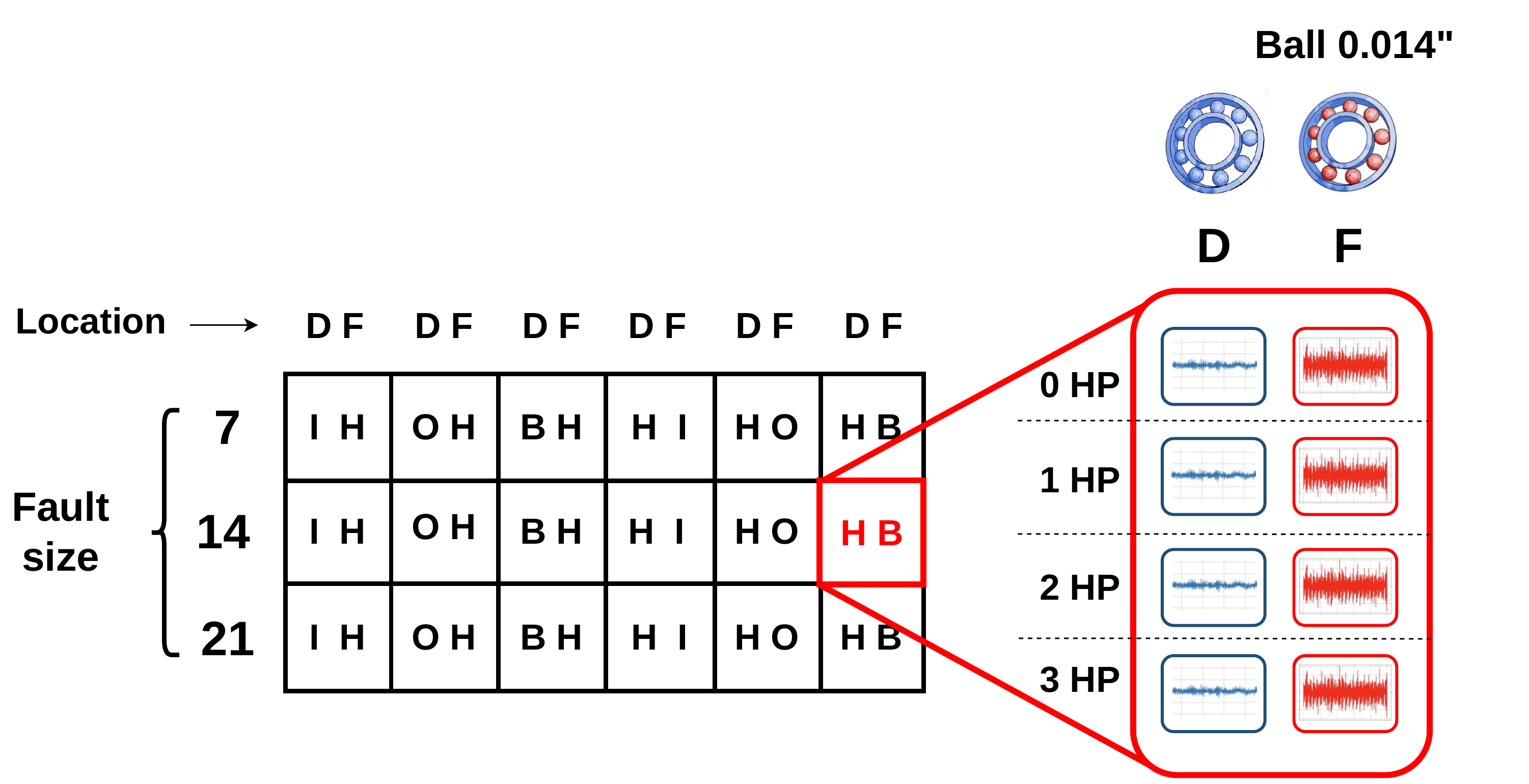}
    \caption{Bearing configurations used in the CWRU dataset. Each cell represents a specific acquisition setup containing two bearings: one located at the drive end (D) and the other at the fan end (F). Fault types are denoted as follows: I for inner race fault, O for outer race fault, B for ball fault, and H for healthy.}
    \label{fig:cwru-configurations}
\end{figure}

\new{In this paper, the evaluated configurations considered loads ranging from 0 to 3 HP and fault sizes of 7, 14, and 21 mils. 
For the faulty configurations (consisting of one faulty bearing and one healthy bearing), signals acquired at both locations were included, while the healthy configuration (consisting of two healthy bearings) was entirely excluded.
}
Measurements with a sampling rate of 12 kHz were used, \new{and signals that} had a sampling rate of 48 kHz were resampled to 12 kHz. Considering the three different fault positions at the outer race fault experiments, the ``Centered @6:00'' experiments were primarily used whenever possible. If the former did not exist, the ``Orthogonal @3:00'' experiments were used, following \cite{hendriks2022towards}. All these configurations can be seen in Figure~\ref{fig:cwru-configurations}, where each box represents a different bearing configuration. 
\new{Since there is one sensor at each side, we consider that each sensor monitors its co-located bearing; in particular, signals acquired at the location of a healthy bearing were considered healthy signals. Considering four load conditions per configuration, there are a total of 72 faulty signals (evenly distributed across fault type, fault size and location) and 72 healthy signals (evenly distributed across location). Note that each faulty configuration corresponds to a unique faulty bearing, so each faulty bearing contributes exactly 4 signals; in contrast each of the two healthy bearings contributes 36 signals.}

\begin{figure}
    \centering
    \includegraphics[width=\linewidth]{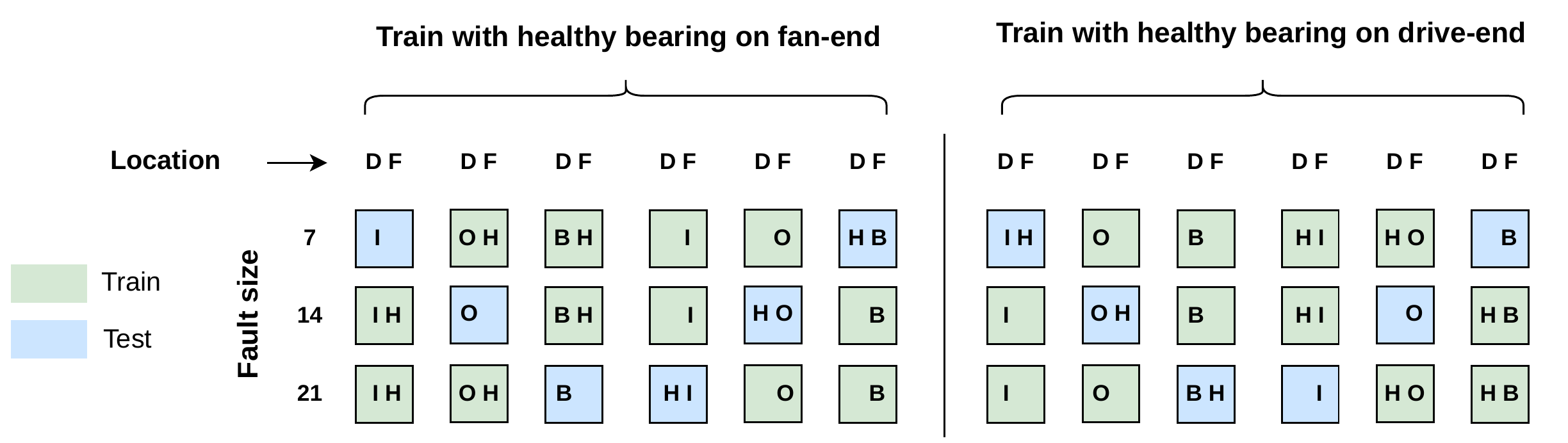}
    \caption{\new{Illustration of the proposed splitting methodology for the CWRU dataset. For each column, which represents bearings from a specific fault mode on a given location (e.g. column 1 represents inner-race fault bearings on the drive end), two faulty bearings are placed in the training set (green), while the remaining one is placed in the test set (blue). This figure also details two possible splitting scenarios that take into account the healthy bearings: training with healthy signals from the fan end (left panel) and training with healthy signals from the drive end (right panel).} Green cells indicate signals used for training, whereas blue cells indicate signals reserved for testing.}
    
    \label{fig:cwru-trainsides}
\end{figure}

\new{Following our methodology, we propose a strict bearing-wise split. This is easier to achieve for the faulty bearings and is explained first. As mentioned above, each cell in Figure \ref{fig:cwru-configurations} represents a unique faulty bearing, characterized by a unique combination of fault size, fault type and fault location. Our splitting configuration uses a ratio of 2:1, where 12 faulty bearings are included in the training set, while the remaining 6 are reserved for the test set. We choose to stratify by faulty type and fault location and randomize the fault size. Thus, for each fault location-type pair (represented as a column in Fig.~\ref{fig:cwru-configurations}), we randomly choose one fault size (a row in Fig.~\ref{fig:cwru-configurations}) and place the resulting faulty bearing in the test set; the remaining faulty bearings are placed in the training set. An example is illustrated in Figure~\ref{fig:cwru-trainsides}.}

\new{Note, however, that we cannot place the entire configuration (including the healthy signals) in the train/test sets as described above, since each fault location-type pair corresponds to a single healthy bearing (at the opposite side of the fault), in which case we would end up with the same healthy bearing being used for training and testing, causing data leakage. Since there are only two healthy bearings, there are only two possible leakage-free choices: either we train with the DE healthy bearing and test with the FE healthy bearing or vice-versa. This can be achieved by excluding DE healthy signals from the test set and FE healthy signals from the training set (or vice-versa). Thus, for each train-test split of the faulty bearings, we produce two different splits of the healthy bearings, as illustrated in Fig.~\ref{fig:cwru-trainsides}.}%
\footnote{\new{In principle, one could also split the single healthy bearing configuration consistently with the chosen split of the healthy signals from the faulty configurations (e.g., assigning signals from the DE to the training set and signals from the FE to the test set when the split of the faulty configurations places healthy signals from DE on the training set). However, because these signals from the single healthy bearing configuration are acquired simultaneously, it is conceivable that vibrations from the opposite bearing may introduce shared interference, potentially leading to data leakage. For this reason, these signals are excluded from our experiments.}}

\new{In addition to this hold-out strategy, a 3-fold cross-validation procedure is applied for hyperparameter optimization following the same selection principles. The first fold is generated by randomly selecting a bearing for each fault location-type pair. The remaining two folds are then created by applying additional hold-out splits on the residual configurations, excluding the first fold. This procedure yields three train-test configurations, each using two folds for training and the remaining fold for testing in a 2:1 proportion. Since each configuration involves training and testing with healthy signals on each side (DE/FE), this results in six experiments in total. To perform evaluation, we created 50 additional 2:1 splits, with each split supporting two train-test configurations, yielding a total of 100 evaluation runs.}

\subsubsection{Summary}

Table \ref{tab:summary} summarizes the main characteristics of the bearing diagnosis datasets considered in this study. Note that, to facilitate the comparison of dataset diversity, we introduce the notion of [bearing, health state] instances. As discussed in the previous sections, the UORED-VAFCLS dataset associates each bearing with three distinct health states, two of which represent different fault severities. Consequently, although the dataset comprises 20 bearings, it yields 60 unique [bearing, health state] instances, offering greater diversity. In contrast, the CWRU and PU datasets provide a larger number of signals overall (specially PU), but only one health state per bearing, resulting in roughly three times fewer instances.

\begin{table}
\caption{Characteristics of the bearing diagnosis datasets.}
\centering
\begin{minipage}{\textwidth}
\centering
\renewcommand{\footnoterule}{}
\begin{tabular}{cccccc}
\toprule
\textbf{Dataset} &
\textbf{\makecell{Number\\ of bearings}} &
\textbf{\makecell{Health states\\ per bearing}} &
\textbf{\makecell{Number of\\ {[}bearing, health state{]}\\  instances}} &
\textbf{\makecell{Number of\\ faulty instances}} &
\textbf{\makecell{Number of faulty\\ instances in training}} \\ 
\midrule
UORED-VAFCLS         & 20                   & 3                    & 60                   & 40                   & 24                   \\
CWRU                 & 20\footnote{Excluding bearings with a 0.028'' fault size.}                   & 1                    & 18\footnote{We only consider instances with a faulty bearing accompanied by a healthy bearing at the opposite side (see Section~\ref{sec:cwru-dataset}).}                  & 18                   & 12                   \\
Paderborn            & 17\footnote{These bearings correspond to the group with real faults, excluding those with combined faults.}                  & 1                    & 17                   & 11                   & 7                    \\
\midrule
\textbf{Dataset} &
\textbf{\makecell{Acquisition\\ locations}} &
\textbf{\makecell{Conditions\\ per instance}} &
\textbf{\makecell{acquisitions\\ per condition\\ per location}} &
\textbf{\makecell{Total number\\ of signals}} &
\textbf{\makecell{Total number of\\ signals in training}} \\
\midrule
UORED-VAFCLS                              & 1                    & 1                    & 1                    & 60    & 36               \\
CWRU                                      & 2                    & 4                    & 1                    & 144     & 72             \\
Paderborn                                 & 1                    & 4                    & 20                   & 1360        & 880        \\
\bottomrule
\end{tabular}%
\label{tab:summary}
\end{minipage}
\end{table}

\subsection{Models and Training Details}

\subsubsection{Shallow Learning models with bearing fault based features
}
\label{sec:shallow}
In shallow (i.e., non-deep) learning experiments, we adopt feature-based approaches using classical machine learning classifiers---specifically Random Forest and Support Vector Machines (SVM). The input features consist of well-established signal descriptors from the literature. Feature extraction is grounded in signal processing techniques applied to vibration data in both time and frequency domains, with prior studies demonstrating their effectiveness for bearing fault diagnosis.

In the work of \cite{abburi2023closer}, statistical metrics extracted from time-domain signals were employed, including mean, absolute median, standard deviation, skewness, kurtosis, crest factor, energy, RMS value, peak count, zero-crossing count, Shapiro-Wilk test statistic, and Kullback-Leibler divergence. In the frequency domain, \cite{alonso:2023} explored the use of bearing-specific fault frequencies, such as BPFI (Ball Pass Frequency of the Inner race), BPFO (Ball Pass Frequency of the Outer race), and BSF (Ball Spin Frequency). Both works reported favorable results using these handcrafted features, which motivated the inclusion of a subset of them in our shallow learning pipeline. 
To highlight modulated fault components distributed across the spectrum, envelope analysis is performed to shift these components into the baseband, following the principles explained in \cite{smith2015rolling}. Subsequently, algorithms are applied to extract the magnitudes of spectral peaks. In this study, the envelope spectrum is computed in the frequency range of 500 Hz to 10 kHz for the PU and UORED-VAFCLS datasets, which offer high sampling rates\footnote{In \cite{smith2015rolling}, envelope analysis is performed without pre-filtering (wide-band analysis). Although the author notes that this approach is sufficient for the CWRU dataset, we restrict the bandwidth to exclude low frequencies (below 500 Hz) and very high frequencies (above 10 kHz), while still preserving a broad range that is potentially applicable to most signals. An inspection of the artificially damaged bearings on PU revealed that the majority of the relevant spectral content is concentrated within this interval.}. For the CWRU dataset, limited to a 12 kHz sampling rate, a reduced band of 500 Hz to 6 kHz is adopted. Fault frequencies are computed based on the rotation speeds and geometric parameters provided within each dataset. Table \ref{tab:features} shows all features used on shallow learning models.

\begin{table}
\centering
\caption{Time and \new{envelope spectrum} domain hand-crafted features for Shallow Learning models.}
\begin{tabular}{cc}
\toprule
\textbf{Set} &
  \textbf{Features} \\  \midrule
Time + \new{Envelope spectrum} &
  \makecell{RMS, peak-to-peak, kurtosis, skewness, crest factor,\\ magnitude of all bearing fault frequencies (1x - 5x) \\ on envelope spectra}\\ \bottomrule
\end{tabular}
\label{tab:features}
\end{table}

A notable limitation of shallow learning approaches lies in their reliance on metadata---particularly rotational speed (RPM) and bearing geometry---for the computation of specialized features. Furthermore, these models do not generalize naturally across datasets with differing fault types or label structures. As a result, we must train a separate classifier for each fault mode, increasing the complexity of model management. In contrast, deep learning models can learn representations directly from raw signals and are capable of handling varied datasets within a unified architecture.

Finally, our preprocessing of the training and test signals consisted of segmenting them into non-overlapping 1-second windows.

\subsubsection{Deep Learning architectures}

Informed by a review of contemporary deep learning models for vibration analysis, we selected the Wide First-layer Kernel Deep Convolutional Neural Network (WDCNN) \cite{wdcnn} as the primary architecture for our investigation. This model, with approximately 172k parameters for an input segment length of 12,000 samples, offers a well-established baseline. To benchmark its performance against more recent or complex 1D convolutional architectures, a comparative analysis was conducted on the UORED-VAFCLS dataset, evaluating the WDCNN against the CDCN \cite{cdcn}, WDTCNN \cite{wdtcnn}, 1D-ConvNet \cite{dcase2018}, and RESNET1D \cite{resnet1d} models. 
The results of this comparative study are presented in Section \ref{sec:ott-results}. To implement our multi-label classification framework, the output layer of each architecture was configured with a number of nodes equal to the number of detectable fault modes (excluding the healthy state). Each output node employs a sigmoid activation function, effectively operating as an independent binary classifier for a specific fault type. Model training was carried out using the Binary Cross-Entropy loss function, which is well suited for this multi-label setting.

A critical challenge posed by the CWRU and UORED-VAFCLS datasets is the limited number of acquisitions, which substantially increases the risk of model overfitting. To mitigate this, we implemented a data augmentation strategy combining two techniques. The first, Random Crop, generates training samples by extracting segments of a predefined length (e.g., 42,000 samples, corresponding to 1 second for UORED-VAFCLS) from random positions within the full signal. The second, Random Gain (referred to in \cite{randomgain} as Scaling), introduces stochastic amplitude variations by multiplying each signal by a scalar drawn from a normal distribution $\mathcal{N}(\mu, \sigma^2)$. These techniques are applied exclusively to the training set, whereas the test set is preserved as non-overlapping segmented signals, as described in the previous section. 

The augmentation hyperparameters were systematically optimized on the UORED-VAFCLS dataset using the WDCNN architecture. Following \cite{randomgain}, the mean $\mu$ for the Random Gain distribution was fixed at 1.0, while the standard deviation $\sigma$ was tuned over the set {0.3, 0.5, 0.7}. Our validation experiments revealed that the optimal configuration was a combination of Random Crop and Random Gain (RC+RG) with $\sigma =0.7$ for time-domain signals and $\sigma =0.3$ for frequency-domain and spectrum envelope inputs. To maintain a feasible computational budget, these optimized augmentation hyperparameters were subsequently applied without modification to all other datasets and experiments.

For the model selection phase (the inner loop of our CVM-CV protocol), we defined a hyperparameter search space to tune batch size, learning rate, and the input normalization strategy, which is detailed in Table \ref{tab:hparam_space}. We considered standard scaling normalization under two approaches: entry-wise and global. Entry-wise normalization scales each signal using its own mean and standard deviation, whereas global normalization uses the mean and standard deviation computed over the training set. Training was conducted for a fixed duration of 600 epochs across all experiments on the UORED-VAFCLS dataset, which contains 36 training signals under our splitting strategy. This number of epochs was chosen to ensure sufficient gradient updates to obtain stable validation performance curves, while the data augmentation strategies were used to help the curves to reach a plateau without subsequently decreasing, avoiding the need for early stopping. Due to their larger sizes, the CWRU (72 training signals) and PU (880 training signals) datasets required 150 and 30 epochs, respectively, to reach a similar plateau on the validation curves.

\begin{table}
\centering
\caption{Hyperparameter search space for model selection, applied across all datasets.}
\begin{tabular}{ccc}
\toprule
\textbf{Batch size}        & \textbf{Learning Rate} & \textbf{Normalization strategy} \\ \midrule
{[}16, 32, 64, 128, 256{]} & {[}1e-2, 1e-3, 1e-4, 1e-5{]} & {[}none, global, entry-wise{]} \\
\bottomrule
\end{tabular}

\label{tab:hparam_space}
\end{table}

\begin{figure}
    \centering
    \includegraphics[width=\linewidth]{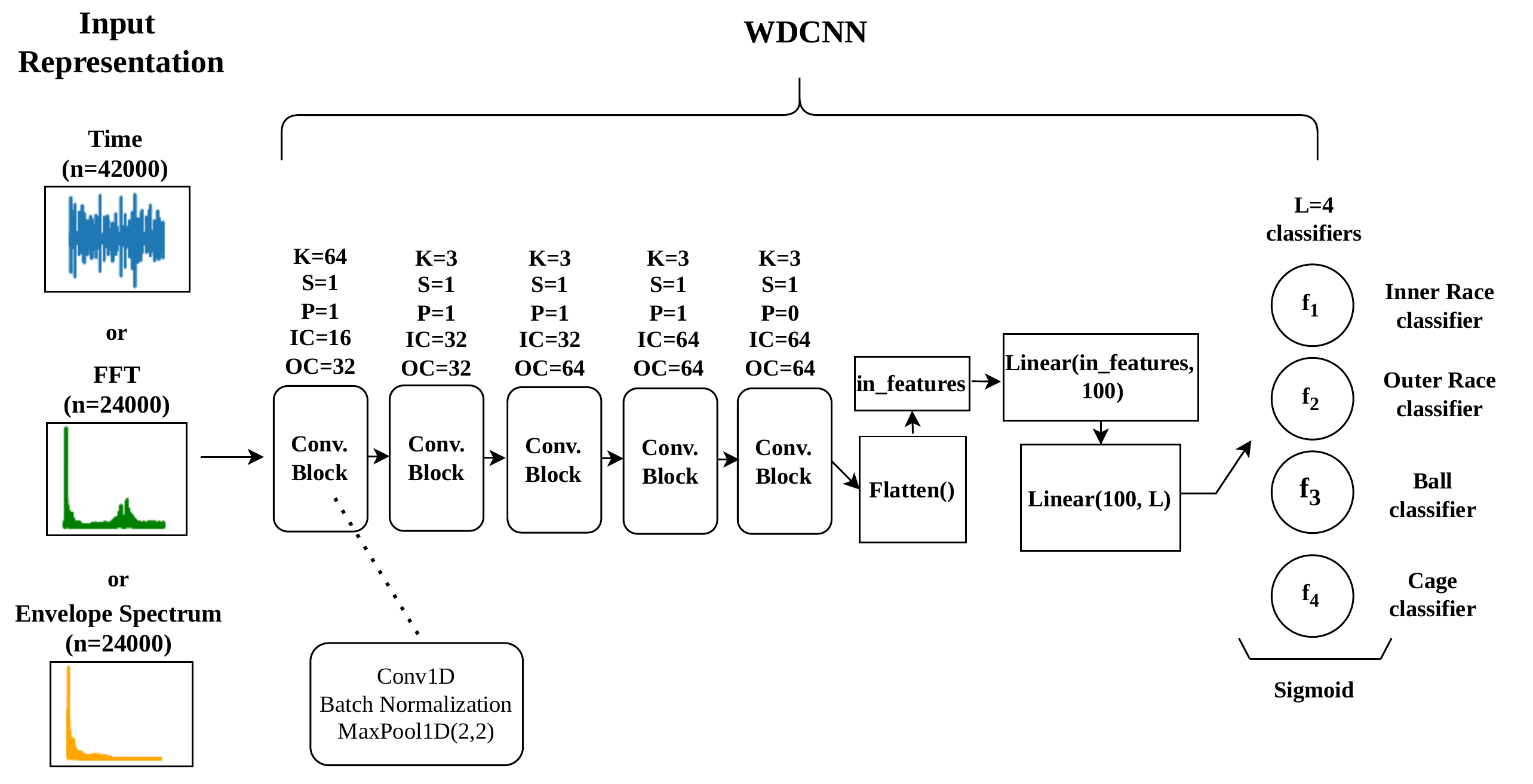}

    \caption{\new{Illustration of the proposed deep learning pipeline based on the WDCNN architecture, adapted for multi-label classification with \textbf{L} binary classifiers, with a choice of three possible input representations. For the UORED-VAFCLS dataset, input segments of length \textbf{n} = 42000 are used, with \textbf{L} = 4 outputs corresponding to the four fault modes present in the dataset. In the figure, each convolutional block is characterized by kernel size \textbf{K}, stride \textbf{S}, padding \textbf{P}, \textbf{IC} input channels, and \textbf{OC} output channels.}}
    \label{fig:dl-pipeline}
\end{figure}

Finally, three input representations were chosen: time-domain signals; frequency-domain \new{representations, obtained through the one-sided FFT computed using the same number of bins as the original segment length; and the envelope spectrum, computed by applying the same one-sided FFT to the temporal envelope. Figure~\ref{fig:dl-pipeline} presents the WDCNN architecture together with the three input representations considered in this work, as well as the modification introduced in the final layer, where a sigmoid activation function replaces the original output mapping to support multi-label classification.}

\section{Experimental results}
\label{sec:experiments}

In this section, we apply the methodology described in Section \ref{sec:methodology} to each dataset. First, the best model architecture is identified through a CVM experiment on the UORED-VAFCLS dataset, comparing performances in the time and frequency domains. Next, we conduct evaluation experiments on all datasets, comparing three input representations across Deep Learning with two Shallow Learning models. Additionally, we test different train-test split proportions in all datasets to assess the impact of the number of bearings on a model’s generalization capacity. \new{Appendix~\ref{app:c} further extends our methodology and experiments to the HUST bearing dataset, including a comparison between the proposed multi-label framework and a conventional multiclass formulation for the detection of co-occurring faults, as well as an assessment of the impact of bearing diversity. Additional details regarding the hyperparameter optimization procedure for shallow learning models, together with the resulting best configurations for all datasets, are provided in Appendix~\ref{app:b}}. 
All experiments were conducted using Python with the PyTorch Lightning framework on a machine equipped with an RTX 3090 GPU and 64GB RAM.

\subsection{UORED-VAFCLS dataset}
\label{sec:ott-results}

The UORED-VAFCLS dataset served as the primary testbed to conduct a rigorous comparative analysis to identify the most effective deep learning architecture and input representation for bearing fault diagnosis. A comparative evaluation of five distinct 1D convolutional neural network architectures was performed using the CVM protocol on the 5 designated tuning splits. The objective was to identify the optimal combination of architecture and hyperparameters across different input representations, such as the time and frequency domains. The envelope spectrum was not included in this comparison due to computational budget constraints. The results of this architecture selection phase, summarized in Table \ref{tab:ott-tuning}, reveal the leading performers.

\begin{table}
\centering
\caption{Best tuning (validation) results on the UORED-VAFCLS dataset across five architectures. Performance is reported as the mean and standard deviation of the Macro AUROC over the 5 designated tuning (validation) splits, along with the optimal hyperparameters identified.}
\begin{tabular}{cccccc}
\toprule
\textbf{Architecture} & \textbf{Input Repr.} & \textbf{Macro AUROC} & \textbf{BS} & \textbf{LR} & \textbf{Normalization} \\
\hline
\multirow{2}{*}{1D-ConvNet} & Time               & 85.99\% ± 7.80\%          & 64  & 0.001  & none \\
                           & Frequency          & 81.25\% ± 6.23\%          & 256 & 0.0001 & none \\
\midrule
\multirow{2}{*}{RESNET1D}  & Time               & 76.05\% ± 4.06\%          & 32  & 0.01   & none \\
                           & Frequency          & 87.6\% ± 10.13\%          & 32  & 0.001  & none \\
\midrule
\multirow{2}{*}{CDCN}      & Time               & 82.45\% ± 10.54\%         & 16  & 0.001  & none \\
                           & Frequency          & 89.36\% ± 10.12\%         & 32  & 0.0001 & none \\
\midrule
\multirow{2}{*}{WDTCNN}    & Time    & 91.09\% ± 5.99\% & 32  & 0.01   & none \\
                           & Frequency          & 87.32\% ± 5.12\%          & 256 & 0.01   & none \\
\midrule
\multirow{2}{*}{WDCNN}     & \textbf{Time}      & \textbf{91.47\% ± 4.87\%} & 128 & 0.0001  & none \\
                           & \textbf{Frequency} & \textbf{93.24\% ± 5.93\%} & 256 & 0.001 & none
\\ \bottomrule
\end{tabular}

\label{tab:ott-tuning}
\end{table}

The WDCNN architecture demonstrated superior performance, achieving the highest mean Macro AUROC scores for both time-domain (91.47\%) and frequency-domain (93.24\%) inputs. The WDTCNN also yielded strong results with time-domain inputs (91.09\%). Based on its consistently high performance and robustness across both input modalities, the WDCNN was selected as the primary architecture for the full performance evaluation.

The final evaluation was conducted using the selected WDCNN model on the 100 disjoint test splits. The results, presented in Table \ref{tab:ou-results}, confirm the superiority of the frequency-domain representation, which achieved a Macro AUROC of 93.12\% ± 4.26\%. This result serves as the primary deep learning benchmark for this dataset. In parallel, we also report the performance of shallow learning models trained on handcrafted features.

\begin{table}
\centering
\caption{Final evaluation results on the UORED-VAFCLS dataset for the selected WDCNN model and corresponding shallow learning benchmarks. Performance is the mean and standard deviation of the Macro AUROC over 100 disjoint test splits.}
\label{tab:ou-results}
\begin{tabular}{ccc}
\toprule
\textbf{Model}                  & \textbf{Input repr. / Features} & \textbf{Macro AUROC}        \\ \midrule
\multirow{3}{*}{\textbf{WDCNN}} & Time                                     & 90.69\% ± 5.43\%          \\
                                & \textbf{Frequency}                       & \textbf{93.12\% ± 4.26\%} \\
                                & Envelope Spectrum                                & 89.86\% ± 6.13\%          \\
                                \midrule
Random Forest                   & Time + \new{Envelope Spectrum}                         & 85.58\% ± 4.74\%          \\
SVM                             & Time + \new{Envelope Spectrum}                         & 81.33\% ± 8.61\%   
\\ \bottomrule
\end{tabular}
\end{table}

\begin{figure}
    \centering
    \includegraphics[width=0.8\linewidth]{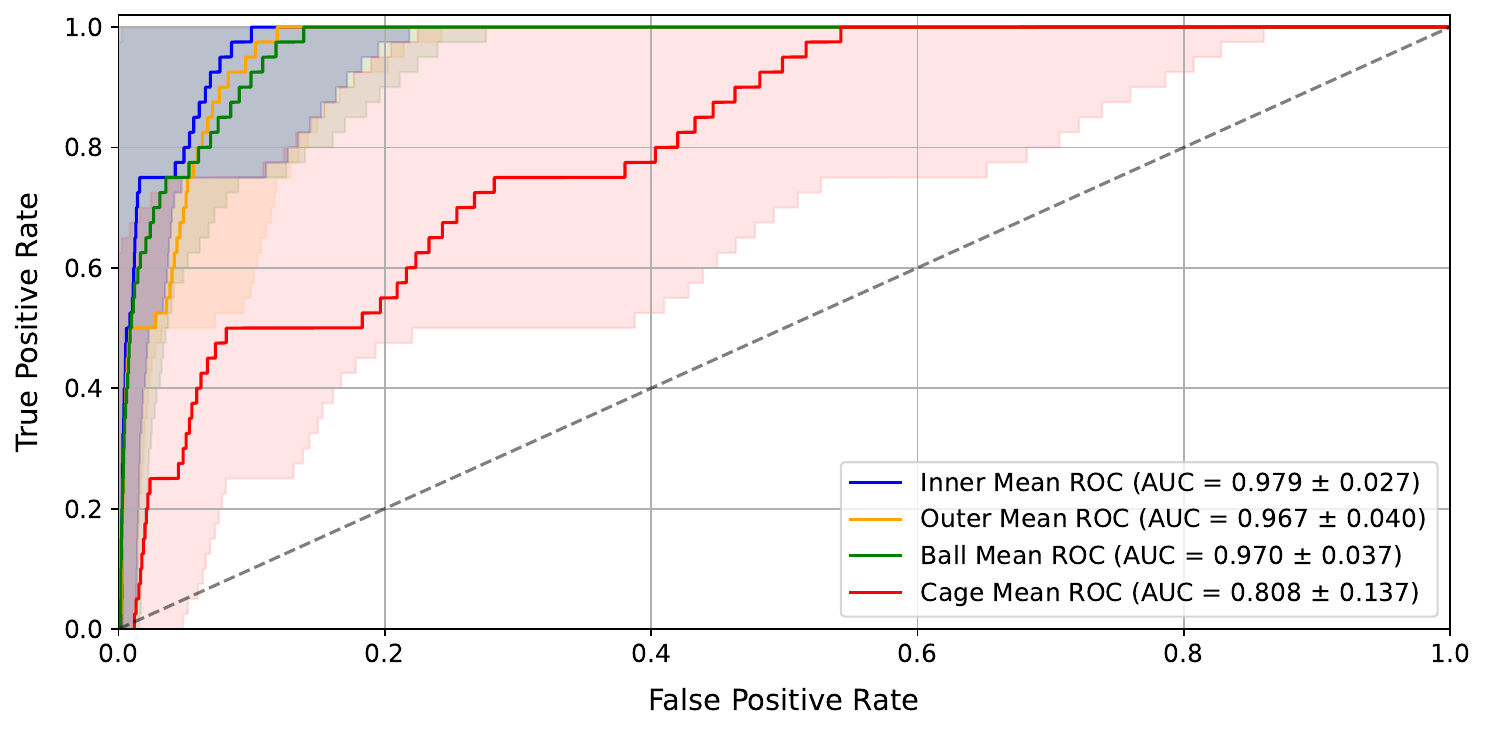}
    
    \caption{ROC curves on the UORED-VAFCLS dataset for the frequency-domain WDCNN. The solid curves represent the horizontal average across 100 runs, while, for each curve, the shaded region represents one standard deviation.
}
    \label{fig:ou-rocs}
\end{figure}

To better understand the performance of each individual classifier in the frequency-domain WDCNN, we analyzed the ROC curves (Figure \ref{fig:ou-rocs}) by calculating the horizontal average \cite{hogan2023on} across 100 runs. 
For instance, at TPR = 90\% (FNR = 10\%), we achieve an average FPR = 6.53\%, 7.59\%, 9.08\% and 46.30\% for the diagnosis of inner, outer, ball and cage faults, respectively.
As can be seen, most classifiers (Inner, Outer, and Ball) show strong performance, achieving AUROC values greater than 96.7\%. In contrast, the Cage classifier showed a significantly decreased performance, yielding an AUC of 80.8\% with a large variance of 13.7\%, indicating a more challenging classification problem for this specific fault type.

It is worth noting that both the original ROC curves and their horizontal averages exhibit step-like behavior, with constant TPR values over ranges of FPR. Although the test set contains up to 240 segments, these are derived from only 24 underlying signals, thus many of these segments are highly similar and receive very similar scores, leading to a limited number of effective operating points. As a result, the lack of smoothness reflects the reduced diversity of the data.

A central tenet of our methodology is the critical role of the number of unique bearings in the training set for model generalization. To empirically validate this principle, we conducted a sensitivity analysis on the UORED-VAFCLS dataset by systematically varying the train-to-test bearing ratio, exploring configurations of 1:4, 2:3, 3:2, and 4:1. \new{These ratios, expressed as N:M, indicate that N bearings per fault mode are assigned to the training set, while the remaining M bearings per fault mode are allocated to the test set.} To isolate the effect of bearing diversity, it was critical to control the total volume of training data seen by each model. \new{We achieved this by adjusting the number of training epochs for each split configuration, scaling it accordingly so that all models were exposed to approximately the same total number of samples from the training set as in the baseline experiment (3:2), while applying the same Data Augmentation techniques. For example, the baseline configuration processes 21,600 training samples (36 signals over 600 epochs), while a 2:3 split provides only 24 training signals; in this case, the number of epochs is increased to 900 to maintain the same total training exposure.} This methodology disentangles the influence of data diversity from the confounding variable of data quantity, clarifying whether performance changes are due to a richer dataset or simply a larger one.

\begin{figure}
    \centering
    \includegraphics[width=0.9\linewidth]{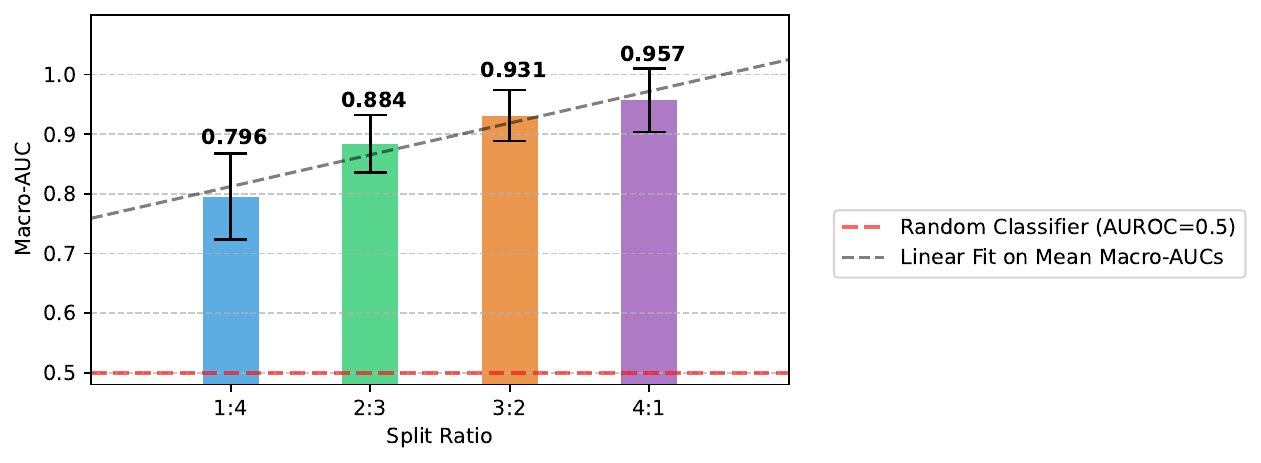}
    \caption{Impact of train-test split ratio on model performance on the UORED-VAFCLS dataset. The plot shows the mean Macro AUROC (and standard deviation as error bars) calculated over 100 evaluation splits for four distinct bearing-level train-to-test ratios.
}
    \label{fig:ou-diversity}
\end{figure}

As hypothesized, a clear trend emerged. As illustrated in Figure \ref{fig:ou-diversity}, increasing the number of training bearings (e.g., a 4:1 ratio) leads to higher mean Macro AUROC scores, as the model benefits from greater component-level diversity during training. However, this comes at the cost of a smaller and less diverse test set, which may limit the reliability of the evaluation---especially in real-world scenarios where generalization to unseen components is critical. On the other hand, splits with more bearings in the test set (e.g., 1:4) provide a broader basis for evaluation, but may yield lower performance due to the reduced diversity in training data. These findings highlight the importance of selecting a partitioning strategy that not only supports model generalization but also ensures that performance estimates are based on sufficiently diverse and representative test sets.

\subsection{PU dataset}

The experiments conducted on the curated Paderborn University (PU) dataset revealed a significant generalization challenge for both time and frequency domains, whereas improved performance was observed when using the envelope spectrum. The final evaluation results, presented in Table \ref{tab:pu-results}, quantify the extent of the generalization challenge. The low mean scores and high standard deviations underscore the model's inability to consistently learn a robust fault signature from the limited training data.

\begin{table}
\centering
\caption{Evaluation results on the curated PU dataset. Performance is reported as the mean and standard deviation of the Macro AUROC over 100 disjoint test splits.}
\begin{tabular}{ccc}
\toprule
\textbf{Model}                  & \textbf{Input Repr. / Features} & \textbf{Macro AUROC}         \\ \hline
\multirow{3}{*}{\textbf{WDCNN}} & Time                                     & 55.08\% ± 19.53\%          \\
                                & Frequency                     & 63.93\% ± 16.06\% \\
                                & \textbf{Envelope Spectrum}                        & \textbf{79.56\% ± 13.07\%} \\
\midrule Random Forest                   & Time + \new{Envelope Spectrum}                         & 69.76\% ± 15.72\%          \\
SVM                             & Time + \new{Envelope Spectrum}                        & 64.41\% ± 15.83\%     \\ \bottomrule    
\end{tabular}

\label{tab:pu-results}
\end{table}

Our primary hypothesis for this poor performance is the limited component diversity in the curated training subset of naturally degraded bearings, especially when compared with the broader variability of the UORED-VAFCLS dataset (Table~\ref{tab:summary}). We first considered the possibility of model underfitting (i.e., insufficient model capacity), but this was ruled out, as the WDCNN consistently achieved near-perfect performance on the training partitions, demonstrating its ability to fit the available data. This observation leads us to conclude that the issue lies not with the model, but with the data itself.

\begin{figure}
    \centering
    \includegraphics[width=0.8\linewidth]{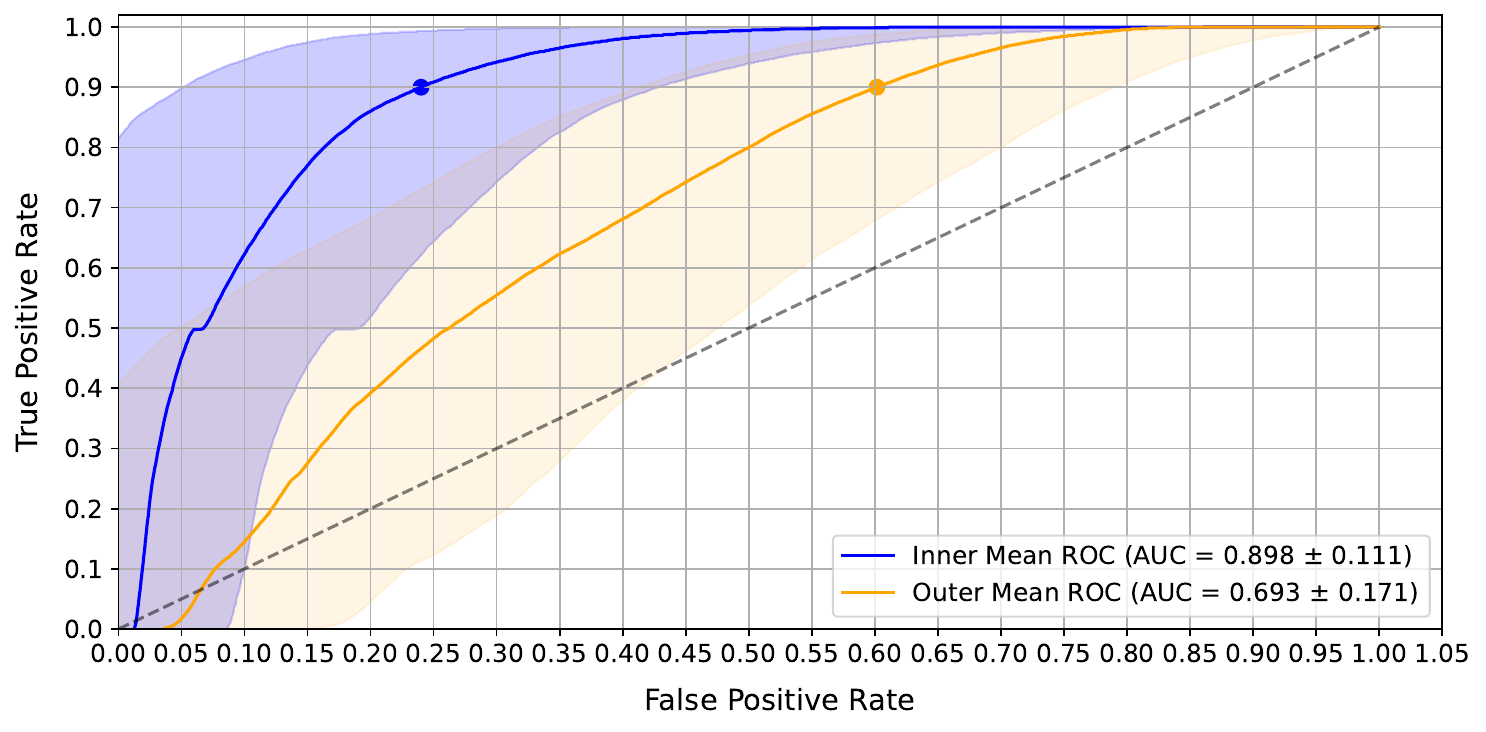}
    
    \caption{ROC curves on the PU dataset for the WDCNN trained with envelope spectrum inputs. The solid curves represent the horizontal average across 100 runs, while, for each curve, the shaded region represents one standard deviation.
}
    \label{fig:pu-rocs}
\end{figure}

Figure \ref{fig:pu-rocs} exhibits the ROC curves obtained for inner and outer race classifiers, where we achieve an average FPR of 24.00\% and 60.15\%, respectively, when fixing $\text{TPR} = 90\%$. From these results, we can observe that the model obtains good performance when classifying inner race faults, reaching a mean AUROC of approximately 90\%. On the other hand, the outer classifier reaches a lower result of 69\% with higher variance. 

While each bearing was recorded under four different operating conditions, our results suggest that this intra-bearing variation is insufficient to compensate for the limited inter-bearing diversity, which is manifested mostly in splits where the model performances reach values around 0.5. In these cases, the unique physical signature of each bearing appears to be the dominant feature, making it difficult for the model to generalize to unseen components. On the other hand, results with Envelope Spectrum, a more specialized input representation, showed us that specialized features may contribute to reduce the impact of memorization. 

To empirically validate this diversity-limitation hypothesis, we replicated the split-proportion analysis performed on the UORED-VAFCLS dataset. Because the number of bearings differs across fault classes (e.g., six healthy bearings and five outer-race bearings) in this dataset, we adapted the experimental protocol to ensure exact train-test proportions. Specifically, for each split configuration, the test set was first constructed by randomly selecting the required number of bearings per health state (e.g., two bearings per class for a 3:2 ratio) and the training set was then formed by randomly sampling the corresponding number of bearings per class from the remaining pool.

\begin{figure}
    \centering
    \includegraphics[width=0.9\linewidth]{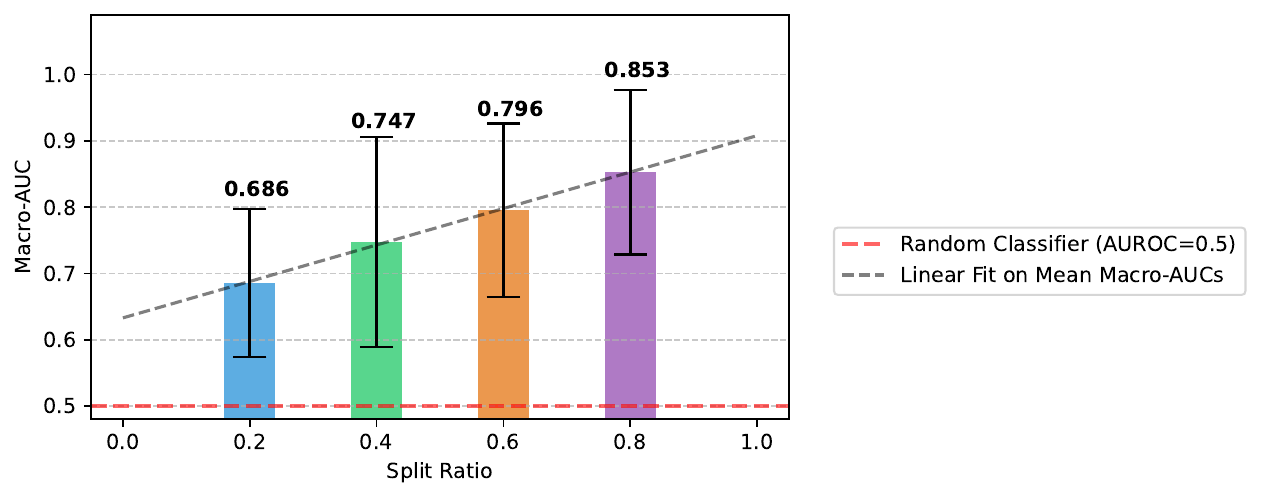}
    \caption{Impact of train-test split ratio on model performance on the PU dataset using WDCNN with envelope spectrum as the input representation. \new{In this figure, the N:M split ratio denotes N as the minimum number of bearings per class allocated to the training set, while M denotes the fixed number of bearings per class in the test set.}}
    \label{fig:pu-diversity}
\end{figure}

As depicted in Figure \ref{fig:pu-diversity}, a clear and monotonic increase in the mean Macro AUROC was observed as the proportion of bearings in the training set increased.
Although the overall performance remains relatively low, this direct correlation provides strong evidence that the lack of component diversity is the principal bottleneck limiting model generalization on this dataset.

\subsection{CWRU dataset}

The results obtained on the CWRU dataset were comparable to those observed on the PU dataset for deep learning approaches, with the envelope spectrum domain yielding the best overall performance. In contrast, shallow learning models---particularly the Random Forest---achieved better results, as shown in Table \ref{tab:cwru-results}.

\begin{table}
\centering
\caption{Macro AUROC (mean ± standard deviation) results obtained applying our methodology in the CWRU dataset using Deep and Shallow Learning Models. }
\begin{tabular}{lll}
\toprule
\textbf{Model}         & \textbf{Input Repr. / Features} & \textbf{Macro AUROC}        \\
\midrule \multirow{3}{*}{WDCNN} & Time                          & 63.22\% ± 10.24\%         \\
                       & Frequency                     & 70.90\% ± 8.95\% \\
                       & Envelope Spectrum                     & 74.51\% ± 9.43\%  \\
\midrule \textbf{Random Forest} & \textbf{Time + \new{Envelope Spectrum}}     & \textbf{85.06\% ± 8.92\%} \\
SVM                    & Time + \new{Envelope Spectrum}              & \new{81.06\% ± 6.76\%}     \\ \bottomrule   
\end{tabular}
\label{tab:cwru-results}
\end{table}

We hypothesize that the limited performance of deep learning models may stem from their tendency to memorize specific bearing signatures seen during training, rather than generalizing to unseen signals. This effect is likely amplified in datasets with limited variability, such as CWRU. We speculate that in more diverse datasets, the deep models’ feature extractors (backbones) would be exposed to a broader range of examples, enabling the learning of more general and predictive representations. In contrast, for the CWRU dataset, the use of handcrafted features---both specialized and general---proves to be a more effective strategy. 

\begin{figure}
    \centering
    \includegraphics[width=0.8\linewidth]{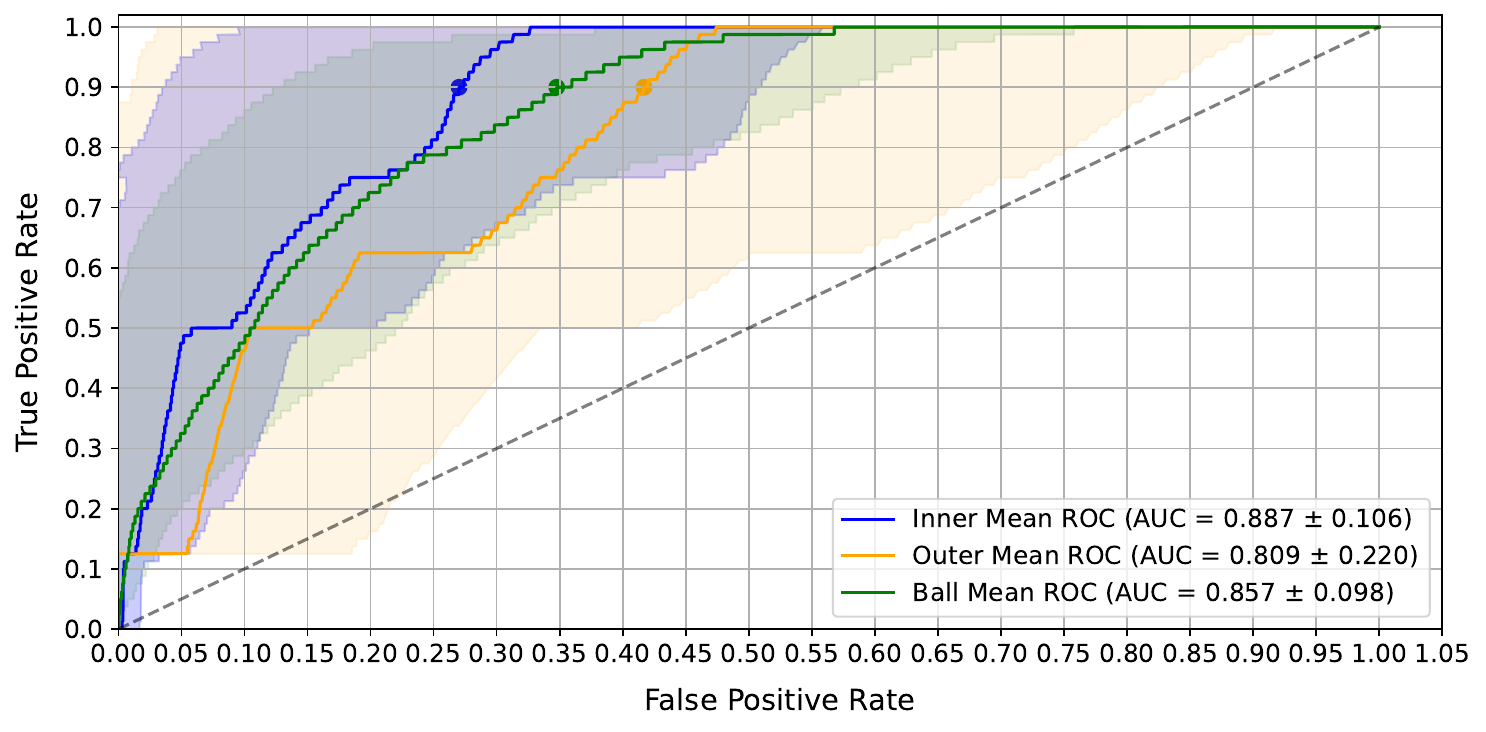}
    
    \caption{ROC curves on the CWRU dataset for the Random Forest trained with handcrafted features. The solid curves represent the horizontal average across 100 runs, while, for each curve, the shaded region represents one standard deviation.
}
    \label{fig:cwru-rocs}
\end{figure}

The ROC Curves calculated for the CWRU dataset are presented in Figure \ref{fig:cwru-rocs}. We observe a similar pattern to the UORED-VAFCLS and PU datasets, where the inner classifier outperforms the others. In this case, we reach FPR = 27.02, 41.69\%, and 34.76\% for the inner race, outer race and ball classifiers, respectively, when fixing  $\text{TPR} = 90\%.$

\begin{figure}
    \centering
    \includegraphics[width=0.8\linewidth]{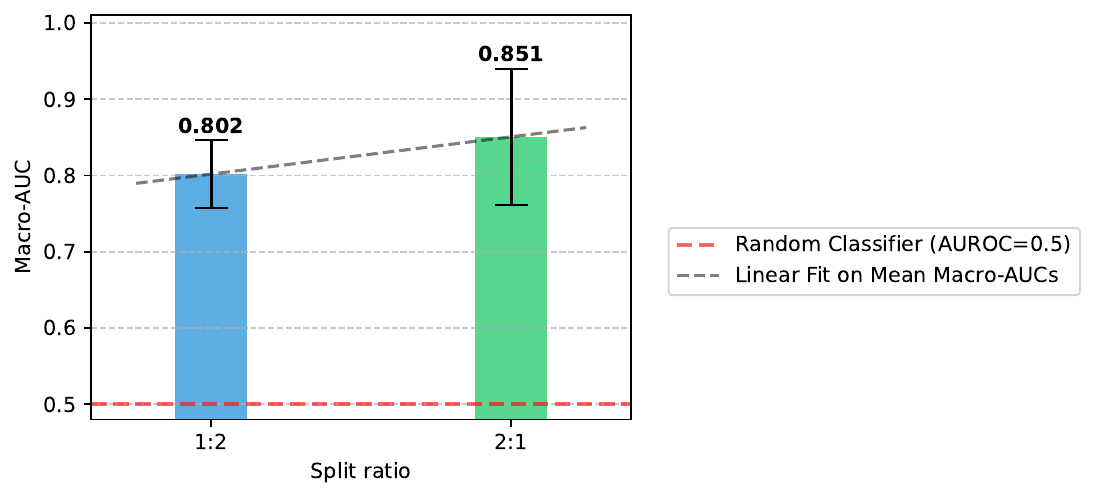}
    \caption{Impact of train-test split ratio on model performance on the CWRU dataset using Random Forest with hand-crafted features.}
    \label{fig:cwru-diversity}
\end{figure}

Finally, we compare two distinct train-test splits on CWRU (1:2 and 2:1), following the same procedure as before, seeking to observe the impact of different split ratios on model performance. For this experiment, in order to guarantee an equal number of samples in both splits, we applied segmentation with a $12000$ size and an overlap of $53\%$, yielding $20$ segments per signal in the 1:2 split. Figure~\ref{fig:cwru-diversity} shows the results of this experiment which, although limited due to available number of configurations, is in line with those obtained for PU and UORED-VAFCLS, suggesting that increased diversity can be beneficial to performance.

\section{Experiments with data leakage}
\label{sec:experiments_leakage}
In this section, we investigate the effects of data leakage across multiple datasets by designing experiments that fit their unique structures. Earlier, in Section \ref{sec:toyexample}, we showed using a toy problem that data leakage produces overoptimistic results, which are not suitable for decision-making. We now present similar findings using real data.

Although previous studies have investigated data leakage in bearing diagnosis, to the best of our knowledge, no prior work has done so solely by altering the test set. In our proposed experiments, we address this by keeping the training set---and thus the model---fixed, while modifying only the test set. This is important since changing both sets introduces a confounding factor: it becomes impossible to know whether a given result is inferior because leakage was avoided or because training diversity was reduced. We eliminate this confounding factor with our setup, in which the presence or absence of data leakage is determined entirely by the test set composition. This approach also explains why our no-leakage results differ from those presented in Section \ref{sec:experiments}, as we designed specific experiments for each dataset according to its inherent limitations. \new{We emphasize that the results presented in this section are not directly comparable to those in Section \ref{sec:experiments}, as both the training and test sets differ between these experimental settings.}

Our experimental design for all datasets contrasts a leakage-free evaluation protocol with common but flawed partitioning methods that introduces data leakage. To simulate a baseline scenario in which no hyperparameters are optimized (e.g., learning rate, batch size, or the use of data augmentation), we adopted a fixed training setup for all datasets and split configurations, using a learning rate of $10^{-4}$, a batch size of 16, no normalization, overlapping training segments, and a fixed training length of 30 epochs. Because in our designs each dataset contains a very different number of training signals---approximately 24 for UORED-VAFCLS, 700 for PU, and only 3 for CWRU---we set overlap ratios of 75\%, 0\%, and 95\%, respectively, to guarantee a sufficient number of training steps. Overall, this design aims to remove the influence of model selection as much as possible, so that any observed performance gains can be attributed exclusively to the proposed evaluation protocol, even under a fixed and non-optimized training configuration. \new{Finally, to ensure the reliability of our experimental findings, we conducted one-tailed paired t-tests and report the corresponding p-values under the null hypothesis that the leakage-based method does not outperform the leakage-free evaluation protocol.}

\subsection{UORED-VAFCLS dataset}

\label{sec:ott-leakage}

The experimental protocol for the UORED-VAFCLS dataset was structured around \new{20} distinct train-test splits, each following our standard partitioning strategy presented in Section \ref{subsec:uored-methodology}. In every split, the training set remained fixed across all test scenarios to ensure a controlled comparison. Specifically, training was performed using only the first 80\% of each signal from the training bearings. Additionally, to simulate a realistic data leakage scenario, all ``severe fault" signals from train bearings were removed from the original train set and set aside for use in one of the leakage scenarios.

Based on this fixed training set, we evaluated two distinct leakage scenarios:
\begin{itemize}
\item \textbf{Segmentation-level leakage:} This scenario examines leakage within individual signals. While the training set only included the first 80\% of each signal from the training bearings, the test set in this condition was constructed exclusively from the remaining 20\% of those same signals---thus exposing the model to future portions of time series it had already partially seen during training.\footnote{The most severe scenario occurs when signal segments are randomly shuffled between training and test sets, allowing the model to observe samples from the beginning, middle, and end of the same signal during evaluation, which is also the most common form of segmentation-level leakage in the literature. Experiments that evaluate this exact configuration are available in papers such as \cite{wheat_impact_2024, hendriks2022towards, matania-leakage, abburi2023closer}, which obtain perfect or near-perfect results.}
\item \textbf{Bearing-level leakage}: In this scenario, there is no segmentation-level leakage, but the test set includes signals from bearings that also appear in training. To create this, the previously removed ``severe fault'' segments from train bearings were reintroduced into the test set.  To maintain the test set size and class balance, an equal number of severe fault test samples were removed and replaced by these leakage samples.
\end{itemize}

\begin{table}
\centering
\caption{Performance comparison on the UORED-VAFCLS dataset, demonstrating the impact of bearing-level and segmentation-level data leakage versus the proposed leakage-free methodology. \new{Reported p-values correspond to one-tailed paired t-tests against the leakage-free setting.}}
\begin{tabular}{ccccc}
\toprule
\textbf{Split} & \textbf{Model} & \textbf{Input Repr. / Features} & \textbf{Macro AUROC} & \new{\textbf{p-value}} \\ \midrule

No leakage & WDCNN &
\begin{tabular}[c]{@{}c@{}}
Time\\
Frequency\\
Envelope
\end{tabular} &
\begin{tabular}[c]{@{}c@{}}
\new{87.25\% ± 6.03\%}\\
\new{90.21\% ± 5.57\%}\\
\new{87.34\% ± 7.71\%}
\end{tabular} &
\begin{tabular}[c]{@{}c@{}}
\new{-}\\
\new{-}\\
\new{-}
\end{tabular} \\ \midrule

Bearing-level leakage & WDCNN &
\begin{tabular}[c]{@{}c@{}}
Time\\
Frequency\\
Envelope
\end{tabular} &
\begin{tabular}[c]{@{}c@{}}
\new{90.23\% ± 4.02\%}\\
\new{94.15\% ± 3.15\%}\\
\new{93.79\% ± 4.42\%}
\end{tabular} &
\begin{tabular}[c]{@{}c@{}}
\new{$3.20 \times 10^{-4}$}\\
\new{$9.50 \times 10^{-4}$}\\
\new{$3.37 \times 10^{-6}$}
\end{tabular} \\ \midrule

Segmentation-level leakage & WDCNN &
\begin{tabular}[c]{@{}c@{}}
Time\\
Frequency\\
Envelope
\end{tabular} &
\begin{tabular}[c]{@{}c@{}}
\new{99.88\% ± 0.39\%}\\
\new{100.00\% ± 0.00\%}\\
\new{99.88\% ± 0.39\%}
\end{tabular} &
\begin{tabular}[c]{@{}c@{}}
\new{$1.33 \times 10^{-8}$}\\
\new{$1.58 \times 10^{-7}$}\\
\new{$5.12 \times 10^{-7}$}
\end{tabular} \\ \bottomrule

\end{tabular}
\label{tab:ou-leakage}
\end{table}

The results reported in Table  \ref{tab:ou-leakage} correspond to the mean Macro AUROC across the \new{20} train-test splits for each split configuration and highlight the substantial and misleading performance inflation caused by both forms of data leakage. Introducing the bearing-level leakage resulted in an artificial performance increase of approximately \new{3.0\%, 3.9\% and 6.4\%} for time-domain, frequency-domain, and envelope spectrum representations, respectively. The more severe segmentation-level leakage produced a higher performance increase, elevating the Macro AUROC by \new{12.6\%, 9.8\% and 12.5\%} across the same representations. Notably, we observed that the performance gap between representations progressively decreased as bearing- and segmentation-level leakage were introduced, specially on the latter case, where their performance was almost equal. This result shows that invalid partitioning can distort not only performance estimates but also conclusions about the most suitable input representation during model development. Importantly, strong results were obtained even without tuning any hyperparameters, simply by introducing data leakage, \new{and the obtained p-values confirm that leakage leads to a statistically significant overestimation of performance compared to the leakage-free setting}. These findings provide compelling empirical evidence for the necessity of strict, bearing-wise data partitioning to ensure the validity and reliability of model evaluation in bearing fault diagnosis.

\subsection{PU dataset}
\label{sec:dataleakage-pu}

For the experimental evaluation on the PU dataset, we also generated \new{20} distinct data splits. In each split, the training set was constructed utilizing data from three of the four available operating conditions. From these selected conditions, 15 measurement repetitions were randomly allocated for model training, while the residual 5 repetitions, along with all data from the fourth operating condition, were reserved to be used in the leakage scenarios. As a key modification, we truncated all training signals by removing the final 25\% of each recording. Using this setup, we proceeded to test our model against three types of data leakage. First, to induce segmentation-level leakage, we followed a procedure similar to that used for the UORED-VAFCLS dataset, but now placing the 25\% segments taken out of the training set into the test set (instead of 20\%). The remaining types of data leakage are detailed below: 

\begin{itemize}

    \item \textbf{Bearing-level leakage (Condition-wise split):} In this scenario, the model is exposed to the same physical bearing in both the training and testing sets, but under different operating conditions. Each training set has three of the four available conditions, while the remaining one is added to the test set, introducing leakage.
    \item \textbf{Bearing-level leakage (Repetition-wise split):} This represents a more severe leakage case. Here, the model is trained on one measurement repetition and tested on a different repetition from the \textit{exact same bearing and operating condition}. We simulate this type of leakage by reserving 15 of the 20 measurement repetitions for model training, while adding the rest to the test set.

\end{itemize}

\begin{table}
\centering
\caption{Performance comparison on the PU dataset, demonstrating the impact of bearing-level and segmentation-level data leakage versus the proposed leakage-free methodology. \new{Reported p-values correspond to one-tailed paired t-tests against the leakage-free setting.}}
\begin{tabular}{ccccc}
\toprule
\textbf{Split} & \textbf{Model} & \textbf{Input Repr. / Features} & \textbf{Macro AUROC}  & \new{\textbf{p-value}} \\ \midrule

No leakage & WDCNN &
\begin{tabular}[c]{@{}c@{}}
Time\\
Frequency\\
Envelope
\end{tabular} &
\begin{tabular}[c]{@{}c@{}}
\new{54.25\% ± 17.79\%}\\
\new{61.25\% ± 17.96\%}\\
\new{76.98\% ± 8.51\%}
\end{tabular}  &
\begin{tabular}[c]{@{}c@{}}
\new{-}\\
\new{-}\\
\new{-}
\end{tabular} \\ \midrule

Bearing-level leakage (Condition) & WDCNN &
\begin{tabular}[c]{@{}c@{}}
Time\\
Frequency\\
Envelope
\end{tabular} &
\begin{tabular}[c]{@{}c@{}}
\new{86.88\% ± 14.55\%}\\
\new{92.27\% ± 11.80\%}\\
\new{87.79\% ± 18.50\%}
\end{tabular}  &
\begin{tabular}[c]{@{}c@{}}
\new{$8.97 \times 10^{-6}$}\\
\new{$4.20 \times 10^{-6}$}\\
\new{$1.07 \times 10^{-2}$}
\end{tabular} \\ \midrule

Bearing-level leakage (Repetition) & WDCNN &
\begin{tabular}[c]{@{}c@{}}
Time\\
Frequency\\
Envelope
\end{tabular} &
\begin{tabular}[c]{@{}c@{}}
\new{98.57\% ± 2.15\%}\\
\new{100.00\% ± 0.00\%}\\
\new{99.99\% ± 0.01\%}
\end{tabular} &
\begin{tabular}[c]{@{}c@{}}
\new{$9.39 \times 10^{-10}$}\\
\new{$7.04 \times 10^{-9}$}\\
\new{$1.76 \times 10^{-10}$}
\end{tabular} \\ \midrule

Segmentation-level leakage & WDCNN &
\begin{tabular}[c]{@{}c@{}}
Time\\
Frequency\\
Envelope
\end{tabular} &
\begin{tabular}[c]{@{}c@{}}
\new{98.56\% ± 2.16\%}\\
\new{100.00\% ± 0.00\%}\\
\new{100.00\% ± 0.01\%}
\end{tabular}  &
\begin{tabular}[c]{@{}c@{}}
\new{$9.67 \times 10^{-10}$}\\
\new{$7.05 \times 10^{-9}$}\\
\new{$1.75 \times 10^{-10}$}
\end{tabular} \\ \bottomrule

\end{tabular}
\label{tab:pu-leakage}
\end{table}

The results in Table~\ref{tab:pu-leakage} reveal a pattern similar to that of the UORED-VAFCLS experiments, showing the impact of bearing-level leakage (with different partitioning methods) and segmentation-level leakage. Particularly, we observe a significant difference between leakage across condition- and repetition-wise splits. The results show that repetitions of the same experiment provide an easy way to memorize signal characteristics (a gain of \new{44.3\%} on time domain), achieving perfect performance with frequency and envelope spectrum inputs and near-perfect performance in the time domain, indicating that repetition-wise splitting is as detrimental as segmentation-level leakage in terms of performance inflation. Meanwhile, the condition-wise split also shows a high performance gain of \new{32.6\%} (on time domain) compared to the valid experiment, indicating that this type of data splitting also provides an easy way for model memorization. Consistent with the UORED-VAFCLS protocol, the reported results correspond to the mean Macro AUROC aggregated over all train-test splits in each configuration.

\subsection{CWRU dataset}

\begin{figure}
    \centering
    
    \includegraphics[width=0.64\linewidth, trim={0 0 15 23},
                     clip]{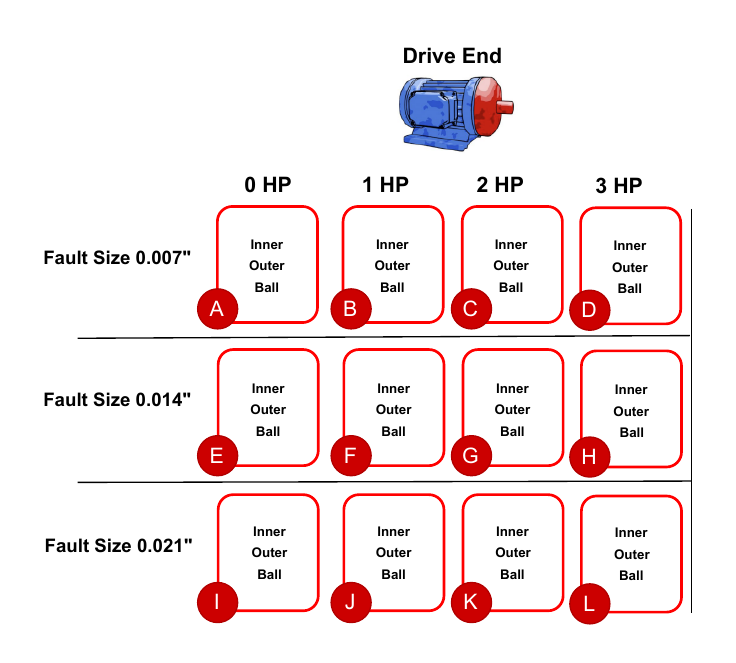}
    \caption{CWRU leakage experiment groups.}
    \label{fig:cwru-leakage}
\end{figure}

The experimental protocol for the CWRU dataset consists in training with samples grouped by condition and fault size, as represented in Figure \ref{fig:cwru-leakage}. In total, there are 12 groups (denoted by A--L), where each group corresponds to signals from faulty bearings located in the drive end with the same load and fault size, yielding 3 signals per group. Note that no signals from healthy bearings are included, due to the difficulty of avoiding data leakage when they are used.  With this setup, every experiment was performed by training with a single group and testing on some of the others. In our leakage-free protocol, each model was evaluated in all of the groups with a different fault size, e.g., a model trained on group A was tested on groups E, F, G, H, I, J, K, and L. With bearing-level leakage, the test set for each group consisted of the three remaining groups with the same fault size, e.g., a model trained on group A was tested on groups B, C and D. Finally, under segmentation-level leakage, we followed the same procedure adopted for the UORED-VAFCLS dataset, in which, for each signal, the first 80\% of its duration was used for training and the remaining 20\% was reserved for testing (note that, in this case, a model is trained and evaluated on the same group). 

The results are presented in Table \ref{tab:cwru-leakage} and follow the same pattern as the UORED-VAFCLS and PU datasets. In order to summarize the performance across all of the groups, the mean Macro AUROC was calculated over the test results in each split configuration.  In this case, both scenarios of bearing-level leakage (with condition-wise splits) and segmentation-level leakage resulted in perfect or near-perfect scores with time, frequency and envelope inputs. Notably, a model trained using only a single signal from a single bearing for each fault type is able to achieve perfect performance when evaluated on other signals from those same bearings under different operating conditions. This reinforces our hypothesis that condition-wise splits induces a strong data leakage, very similar to segmentation-level leakage on the CWRU dataset.

\begin{table}
\centering
\caption{Performance comparison on the CWRU dataset, demonstrating the impact of bearing-level and segmentation-level data leakage versus the proposed leakage-free methodology. \new{Reported p-values correspond to one-tailed paired t-tests against the leakage-free setting.}}
\begin{tabular}{ccccc}
\toprule
\textbf{Split} & \textbf{Model} & \textbf{Input Repr. / Features} & \textbf{Macro AUROC} & \new{\textbf{p-value}} \\ \midrule

No leakage & WDCNN &
\begin{tabular}[c]{@{}c@{}}
Time\\
Frequency\\
Envelope
\end{tabular} &
\begin{tabular}[c]{@{}c@{}}
63.17\% ± 10.84\%\\
64.34\% ± 9.29\%\\
63.89\% ± 9.26\%
\end{tabular}  &
\begin{tabular}[c]{@{}c@{}}
\new{-}\\
\new{-}\\
\new{-}
\end{tabular} \\ \midrule

Bearing-level leakage (Condition) & WDCNN &
\begin{tabular}[c]{@{}c@{}}
Time\\
Frequency\\
Envelope
\end{tabular} &
\begin{tabular}[c]{@{}c@{}}
100.00\% ± 0.00\%\\
99.96\% ± 0.13\%\\
99.79\% ± 0.54\%
\end{tabular}  &
\begin{tabular}[c]{@{}c@{}}
\new{$7.11 \times 10^{-8}$}\\
\new{$1.93 \times 10^{-8}$}\\
\new{$1.88 \times 10^{-8}$}
\end{tabular} \\ \midrule

Segmentation-level leakage & WDCNN &
\begin{tabular}[c]{@{}c@{}}
Time\\
Frequency\\
Envelope
\end{tabular} &
\begin{tabular}[c]{@{}c@{}}
100.00\% ± 0.00\%\\
100.00\% ± 0.00\%\\
100.00\% ± 0.00\%
\end{tabular}  &
\begin{tabular}[c]{@{}c@{}}
\new{$7.11 \times 10^{-8}$}\\
\new{$2.02 \times 10^{-8}$}\\
\new{$1.70 \times 10^{-8}$}
\end{tabular} \\ \bottomrule

\end{tabular}
\label{tab:cwru-leakage}
\end{table}

While these results are clear indicators of data leakage, we now provide a deeper analysis on model failure, specifically showing how feature distributions on models without diversity are similar for signals from the same bearings. Our analysis uses the model trained on group A with time-domain inputs. This model achieves perfect AUROC for each classifier in the bearing-level leakage scenario, but shows more realistic performance with the valid (no-leakage) split: 94.8\%, 53.9\%, and 78.7\% for inner race, outer race, and ball faults, respectively. We selected a baseline decision threshold of $0.5$ for all classifiers. Next, we extracted the model weights for each classifier and kept the top 15 values, which correspond to the features that most strongly drive failure predictions. We then selected a random 1-second segment from each signal and extracted the corresponding features. Using this weight ranking, we organized the feature distributions for all bearings and visualized the classifier output for each signal together with its group and fault type.

\begin{figure}
    \centering
    
    \includegraphics[width=1.05\linewidth, 
                     clip]{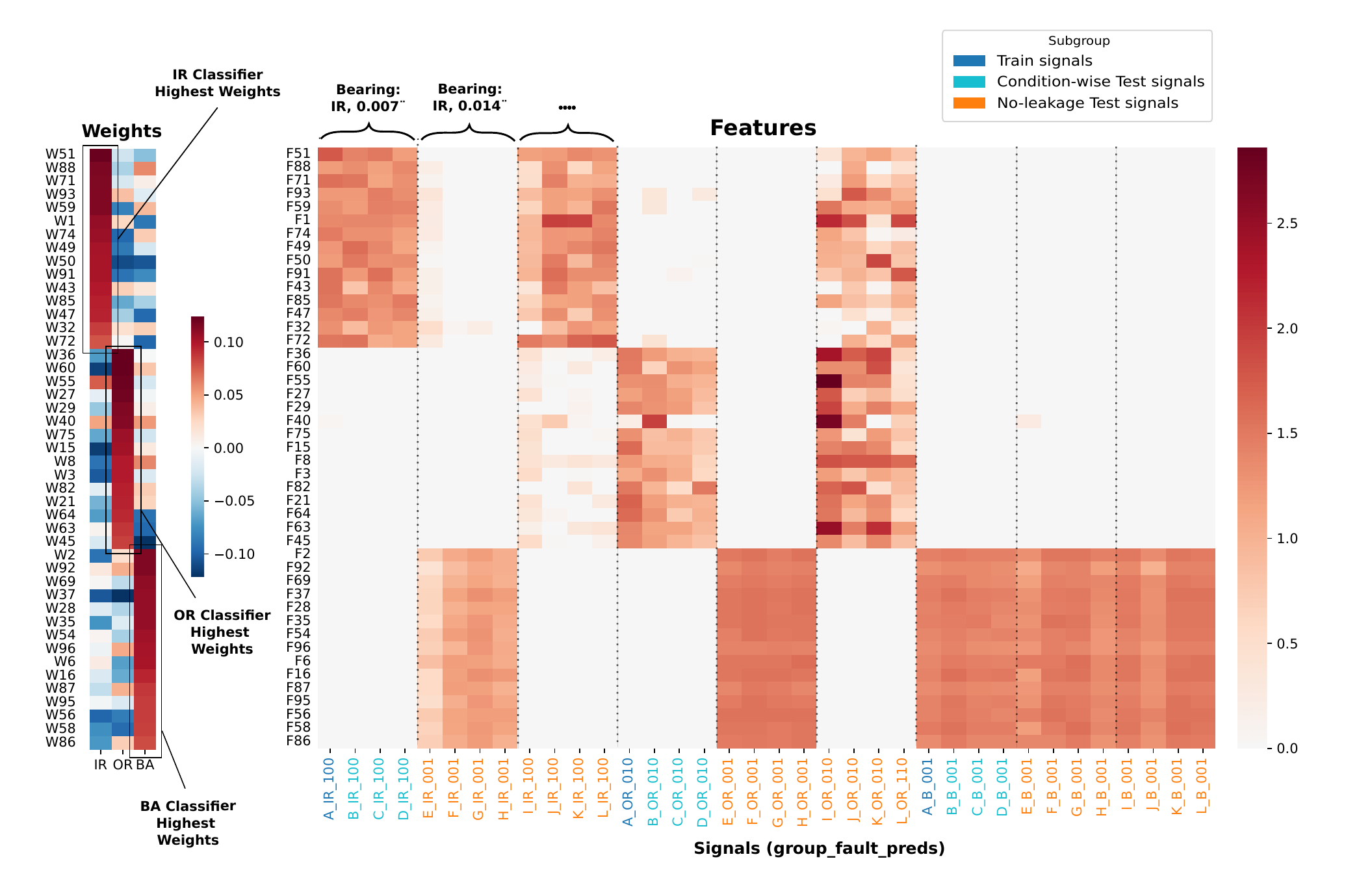}
    \caption{\new{Feature weight activation map for the WDCNN trained on Group A using time domain inputs from the CWRU dataset. The left panel displays the 45 most relevant classifier weights, obtained by selecting the 15 highest magnitude weights from each classification head (IR, OR, and BA). Rows labeled with \textbf{W} correspond to these classifier weights, grouped according to the classifier for which they are most relevant. The right panel display the corresponding feature activations extracted by the model backbone for the same feature indices, allowing direct comparison between classifier importance and feature responses across signals. Rows labeled with \textbf{F} correspond to feature indices associated with the selected weights. The columns represent signals labeled according to \textbf{Group\_FaultType\_Prediction}, where each column denotes a specific group (A--L), fault mode (IR, OR, or B), and predicted output in the same order (e.g., ``110'' indicates a prediction containing both IR and OR faults). Bearings are arranged into groups of four columns, with each column corresponding to a different load condition. For example, signals associated with the bearing presenting an IR fault of size 0.007'' are distributed across columns (A--D)\_IR, whereas signals associated with the bearing presenting an OR fault of size 0.021'' are distributed across columns (I--L)\_OR. This visualization highlights how signals originating from the same bearing exhibit highly similar feature activation patterns, revealing the tendency of the model to memorize bearing specific signal characteristics rather than learn generalizable fault related predictors.}}
    \label{fig:cwru-features}
\end{figure}

Figure \ref{fig:cwru-features} shows our results in detail and highlights several important points. First, we clearly observe that a model trained with a given bearing produces similar feature distributions for other signals from that same bearing. In this experiment, groups that share the same bearing between training and testing but differ in operating conditions---such as B, C, and D---form blocks of similar features. These feature blocks are also associated with the highest weights of each classifier, indicating that the model learned to prioritize them when detecting faults. However, other bearings (such as the inner and outer race in groups E to H) show inconsistent values for these same features, suggesting that they are not true fault indicators but rather spurious correlations with the bearings seen during training. For ball faults, feature distributions appear more consistent across bearings, since most ball fault signals activate similar high-importance features. Even so, these same features are also triggered by bearings with other fault types in groups E to H. Overall, these results suggest that a model trained on only a few bearings tends to memorize intrinsic signal characteristics instead of learning fault-related indicators. Evaluating such models on signals from the same bearings is therefore close to measuring training performance rather than true generalization.

\new{\section{Discussion}}

\new{Our experimental results show that, under a leakage-free methodology, the deep learning models we were able to evaluate within our computational budget and development time did not always achieve the best performance. Specifically, the results presented in Section~\ref{sec:experiments} show that, on the CWRU dataset, a Random Forest trained on handcrafted features outperformed the evaluated deep learning baselines, achieving a Macro AUROC of 85.06\% ± 8.92\%, compared to 74.51\% ± 9.43\% obtained by the WDCNN using envelope spectrum inputs. Similarly, results on the HUST bearing dataset (Appendix~\ref{app:c}) also indicate improved performance with the Random Forest, which achieved a Macro AUROC of 80.50\% ± 5.52\%, compared to 76.86\% ± 5.27\% obtained by the WDCNN using time-domain inputs. These results contrast with many papers that use exclusively deep learning, as we observe cases where shallow learning outperforms it. Such findings reinforce the importance of also considering non-deep learning techniques as baselines, as they often have less capacity and computational cost.}

 \new{We hypothesize that the widespread adoption of deep learning in this field is partly driven by the strong performance these methods often report, particularly in experimental settings where flaws in data-splitting introduce data leakage. Since higher-capacity models are inherently more prone to memorizing training patterns, their performance can be artificially inflated when evaluated under such conditions. This behavior is illustrated in our Toy Problem (Section~\ref{sec:toyexample}) experiments, where the higher-capacity model (LR) consistently outperformed its lower-capacity counterpart (DT) in leakage scenarios; however, when evaluated under leakage-free conditions, this apparent advantage was reversed, with the DT achieving superior performance. Taken together, these results suggest that part of the reported superiority of deep learning methods in prior works may reflect evaluation biases introduced by non-bearing-wise data-splitting methodologies, rather than genuine gains in generalization.}

\new{Another notable aspect observed from the results is the impact of input representations when using Deep Learning techniques. On the PU dataset, the WDCNN with envelope spectrum input yielded the best performance (79.56\% ± 13.07\%),  while on the UORED-VAFCLS dataset, WDCNN with frequency-domain input outperformed all other configurations, reaching a Macro AUROC of 93.12\% ± 4.26\%. This highlights the importance of choosing a good input representation for the dataset at hand.}

\new{Our results also underscore the critical role of bearing diversity in the training set, as increasing the number of distinct bearings consistently led to improved performance across all evaluated datasets.
 Importantly, our experimental protocol demonstrates that simply enlarging the training set is not sufficient when this increase does not introduce new bearings, since all split configurations (1:4, 2:3, 3:2 and 4:1) were trained with the same number of samples (the total number of samples seen by the model during the entire training is kept fixed) and yet exhibited substantially different performance depending on the number of bearings available for training. This behavior indicates that bearing diversity, rather than sample count alone, plays a critical role in achieving generalizable performance.}

\new{Finally, our findings further emphasize the difficulty of avoiding data leakage in the CWRU dataset. Despite being the most widely used benchmark in bearing fault diagnosis, CWRU presents significant challenges for rigorous evaluation due to its structure. In contrast, the PU, UORED-VAFCLS and HUST bearing datasets are more amenable to clean bearing-wise splits, making them more suitable for assessing true model generalization.}

\section{Deployment considerations}
\label{sec:deploy}

Considering the importance of a realistic, leakage-free evaluation highlighted by the previous experiments, we encourage researchers to apply our methodology to their own datasets. In many industrial scenarios, implementing a bearing-wise split may not be straightforward, especially when bearing identifiers are not explicitly available and data acquisition is typically organized by sensor location rather than by individual bearings. A common setup involves multiple sensors mounted at different points on a rotating machine (e.g., a conveyor belt), with each sensor positioned close to a bearing. In such cases, a sensor-based split, in which models are evaluated on data from unseen sensors, would be comparable to a bearing-wise split.

Although this approach adapts the bearing-wise split to realistic industrial settings, it may still be susceptible to data leakage, particularly because sensors can capture vibrations originating from nearby bearings (e.g., bearings located in side positions). When possible, we therefore recommend more robust strategies to promote proper generalization, such as splitting data by groups of nearby sensors (which are attached to individual bearings or other components) or by entire machines when multiple machines are available. This requirement extends naturally to online learning scenarios, where model updates must be performed without using data from the evaluation group in order to avoid introducing data leakage.

With these progressively stricter data partitioning strategies, measured model performance can more realistically reflect deployment conditions, e.g., if a model is intended for operation on unseen machines, a machine-wise split would be the appropriate choice. Nevertheless, machine-wise splitting can be challenging due to domain shift, which may require the use of domain adaptation or domain generalization techniques. A similar issue arises when models are trained on public benchmark datasets and subsequently tested on data from different machines, a scenario that likewise characterizes domain shift comparable to machine-wise partitioning.

Finally, although this work focuses on bearing fault diagnosis, our multilabel framework can be extended to include multiple fault modes (e.g., lack of lubrication), allowing a specific operating point to be defined for each classifier and enabling an evaluation methodology that is robust to class prevalence, which is particularly useful in condition monitoring applications.

\new{
\section{Conclusions}
\label{sec:conclusions}

This paper demonstrated the importance of realistic evaluation protocols in bearing fault diagnosis, showing that improper data splitting methodologies introduce data leakage and consequently lead to overestimated and unreliable results. By employing bearing-wise splits across four public datasets (CWRU, PU, UORED-VAFCLS and HUST bearing), we observed substantial variations in model performance depending on dataset diversity, input representation, and learning approach.

Beyond data splitting, we proposed a multi-label framework that offers greater data efficiency than the multiclass formulation, enables the diagnosis of co-occurring faults even when such are absent during training, and supports the use of fine-grained, precision-independent evaluation metrics with explicit control over the operating point, such as ROC curves and Macro AUROC.

Our results further highlighted the critical role of bearing diversity during training, showing that each dataset may require a different number of training bearings to achieve strong generalization performance. Since most public datasets contain only a limited number of bearings, evaluating techniques across multiple datasets is essential for obtaining realistic estimates of their effectiveness. Finally, we observed in two of the four evaluated datasets (CWRU and HUST bearing) that shallow learning techniques outperformed the deep learning models considered in this work, reinforcing the importance of including such approaches as baselines, given their significantly lower computational cost.
}

Based on these insights, we provide the following recommendations to researchers in the field:

\begin{itemize}
    \item \textbf{Bearing-wise split:} Ensure strict separation of bearings across training and test splits to avoid misleading, inflated performance estimates due to data leakage. 
    \item \textbf{Datasets with different properties:} We encourage the community to systematically test models on datasets with varying properties (e.g., PU, UORED-VAFCLS) to better assess robustness, as well as to explore multiple diversity configurations within each dataset. It is also crucial to select datasets with a structure that supports proper bearing-wise splitting and contain a sufficient number of bearings per class to enable meaningful evaluation (which is not the case of the CWRU).
    \item \textbf{Model selection and evaluation protocol:} The process for tuning hyperparameters must be clearly detailed. These parameters must be chosen without using the evaluation performance.
    \item \textbf{Multi-label formulation:} This formulation provides a natural setup for diagnosing co-occurring faults and enables a more precise evaluation with prevalence-independent metrics, such as the Macro AUROC metric for model development/selection and the individual ROC curves for final evaluation and operating-point selection.
    \item  \textbf{Models:} We suggest that a comprehensive evaluation should not default to deep learning architectures. Shallow models often provide competitive or even superior results depending on the dataset, and we encourage their inclusion as robust baselines. %
    \item \textbf{Share code, data splits, and evaluation pipelines:} Reproducibility is key for advancing the field and ensuring fair comparisons between proposed methods.
    
\end{itemize}

By following these practices, the community can foster more reliable and generalizable machine learning systems for bearing fault diagnosis applications.

\printcredits

\section*{Acknowledgements}

This research was supported by the Conselho Nacional de Desenvolvimento Científico e Tecnológico (CNPq), grants 164299/2021-1 and 304619/2022-1, and the Coordenação de Aperfeiçoamento de Pessoal de Nível Superior (CAPES), grants 88887.951193/2024-00 and 88887.137796/2025-00.

\appendix
\section{Derivation of the maximum achievable accuracy in the toy example}
\label{app:a}

Consider a discrete random variable $y \in \{0,1\}$ representing the state of a given bearing, where $y=0$ denotes the healthy state and $y=1$ denotes the faulty state.

We observe a continuous random vector $X \in\mathbb{R}^N$ representing $N$ fault predictive features, whose componentes are i.i.d.\ Gaussian random variables conditioned on $y$, given by $X_i \mid y \sim \mathcal N(yA, 1)$, $i=1,2,..., N$.
More precisely,
\begin{equation}
P({X}\mid y=0) = \prod_{i=1}^{N}\sqrt{\frac{1}{2\pi}}e^{-\frac{{X_i}^2}{2}} \quad \text{and} \quad P({X}\mid y=1) = \prod_{i=1}^{N}\sqrt{\frac{1}{2\pi}}e^{-\frac{({X_i}-A)^2}{2}}.
\end{equation}

It is well-known that the maximum a posteriori (MAP) decision rule minimizes the error probability (and therefore maximizes accuracy). The MAP rule selects the hypothesis $\hat{y}$ for the bearing health state as $\hat{y} = \arg\max_{y} P(y\mid X)$. In this case, this amounts to deciding $\hat{y}=1$ whenever $P(y=1\mid {X}) > P(y=0\mid {X})$. It follows that
\begin{align}
\hat{y}=1 
&\iff P(y=1\mid {X}) > P(y=0\mid {X}) \\
&\iff \frac{P({X}\mid y=1)P(y=1)}{P({X})} > \frac{P({X}\mid y=0)P(y=0)}{P({X})} \\
&\iff \frac{P({X}\mid y=1)}{P({X}\mid y=0)} > \frac{P(y=0)}{P(y=1)} \\
&\iff e^{\sum_{i=1}^{N}\frac{-(X_i-A)^2}{2}+\frac{X_i^2}{2}} > \frac{P(y=0)}{P(y=1)} \\
&\iff \sum_{i=1}^{N}\frac{-(X_i-A)^2}{2}+\frac{X_i^2}{2} > \log \frac{P(y=0)}{P(y=1)} \\
&\iff A\sum_{i=1}^N X_i - \frac{NA^2}{2} > \log \frac{P(y=0)}{P(y=1)} \\
&\iff \frac{1}{N}\sum_{i=1}^N X_i > \frac{1}{NA}\log \frac{P(y=0)}{P(y=1)} + \frac{A}{2}.
\end{align}

Since we assume $P(y=0) = P(y=1) = \frac{1}{2}$, we have that, under the MAP rule, 
\begin{equation}
\hat{y} = 
\begin{cases}
0, & \text{if } \bar{X} \leq A/2, \\
1, & \text{if } \bar{X} > A/2,
\end{cases} \quad \text{where $\bar{X} = \frac{1}{N}\sum_{i=1}^N X_i$}.
\end{equation}

We now wish to find the corresponding accuracy $P(\hat{y} = y)$. Note that $\bar{X} \mid y \sim \mathcal N(yA, 1/N)$. Since
\begin{align}
P(\hat{y} = 0 \mid y=0) &= P(\bar{X} \le A/2 \mid y=0) = \Phi \left(\frac{A/2}{1/\sqrt{N}} \right) = \Phi\left(\frac{A\sqrt{N}}{2}\right) \\
P(\hat{y} = 1 \mid y=1) &= P(\bar{X} > A/2 \mid y=1) = 1 - \Phi \left(\frac{A/2-A}{1/\sqrt{N}} \right) = 1 - \Phi\left(\frac{-A\sqrt{N}}{2}\right) = \Phi\left(\frac{A\sqrt{N}}{2}\right)
\end{align}
where $\Phi$ denotes the standard Gaussian c.d.f., it follows that
\begin{equation}
P(\hat{y} = y) = P(\hat{y} = 0 \mid y=0)P(y=0) + P(\hat{y} = 1 \mid y=1)P(y=1) = \Phi\left(\frac{A\sqrt{N}}{2}\right).
\end{equation}
For the special case of $N = 3$ and $A = 1.5$, we have $P(\hat{y} = y) = \Phi\left(\frac{1.5\sqrt{3}}{2}\right) \approx 0.9030$.

\new{
\section{Detection of co-occurring faults}
\label{app:c}

As mentioned in Section \ref{sec:problem-formulation}, the multi-label formulation proposed in this paper possesses two main advantages: achieving higher data efficiency when modeling co-occurring faults, which applies to the model training; and enabling the use of fine-grained precision-independent evaluation metrics, which applies to the model evaluation. The second advantage speaks for itself, by the fact that we can plot a ROC curve for each fault type, as in Figures \ref{fig:ou-rocs}, \ref{fig:pu-rocs}, \ref{fig:cwru-rocs}; in contrast, a multi-class model would produce a single point in the ROC space for each fault type, or a just single number (harder to interpret and that could change under a different fault distribution) if evaluated using multi-class accuracy.

In this appendix, following a reviewer suggestion, we would like to illustrate the first advantage. For this goal we use a high-quality vibration dataset that contains a sufficient number of co-occurring faults, the HUST bearing dataset \cite{thuan_hust_2023}. This dataset is composed of acquisitions from 33 bearings, of which 5 are healthy, 14 contain single faults, and the remaining 14 exhibit co-occurring fault conditions, including combinations such as IR+OR, OR+B, and IR+B. The faults were artificially induced, and the signals were acquired from an electric motor operating at approximately 1440 RPM. Each acquisition lasts around 10 seconds and was originally sampled at 51.2 kHz. In our experiments, the signals were resampled to 42 kHz to align with the UORED-VAFCLS setup while slightly reducing segment size. Additionally, each bearing was evaluated under three load conditions (0 W, 200 W, and 400~W), yielding 3 signals per bearing. Table \ref{tab:hust-bearings} summarizes the dataset by listing the bearing IDs and their corresponding fault categories.

\begin{table}[]
\caption{Bearing-level structure of the HUST bearing dataset. }
\label{tab:hust-bearings}
\centering
\resizebox{0.7\textwidth}{!}{%
\begin{tabular}{ccccccc}
\hline
\textbf{Healthy} & \textbf{Inner Ring (IR)} & \textbf{Outer Ring (OR)} & \textbf{Ball (B)} & \textbf{IR + B} & \textbf{OR + B} & \textbf{IR + OR} \\ \hline
N4 & I4 & O4 & -  & -   & OB4 & IO4 \\
N5 & I5 & O5 & B5 & IB5 & OB5 & IO5 \\
N6 & I6 & O6 & B6 & IB6 & OB6 & IO6 \\
N7 & I7 & O7 & B7 & IB7 & OB7 & IO7 \\
N8 & I8 & O8 & B8 & IB8 & OB8 & IO8
\end{tabular}%
}

\end{table}

\subsection{Training with single and combined faults}

For this dataset, a strict 2:2 split was adopted, in which two bearings per class are randomly selected to form the test set, and an additional two bearings per class are selected to form the training set. This choice ensures that all classes have equal representation in both training and test sets, enabling a fairer comparison with a multiclass model trained with the exact same split configuration. The CVM-CV protocol was also employed, using 5 train-test splits for hyperparameter tuning and 100 splits for evaluation. The same hyperparameter search space displayed in Tables \ref{tab:hparam_space} and \ref{tab:shallowspace} were used for both deep and shallow learning models. The results, presented in Table \ref{tab:hust-results}, exhibit a pattern similar to that observed for the CWRU dataset (Table~\ref{tab:cwru-results}), where the Random Forest outperforms WDCNN across multiple input representations.

\begin{table}
\centering
\caption{Macro AUROC (mean ± standard deviation) results obtained by applying our methodology with a 2:2 train-test split on the HUST bearing dataset using Deep and Shallow Learning Models. }
\begin{tabular}{lll}
\toprule
\textbf{Model}         & \textbf{Input Repr. / Features} & \textbf{Macro AUROC}        \\
\midrule \multirow{3}{*}{WDCNN} & Time                          &  76.86\% ± 5.27\%         \\
                       & Frequency                     & 69.44\% ± 6.72\% \\
                       & Envelope Spectrum                     & 74.33\% ± 4.55\%  \\
\midrule \textbf{Random Forest} & \textbf{Time + Envelope Spectrum}     & \textbf{80.50\% ± 5.52\%} \\
SVM                    & Time + Envelope Spectrum            & 74.57\% ± 5.96\%     \\ \bottomrule   
\end{tabular}
\label{tab:hust-results}
\end{table}

Following this, a Random Forest with 7 classes (Normal, IR, OR, B, IR+B, OR+B and IR+OR) was trained under the same methodology, with multiclass accuracy as the optimization metric. To directly compare both frameworks under this setup, we converted the multiclass model predictions into multi-label outputs. Specifically, each multiclass prediction was decomposed according to the faults it contained. For instance, any prediction containing a specific fault type (e.g., inner-race), whether as an isolated fault or as part of a combined fault, was treated as a positive prediction for the corresponding binary classifier (e.g., the inner-race classifier), while predictions that did not include that fault type were treated as negative. The same procedure was applied to the remaining classifiers, yielding corresponding mean TPR and FPR values across 100 splits.

In sequence, ROC curves were calculated for the multi-label framework and compared directly with the points obtained from the multiclass models (Figure~\ref{fig:hust-rocs}). The results show that the multiclass framework consistently underperforms the multi-label approach on inner-race and ball classifiers, while reaches a similar performance on the outer-race classifier. For instance, when diagnosing ball faults, the multi-label model achieves a mean TPR more than 20\% higher than that of the multiclass model at the same FPR.

\begin{figure}
    \centering
    \includegraphics[width=0.8\linewidth]{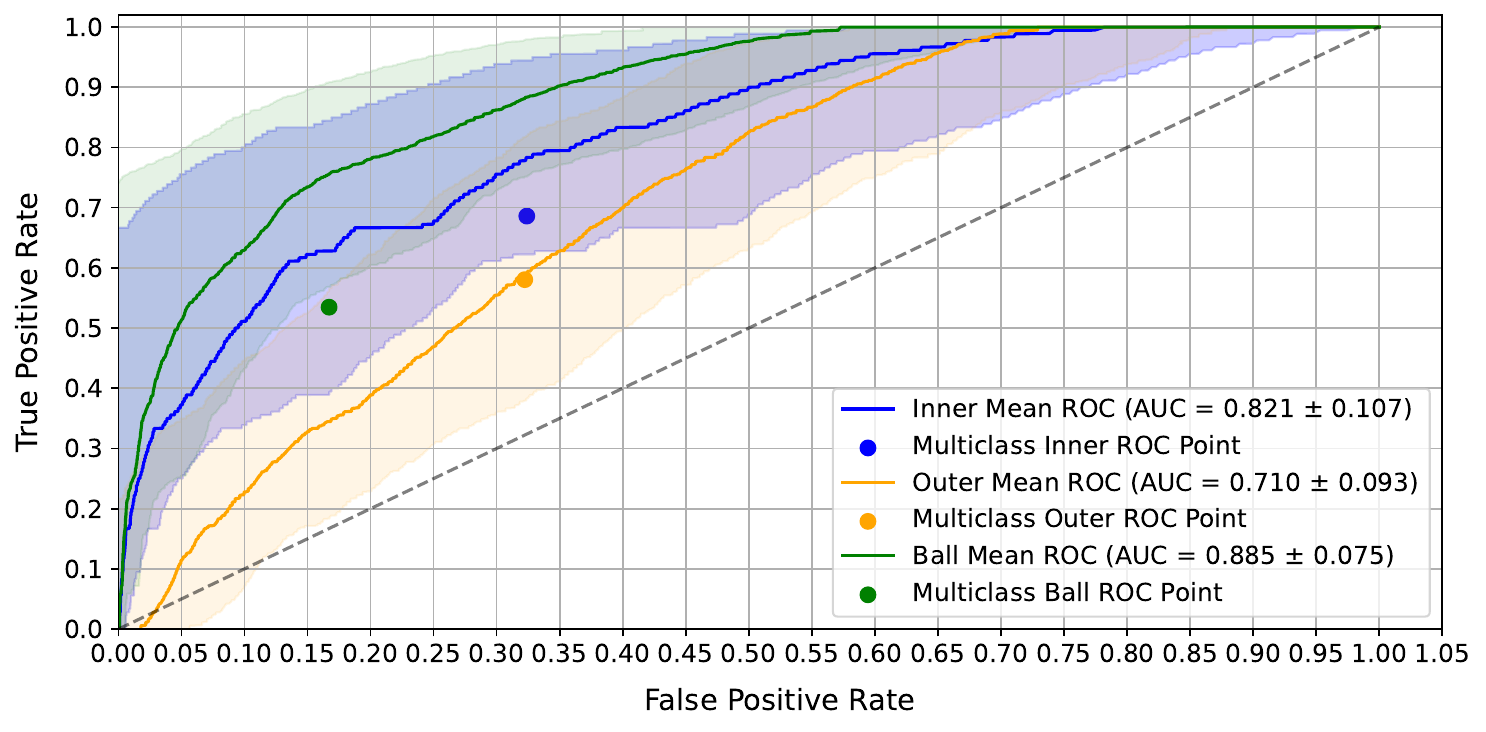}
    
    \caption{\new{ROC curves on the HUST bearing dataset for the Random Forest trained with handcrafted features on a 2:2 split. The individual points represent TPR x FPR mean values obtained from multiclass models when evaluated under a multi-label setup (independent binary fault classifiers).}
}
    \label{fig:hust-rocs}
\end{figure}

\subsection{Training exclusively with single-faults}

In the next experiment, we simulate a realistic scenario in which no combined faults are available in the training data. For that, we maintain the exact 2:2 split performed in the last experiment, but then eliminate all combined faults on training. This results in a 2:2 split for single-faults and a 0:2 split for combined faults, keeping the test set balanced. The same evaluation procedure was applied to the multiclass model (which now had only Normal, IR, OR and B classes) and to the multi-label model, yielding the ROC curves and points displayed in Figure \ref{fig:hust-rocs-nocomb}.

\begin{figure}
    \centering
    \includegraphics[width=0.8\linewidth]{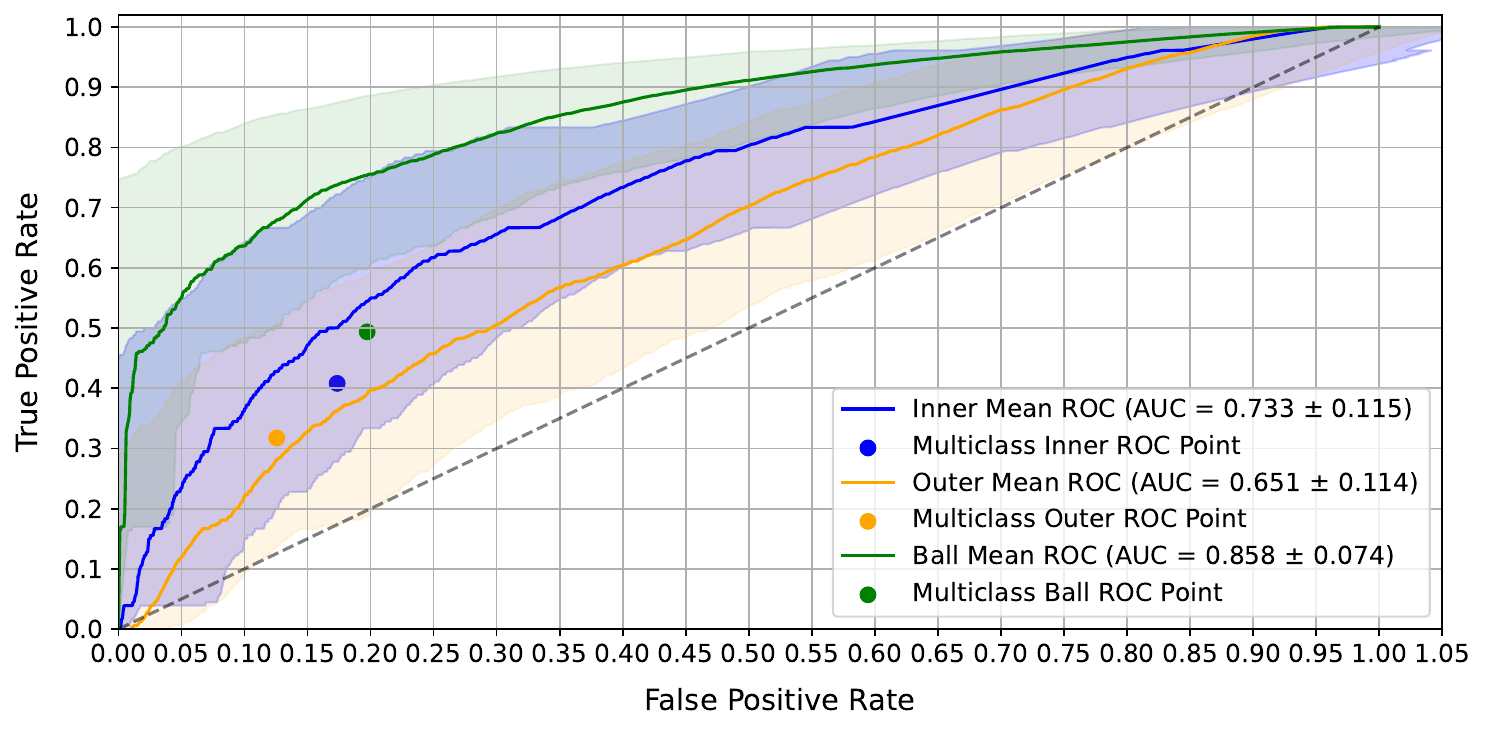}
    
    \caption{\new{ROC curves on the HUST bearing dataset using the (2:2, 0:2) split, where no co-occurring faults are present during training. The individual points represent TPR x FPR mean values obtained from multiclass models when evaluated under a multi-label setup (independent binary fault classifiers).}
}
    \label{fig:hust-rocs-nocomb}
\end{figure}

The results show significant performance decrease on the multiclass model, as it is no longer available to predict combined-fault classes. Therefore, any sample with a combined fault is diagnosed simply as healthy or a single-fault. The multi-label models also provided a decreased performance---but less impactful---in comparison to the 2:2 baseline (Table \ref{tab:hust-results}), reaching a mean Macro AUROC of 74.80\% ± 6.20\%, while still significantly outperforming the multiclass model on inner-race and ball diagnosis.

As the ROC curves provide an isolated analysis of each fault mode---effectively operating as marginal probabilities---they do not fully characterize the classifiers as a whole. Therefore, we further examine both frameworks by computing their corresponding confusion matrices, thereby evaluating their joint predictive distributions. To enable this analysis, we map the multi-label framework predictions to each possible class. Specifically, each prediction, originally composed of three outputs (one per classifier), is converted into a discrete multiclass label. To achieve this, we first select a threshold for each model that yields an FPR of approximately 16.55\%, corresponding to the mean FPR across the multiclass models for the inner, outer, and ball classifiers under the new (2:2, 0:2) split. The resulting binary predictions are then converted into multiclass labels, from which the confusion matrices are computed. Since this procedure is repeated across 100 train-test splits, we report the aggregated confusion matrices for each framework.

\begin{figure}

    \centering
    \includegraphics[width=1.0\linewidth]{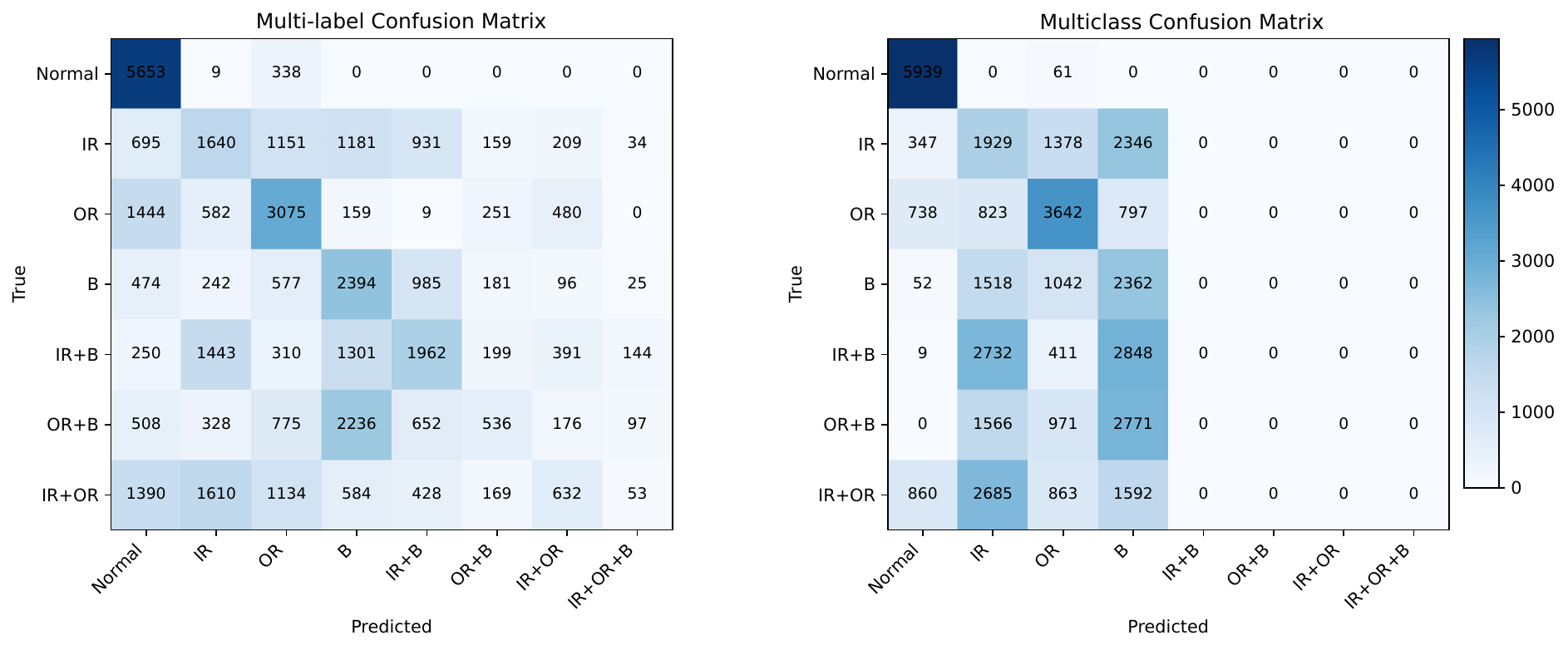}
    
    \caption{\new{Sum of confusion matrices obtained from models trained under the multi-label and multiclass frameworks on HUST bearing dataset. Each cell represents the sum of values across 100 train-test splits.}}

    \label{fig:hust-cm}

\end{figure}

The confusion matrices presented in Figure~\ref{fig:hust-cm} show that, by design, the multi-label framework is able to predict co-occurring faults even without containing them during training, which is a strong advantage on the proposed scenario. This is specially notable on the IR+B class, which has the best performance among the combined-fault classes and correspond to the individual classifiers with best performance among the multi-label evaluation. 
Overall, our results suggest that the proposed multi-label framework provides greater data efficiency when modeling co occurring faults, particularly in low diversity scenarios.

\subsection{Impact of bearing diversity}

Following a similar approach to the experiments presented in Section~\ref{sec:experiments}, we evaluate the impact of bearing diversity on the HUST bearing dataset. Three split ratios were considered: 1:3, 2:2, and 3:1. Figure~\ref{fig:hust-diversity} presents the obtained results, which reveal the same trend observed across all evaluated datasets: increasing the number of bearings in the training set contributes to improved model generalization.

\begin{figure}
    \centering
    \includegraphics[width=0.8\linewidth]{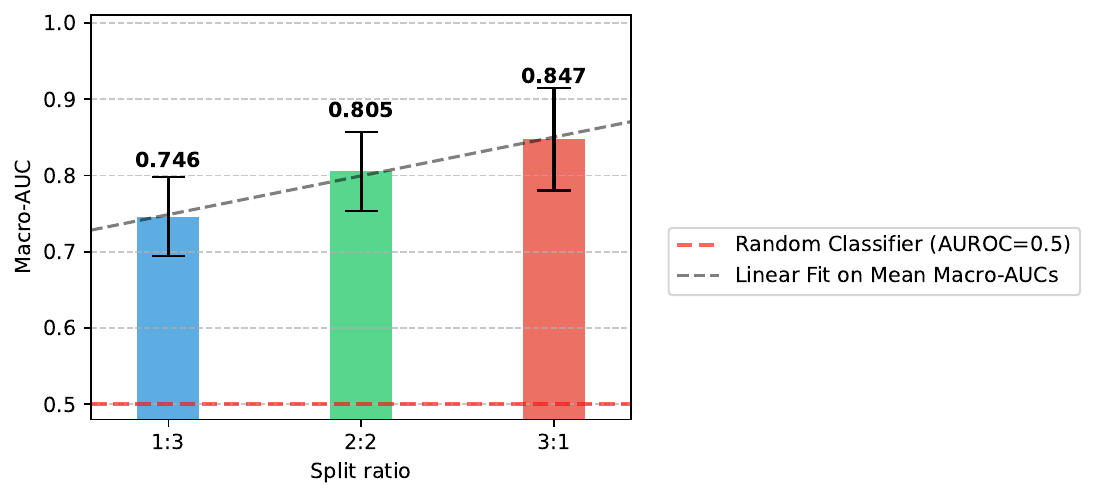}
    \caption{\new{Impact of train-test split ratio on model performance on the HUST bearing dataset using Random Forest with hand-crafted features.}}
    \label{fig:hust-diversity}
\end{figure}

}

\section{Details of hyperparameter optimization for all models}
\label{app:b}

In the shallow learning experiments, two models were considered: Random Forest with 200 estimators and SVM with RBF kernel. Since these models natively operate as either binary or multiclass classifiers, we employed the MultiOutputClassifier\footnote{Documentation for this wrapper is available in \href{https://scikit-learn.org/stable/modules/generated/sklearn.multioutput.MultiOutputClassifier.html}{https://scikit-learn.org/stable/modules/generated/sklearn.multioutput.MultiOutputClassifier.html}.} wrapper from scikit-learn to support the proposed multi-label framework by training one independent classifier per label, and then aggregating their results. Hyperparameter optimization was performed using RandomizedSearchCV from scikit-learn with 250 iterations, which samples random combinations of hyperparameters from predefined search spaces at each iteration. The corresponding hyperparameter distributions for both Random Forest and SVM models are reported in Table \ref{tab:shallowspace}.

\begin{table} 
\centering
\caption{Hyperparameter search spaces used for shallow learning models with RandomizedSearchCV.}
\begin{tabular}{lll}
\toprule
\textbf{Model} & \textbf{Hyperparameter} & \textbf{Search Space} \\
\midrule
\multirow{7}{*}{Random Forest} 
& max\_features & \{sqrt, log2\} \\
& criterion & \{gini, entropy, log\_loss\} \\
& max\_depth & randint(2, 60) \\
& min\_samples\_split & randint(2, 20) \\
& min\_samples\_leaf & randint(1, 20) \\
& ccp\_alpha & loguniform($10^{-5}$, 1) \\
\midrule
\multirow{3}{*}{SVM} 
& C & loguniform($10^{-3}$, $10^{3}$) \\
& gamma & \{scale, auto\} \\
& \new{Feature Scaling} & \new{\{none, StandardScaler\}} \\
\bottomrule
\end{tabular}
\label{tab:shallowspace}
\end{table}

Finally, in Table \ref{tab:best_hparams}, we provide the best found hyperparameters for all models and input representations.

\begin{table}[h!]
\centering
\caption{Best hyperparameter configurations obtained for all models and input representations across the UORED-VAFCLS, PU, CWRU \new{and HUST bearing} datasets.}
\label{tab:best_hparams}
\small
\begin{tabular}{lll m{6.4cm}}
\toprule
\textbf{Dataset} & \textbf{Model} & \textbf{Input Repr. / Features} & \textbf{Best Hyperparameters} \\
\midrule

\multirow{5}{*}{UORED-VAFCLS}
& Random Forest & Time + \new{Envelope Spectrum} &
 max\_features=log2, criterion=entropy, max\_depth=10, 
min\_samples\_split=14, min\_samples\_leaf=16, ccp\_alpha=$9.38\times10^{-5}$ \\

& SVM & Time + \new{Envelope Spectrum} &
$C=4.81\times10^{2}$, gamma=scale \\

& WDCNN & Time &
lr=$10^{-4}$, batch\_size=128, normalization=none \\

& WDCNN & Frequency &
lr=$10^{-3}$, batch\_size=256, normalization=none \\

& WDCNN & Envelope Spectrum &
lr=$10^{-4}$, batch\_size=16, normalization=global \\

\midrule
\multirow{5}{*}{PU}
& Random Forest & Time + \new{Envelope Spectrum} &
 max\_features=log2, criterion=log\_loss, max\_depth=24, 
min\_samples\_split=15, min\_samples\_leaf=18, ccp\_alpha=$2.65\times10^{-2}$ \\

& SVM & Time + \new{Envelope Spectrum} &
$C=4.41\times10^{2}$, gamma=scale \\

& WDCNN & Time &
lr=$10^{-2}$, batch\_size=32, normalization=none \\

& WDCNN & Frequency &
lr=$10^{-3}$, batch\_size=128, normalization=global \\

& WDCNN & Envelope Spectrum &
lr=$10^{-3}$, batch\_size=32, normalization=global \\

\midrule
\multirow{5}{*}{CWRU}
& Random Forest & Time + \new{Envelope Spectrum} &
max\_features=sqrt, criterion=log\_loss, max\_depth=47, 
min\_samples\_split=11, min\_samples\_leaf=8, ccp\_alpha=$3.20\times10^{-5}$ \\

& SVM & Time + \new{Envelope Spectrum} &
$C=1.84\times10^{2}$, gamma=scale \\

& WDCNN & Time &
lr=$10^{-4}$, batch\_size=32, normalization=none \\

& WDCNN & Frequency &
lr=$10^{-3}$, batch\_size=64, normalization=global \\

& WDCNN & Envelope Spectrum &
lr=$10^{-2}$, batch\_size=32, normalization=none \\

\midrule
\multirow{5}{*}{\new{HUST bearing}}
& \new{Random Forest} & \new{Time + Envelope Spectrum} &
\new{max\_features=log2, criterion=entropy, max\_depth=3, 
min\_samples\_split=7, min\_samples\_leaf=6, ccp\_alpha=$1.23\times10^{-2}$} \\

& \new{SVM} & \new{Time + Envelope Spectrum} &
\new{$C=1.84\times10^{-1}$, gamma=auto} \\

& \new{WDCNN} & \new{Time} &
\new{lr=$10^{-4}$, batch\_size=16, normalization=none} \\

& \new{WDCNN} & \new{Frequency} &
\new{lr=$10^{-5}$, batch\_size=16, normalization=none} \\

& \new{WDCNN} & \new{Envelope Spectrum} &
\new{lr=$10^{-2}$, batch\_size=128, normalization=none} \\

\bottomrule

\end{tabular}
\end{table}

\bibliographystyle{elsarticle-num}
\bibliography{refs}

@ARTICLE{hendriks2022towards,
  title={Towards better benchmarking using the {CWRU} bearing fault dataset},
  author={Hendriks, Jacob and Dumond, Patrick and Knox, DA},
  journal={Mechanical Systems and Signal Processing},
  volume={169},
  pages={108732},
  year={2022},
  publisher={Elsevier}
}

@ARTICLE{tsamardinos2015performance,
  title={Performance-estimation properties of cross-validation-based protocols with simultaneous hyper-parameter optimization},
  author={Tsamardinos, Ioannis and Rakhshani, Amin and Lagani, Vincenzo},
  journal={International Journal on Artificial Intelligence Tools},
  volume={24},
  number={05},
  pages={1540023},
  year={2015},
  publisher={World Scientific}
}

@article{smith2015rolling,
title = {Rolling element bearing diagnostics using the Case Western Reserve University data: A benchmark study},
journal = {Mechanical Systems and Signal Processing},
volume = {64-65},
pages = {100-131},
year = {2015},
issn = {0888-3270},
doi = {https://doi.org/10.1016/j.ymssp.2015.04.021},
url = {https://www.sciencedirect.com/science/article/pii/S0888327015002034},
author = {Wade A. Smith and Robert B. Randall}
}

@ARTICLE{lei2020applications,
  title={Applications of machine learning to machine fault diagnosis: A review and roadmap},
  author={Lei, Yaguo and Yang, Bin and Jiang, Xinwei and Jia, Feng and Li, Naipeng and Nandi, Asoke K},
  journal={Mechanical Systems and Signal Processing},
  volume={138},
  pages={106587},
  year={2020},
  publisher={Elsevier}
}

@ARTICLE{hogan2023on,
  title={On Averaging {ROC} Curves},
  author={Jack Hogan and Niall M. Adams},
  journal={Transactions on Machine Learning Research},
  issn={2835-8856},
  year={2023},
  url={https://openreview.net/forum?id=FByH3qL87G},
  note={Survey Certification}
}

@article{kapoorleakage2023,
title = {Leakage and the reproducibility crisis in machine-learning-based science},
journal = {Patterns},
volume = {4},
number = {9},
pages = {100804},
year = {2023},
issn = {2666-3899},
url = {https://www.sciencedirect.com/science/article/pii/S2666389923001599},
author = {Sayash Kapoor and Arvind Narayanan},
doi = {https://doi.org/10.1016/j.patter.2023.100804},
}

@inproceedings{abburi2023closer,
  title={A Closer Look at Bearing Fault Classification Approaches},
  author={Abburi, Harika and Chaudhary, Tanya and Ilyas, Sardar Haider Waseem and Manne, Lakshmi and Mittal, Deepak and Williams, Don and Snaidauf, Derek and Bowen, Edward and Veeramani, Balaji},
  booktitle={Annual Conference of the PHM Society},
  volume={15},
  number={1},
  year={2023}
}

@inproceedings{lessmeier2016condition,
  title={Condition monitoring of bearing damage in electromechanical drive systems by using motor current signals of electric motors: A benchmark data set for data-driven classification},
  author={Lessmeier, Christian and Kimotho, James Kuria and Zimmer, Detmar and Sextro, Walter},
  booktitle={PHM society European conference},
  volume={3},
  number={1},
  year={2016}
}

@article{wheat_impact_2024,
	title = {Impact of {Data} {Leakage} in {Vibration} {Signals} {Used} for {Bearing} {Fault} {Diagnosis}},
	volume = {12},
	copyright = {https://creativecommons.org/licenses/by/4.0/legalcode},
	issn = {2169-3536},
	url = {https://ieeexplore.ieee.org/document/10752530/},
	doi = {10.1109/access.2024.3497716},
	language = {en},
	urldate = {2025-07-11},
	journal = {IEEE Access},
	author = {Wheat, Lesley and Mohrenschildt, Martin V. and Habibi, Saeid and Al-Ani, Dhafar},
	year = {2024},
	note = {Publisher: Institute of Electrical and Electronics Engineers (IEEE)},
	pages = {169879--169895},
}

@article{varejao_similarity_2025,
	title = {The similarity bias problem: {What} it is and how it impacts vibration based intelligent fault diagnosis},
	volume = {235},
	copyright = {https://www.elsevier.com/tdm/userlicense/1.0/},
	issn = {0888-3270},
	shorttitle = {The similarity bias problem},
	url = {https://linkinghub.elsevier.com/retrieve/pii/S0888327025005230},
	doi = {10.1016/j.ymssp.2025.112822},
	language = {en},
	urldate = {2025-07-11},
	journal = {Mechanical Systems and Signal Processing},
	author = {Varejão, Igor Mattos Dos Santos and Costa, Lucas Gabriel De Oliveira and Silva, Luciano Henrique Peixoto Da and Rodrigues, Alexandre and Ribeiro, Marcos Pellegrini and Varejão, Flávio Miguel and Oliveira-Santos, Thiago},
	month = jul,
	year = {2025},
	note = {Publisher: Elsevier BV},
	pages = {112822},
}

@article{ottawa,
	title = {University of {Ottawa} constant load and speed rolling-element bearing vibration and acoustic fault signature datasets},
	volume = {49},
	copyright = {https://www.elsevier.com/tdm/userlicense/1.0/},
	issn = {2352-3409},
	url = {https://linkinghub.elsevier.com/retrieve/pii/S2352340923004456},
	doi = {10.1016/j.dib.2023.109327},
	language = {en},
	urldate = {2025-07-11},
	journal = {Data in Brief},
	author = {Sehri, Mert and Dumond, Patrick and Bouchard, Michel},
	month = aug,
	year = {2023},
	note = {Publisher: Elsevier BV},
	pages = {109327},
}

@ARTICLE{alonso:2023,
  author={González, Miguel and Díaz, Vicente García and Pérez, Benjamìn López and G-Bustelo, B. Cristina Pelayo and Anzola, John Petearson},
  journal={IEEE Access}, 
  title={Bearing Fault Diagnosis With Envelope Analysis and Machine Learning Approaches Using CWRU Dataset}, 
  year={2023},
  volume={11},
  number={},
  pages={57796-57805},
  doi={10.1109/ACCESS.2023.3283466}}

@article{wdcnn,
	title = {A {New} {Deep} {Learning} {Model} for {Fault} {Diagnosis} with {Good} {Anti}-{Noise} and {Domain} {Adaptation} {Ability} on {Raw} {Vibration} {Signals}},
	volume = {17},
	copyright = {https://creativecommons.org/licenses/by/4.0/},
	issn = {1424-8220},
	url = {https://www.mdpi.com/1424-8220/17/2/425},
	doi = {10.3390/s17020425},
	language = {en},
	number = {2},
	urldate = {2025-07-11},
	journal = {Sensors},
	author = {Zhang, Wei and Peng, Gaoliang and Li, Chuanhao and Chen, Yuanhang and Zhang, Zhujun},
	month = feb,
	year = {2017},
	note = {Publisher: MDPI AG},
	pages = {425},
}

@article{cdcn,
	title = {Deep {Coupled} {Dense} {Convolutional} {Network} {With} {Complementary} {Data} for {Intelligent} {Fault} {Diagnosis}},
	volume = {66},
	copyright = {https://ieeexplore.ieee.org/Xplorehelp/downloads/license-information/IEEE.html},
	issn = {0278-0046, 1557-9948},
	url = {https://ieeexplore.ieee.org/document/8663605/},
	doi = {10.1109/tie.2019.2902817},
	language = {en},
	number = {12},
	urldate = {2025-07-11},
	journal = {IEEE Transactions on Industrial Electronics},
	author = {Jiao, Jinyang and Zhao, Ming and Lin, Jing and Ding, Chuancang},
	month = dec,
	year = {2019},
	note = {Publisher: Institute of Electrical and Electronics Engineers (IEEE)},
	pages = {9858--9867},
}

@inproceedings{wdtcnn,
	address = {Sorrento, Italy},
	title = {An {Improved} {Wide}-{Kernel} {CNN} for {Classifying} {Multivariate} {Signals} in {Fault} {Diagnosis}},
	copyright = {https://ieeexplore.ieee.org/Xplorehelp/downloads/license-information/IEEE.html},
	url = {https://ieeexplore.ieee.org/document/9346555/},
	doi = {10.1109/icdmw51313.2020.00046},
	language = {en},
	urldate = {2025-07-11},
	booktitle = {2020 {International} {Conference} on {Data} {Mining} {Workshops} ({ICDMW})},
	publisher = {IEEE},
	author = {Van Den Hoogen, J.O.D. and Bloemheuvel, S.D. and Atzmueller, M.},
	month = nov,
	year = {2020},
	pages = {275--283},
}

@inproceedings{dcase2018,
author = {Wei, Qingkai and Liu, Yanfang and Ruan, Xiaohui},
year = {2018},
month = {11},
pages = {},
title = {A report on audio tagging with deeper CNN, 1D-CONVNET and 2D-CONVNET},
publisher= {DCASE},
url={https://dcase.community/documents/challenge2018/technical_reports/DCASE2018_WEI_53.pdf}
}

@article{resnet1d,
	title = {Application of {1D} {ResNet} for {Multivariate} {Fault} {Detection} on {Semiconductor} {Manufacturing} {Equipment}},
	volume = {23},
	copyright = {https://creativecommons.org/licenses/by/4.0/},
	issn = {1424-8220},
	url = {https://www.mdpi.com/1424-8220/23/22/9099},
	doi = {10.3390/s23229099},
	language = {en},
	number = {22},
	urldate = {2025-07-11},
	journal = {Sensors},
	author = {Tchatchoua, Philip and Graton, Guillaume and Ouladsine, Mustapha and Christaud, Jean-François},
	month = nov,
	year = {2023},
	note = {Publisher: MDPI AG},
	pages = {9099},
}

@article{randomgain,
	title = {{SMoCo}: {A} {Powerful} and {Efficient} {Method} {Based} on {Self}-{Supervised} {Learning} for {Fault} {Diagnosis} of {Aero}-{Engine} {Bearing} under {Limited} {Data}},
	volume = {10},
	copyright = {https://creativecommons.org/licenses/by/4.0/},
	issn = {2227-7390},
	shorttitle = {{SMoCo}},
	url = {https://www.mdpi.com/2227-7390/10/15/2796},
	doi = {10.3390/math10152796},
	language = {en},
	number = {15},
	urldate = {2025-07-11},
	journal = {Mathematics},
	author = {Yan, Zitong and Liu, Hongmei},
	month = aug,
	year = {2022},
	note = {Publisher: MDPI AG},
	pages = {2796},
}

@inproceedings{datamining,
author = {Kaufman, Shachar and Rosset, Saharon and Perlich, Claudia},
year = {2011},
month = {01},
pages = {556-563},
title = {Leakage in Data Mining: Formulation, Detection, and Avoidance},
volume = {6},
journal = {Proceedings of the ACM SIGKDD International Conference on Knowledge Discovery and Data Mining},
doi = {10.1145/2020408.2020496}
}

@article{pitfalls,
  author       = {Michael A. Lones},
  title        = {How to avoid machine learning pitfalls: a guide for academic researchers},
  journal      = {CoRR},
  volume       = {abs/2108.02497},
  year         = {2021},
  url          = {https://arxiv.org/abs/2108.02497},
  eprinttype    = {arXiv},
  eprint       = {2108.02497},
  bibsource    = {dblp computer science bibliography, https://dblp.org}
}

@ARTICLE{wang2021,
  author={Wang, Daichao and Li, Yibin and Jia, Lei and Song, Yan and Liu, Yanjun},
  journal={IEEE Transactions on Instrumentation and Measurement}, 
  title={Novel Three-Stage Feature Fusion Method of Multimodal Data for Bearing Fault Diagnosis}, 
  year={2021},
  volume={70},
  number={},
  pages={1-10},
  doi={10.1109/TIM.2021.3071232}}

@ARTICLE{wang2023,
  author={Wang, Daichao and Li, Yibin and Jia, Lei and Song, Yan and Wen, Tao},
  journal={IEEE/ASME Transactions on Mechatronics}, 
  title={Attention-Based Bilinear Feature Fusion Method for Bearing Fault Diagnosis}, 
  year={2023},
  volume={28},
  number={3},
  pages={1695-1705},
  doi={10.1109/TMECH.2022.3223358}}

@article{Lones_2024, title={Avoiding common machine learning pitfalls}, ISSN={2666-3899}, DOI={10.1016/j.patter.2024.101046}, abstractNote={Mistakes in machine learning practice are commonplace and can result in loss of confidence in the findings and products of machine learning. This tutorial outlines common mistakes that occur when using machine learning and what can be done to avoid them. While it should be accessible to anyone with a basic understanding of machine learning techniques, it focuses on issues that are of particular concern within academic research, such as the need to make rigorous comparisons and reach valid conclusions. It covers five stages of the machine learning process: what to do before model building, how to reliably build models, how to robustly evaluate models, how to compare models fairly, and how to report results.}, journal={Patterns}, author={Lones, Michael A.}, year={2024}, month=aug, pages={101046} }

@article{Roberts_2017, title={Cross‐validation strategies for data with temporal, spatial, hierarchical, or phylogenetic structure}, volume={40}, ISSN={0906-7590, 1600-0587}, DOI={10.1111/ecog.02881}, abstractNote={Ecological data often show temporal, spatial, hierarchical (random effects), or phylogenetic structure. Modern statistical approaches are increasingly accounting for such dependencies. However, when performing cross‐validation, these structures are regularly ignored, resulting in serious underestimation of predictive error. One cause for the poor performance of uncorrected (random) cross‐validation, noted often by modellers, are dependence structures in the data that persist as dependence structures in model residuals, violating the assumption of independence. Even more concerning, because often overlooked, is that structured data also provides ample opportunity for overfitting with non‐causal predictors. This problem can persist even if remedies such as autoregressive models, generalized least squares, or mixed models are used. Block cross‐validation, where data are split strategically rather than randomly, can address these issues. However, the blocking strategy must be carefully considered. Blocking in space, time, random effects or phylogenetic distance, while accounting for dependencies in the data, may also unwittingly induce extrapolations by restricting the ranges or combinations of predictor variables available for model training, thus overestimating interpolation errors. On the other hand, deliberate blocking in predictor space may also improve error estimates when extrapolation is the modelling goal. Here, we review the ecological literature on non‐random and blocked cross‐validation approaches. We also provide a series of simulations and case studies, in which we show that, for all instances tested, block cross‐validation is nearly universally more appropriate than random cross‐validation if the goal is predicting to new data or predictor space, or for selecting causal predictors. We recommend that block cross‐validation be used wherever dependence structures exist in a dataset, even if no correlation structure is visible in the fitted model residuals, or if the fitted models account for such correlations.}, number={8}, journal={Ecography}, author={Roberts, David R. and Bahn, Volker and Ciuti, Simone and Boyce, Mark S. and Elith, Jane and Guillera‐Arroita, Gurutzeta and Hauenstein, Severin and Lahoz‐Monfort, José J. and Schröder, Boris and Thuiller, Wilfried and Warton, David I. and Wintle, Brendan A. and Hartig, Florian and Dormann, Carsten F.}, year={2017}, month=aug, pages={913–929}, language={en} }

@article{Kapoor2024, title={REFORMS: Consensus-based Recommendations for Machine-learning-based Science}, volume={10}, ISSN={2375-2548}, DOI={10.1126/sciadv.adk3452}, abstractNote={Machine learning (ML) methods are proliferating in scientific research. However, the adoption of these methods has been accompanied by failures of validity, reproducibility, and generalizability. These failures can hinder scientific progress, lead to false consensus around invalid claims, and undermine the credibility of ML-based science. ML methods are often applied and fail in similar ways across disciplines. Motivated by this observation, our goal is to provide clear recommendations for conducting and reporting ML-based science. Drawing from an extensive review of past literature, we present the REFORMS checklist (recommendations for machine-learning-based science). It consists of 32 questions and a paired set of guidelines. REFORMS was developed on the basis of a consensus of 19 researchers across computer science, data science, mathematics, social sciences, and biomedical sciences. REFORMS can serve as a resource for researchers when designing and implementing a study, for referees when reviewing papers, and for journals when enforcing standards for transparency and reproducibility.
          , 
            We provide a checklist to improve reporting practices in ML-based science based on a review of best practices and common errors.}, number={18}, journal={Science Advances}, author={Kapoor, Sayash and Cantrell, Emily M. and Peng, Kenny and Pham, Thanh Hien and Bail, Christopher A. and Gundersen, Odd Erik and Hofman, Jake M. and Hullman, Jessica and Lones, Michael A. and Malik, Momin M. and Nanayakkara, Priyanka and Poldrack, Russell A. and Raji, Inioluwa Deborah and Roberts, Michael and Salganik, Matthew J. and Serra-Garcia, Marta and Stewart, Brandon M. and Vandewiele, Gilles and Narayanan, Arvind}, year={2024}, month=may, pages={eadk3452}, language={en} }

@article{Rauber_2021, title={An experimental methodology to evaluate machine learning methods for fault diagnosis based on vibration signals}, volume={167}, ISSN={0957-4174}, DOI={10.1016/j.eswa.2020.114022}, abstractNote={This paper presents a systematic procedure to fairly compare experimental performance scores for machine learning methods for fault diagnosis based on vibration signals. In the vast majority of related scientific publications, the estimated accuracy and similar performance criteria are the sole quality parameter presented. However, the experimental design giving rise to these results is mostly biased, based on unacceptably simple validation methods and on recycling identical patterns in test data sets, previously used for training. Moreover, the methods in general overfit their hyperparameters, introducing additional overoptimistic results. In order to remedy this defect, we critically analyse the usual training-validation-test division and propose an algorithmic guideline in the form of a validation framework. This allows a well defined comparison of experimental results. In order to illustrate the ideas of the paper, the Case Western Reserve University Bearing Data benchmark is used as a case study. Four distinct classifiers are experimentally compared, under gradually more difficult generalization tasks using the proposed evaluation framework: K-Nearest-Neighbor, Support Vector Machine, Random Forest and One-Dimensional Convolutional Neural Network. An extensive literature review suggests that most vibration based research papers, particularly for the Case Western Reserve University Bearing Data, use similar patterns for training and testing, making their classification an easy task.}, journal={Expert Systems with Applications}, author={Rauber, Thomas Walter and da Silva Loca, Antonio Luiz and Boldt, Francisco de Assis and Rodrigues, Alexandre Loureiros and Varejão, Flávio Miguel}, year={2021}, pages={114022} }

@article{sklearn,
  title={Scikit-learn: Machine Learning in {P}ython},
  author={Pedregosa, F. and Varoquaux, G. and Gramfort, A. and Michel, V.
          and Thirion, B. and Grisel, O. and Blondel, M. and Prettenhofer, P.
          and Weiss, R. and Dubourg, V. and Vanderplas, J. and Passos, A. and
          Cournapeau, D. and Brucher, M. and Perrot, M. and Duchesnay, E.},
  journal={Journal of Machine Learning Research},
  volume={12},
  pages={2825--2830},
  year={2011}
}

@article{matania-leakage,
	title = {Test-{Training} {Leakage} in {Evaluation} of {Machine} {Learning} {Algorithms} for {Condition}-{Based} {Maintenance}},
	volume = {8},
	copyright = {http://creativecommons.org/licenses/by/3.0/us/},
	issn = {2325-016X, 2325-016X},
	url = {https://papers.phmsociety.org/index.php/phme/article/view/4125},
	doi = {10.36001/phme.2024.v8i1.4125},
	language = {en},
	number = {1},
	urldate = {2025-11-28},
	journal = {PHM Society European Conference},
	author = {Matania, Omri and Cohen, Roee and Bechhoefer, Eric and Bortman, Jacob},
	month = jun,
	year = {2024},
	pages = {13},
}

@article{leakage1,
	title = {Data-{Driven} {Fault} {Diagnosis} for {Rolling} {Bearings} {Based} on {Machine} {Learning} and {Multisensor} {Information} {Fusion}},
	volume = {25},
	copyright = {https://ieeexplore.ieee.org/Xplorehelp/downloads/license-information/IEEE.html},
	issn = {1530-437X, 1558-1748, 2379-9153},
	url = {https://ieeexplore.ieee.org/document/10768937/},
	doi = {10.1109/JSEN.2024.3499365},
	language = {en},
	number = {2},
	urldate = {2025-11-28},
	journal = {IEEE Sensors Journal},
	author = {Shuming, Yang and Changlin, Xie and Yuqiang, Cheng and Biao, Wang and Xunyi, Ma and Zinuo, Wang},
	month = jan,
	year = {2025},
	pages = {3452--3464},
}

@article{leakage2,
	title = {A novel weighted sparse classification framework with extended discriminative dictionary for data-driven bearing fault diagnosis},
	volume = {222},
	issn = {08883270},
	url = {https://linkinghub.elsevier.com/retrieve/pii/S0888327024006757},
	doi = {10.1016/j.ymssp.2024.111777},
	language = {en},
	urldate = {2025-11-28},
	journal = {Mechanical Systems and Signal Processing},
	author = {Cui, Lingli and Jiang, Zhichao and Liu, Dongdong and Zhen, Dong},
	month = jan,
	year = {2025},
	pages = {111777},
}

@article{leakage3,
	title = {Adaptive {Convergent} {Visibility} {Graph} {Network}: {An} interpretable method for intelligent rolling bearing diagnosis},
	volume = {222},
	issn = {08883270},
	shorttitle = {Adaptive {Convergent} {Visibility} {Graph} {Network}},
	url = {https://linkinghub.elsevier.com/retrieve/pii/S0888327024006599},
	doi = {10.1016/j.ymssp.2024.111761},
	language = {en},
	urldate = {2025-11-28},
	journal = {Mechanical Systems and Signal Processing},
	author = {Li, Xinming and Wang, Yanxue and Zhao, Shuangchen and Yao, Jiachi and Li, Meng},
	month = jan,
	year = {2025},
	pages = {111761},
}

@article{leakage4,
	title = {Lightweight pyramid attention residual network for intelligent fault diagnosis of machine under sharp speed variation},
	volume = {223},
	issn = {08883270},
	url = {https://linkinghub.elsevier.com/retrieve/pii/S0888327024007222},
	doi = {10.1016/j.ymssp.2024.111824},

	language = {en},
	urldate = {2025-11-28},
	journal = {Mechanical Systems and Signal Processing},
	author = {Xie, Zongliang and Chen, Jinglong and Shi, Zhen and Liu, Shen and He, Shuilong},
	month = jan,
	year = {2025},
	pages = {111824},
}

@article{leakage5,
	title = {Domain-invariant feature exploration for intelligent fault diagnosis under unseen and time-varying working conditions},
	volume = {224},
	issn = {08883270},
	url = {https://linkinghub.elsevier.com/retrieve/pii/S0888327024010926},
	doi = {10.1016/j.ymssp.2024.112193},
	language = {en},
	urldate = {2025-11-28},
	journal = {Mechanical Systems and Signal Processing},
	author = {Hua, Zehui and Shi, Juanjuan and Dumond, Patrick},
	month = feb,
	year = {2025},
	pages = {112193},
}

@article{leakage6,
	title = {Towards {Enhanced} {Interpretability}: {A} {Mechanism}-{Driven} domain adaptation model for bearing fault diagnosis across operating conditions},
	volume = {225},
	issn = {08883270},
	shorttitle = {Towards {Enhanced} {Interpretability}},
	url = {https://linkinghub.elsevier.com/retrieve/pii/S0888327024011439},
	doi = {10.1016/j.ymssp.2024.112244},
	language = {en},
	urldate = {2025-11-28},
	journal = {Mechanical Systems and Signal Processing},
	author = {Jiang, Fei and Kuang, Yicong and Li, Tao and Zhang, Shaohui and Wu, Zhaoqian and Feng, Ke and Li, Weihua},
	month = feb,
	year = {2025},
	pages = {112244},
}

@article{leakage7,
	title = {Fault diagnosis method based on multimodal-deep tensor projection network under variable working conditions},
	volume = {225},
	issn = {08883270},
	url = {https://linkinghub.elsevier.com/retrieve/pii/S0888327025000378},
	doi = {10.1016/j.ymssp.2025.112336},
	language = {en},
	urldate = {2025-11-28},
	journal = {Mechanical Systems and Signal Processing},
	author = {Li, Zhinong and Liu, Chenyu and Huang, Wenjing and Wang, Fengtao and Yang, Wenxian},
	month = feb,
	year = {2025},
	pages = {112336},
}

@article{leakage8,
	title = {Classifier-guided neural blind deconvolution: {A} physics-informed denoising module for bearing fault diagnosis under noisy conditions},
	volume = {222},
	issn = {08883270},
	shorttitle = {Classifier-guided neural blind deconvolution},
	url = {https://linkinghub.elsevier.com/retrieve/pii/S0888327024006484},
	doi = {10.1016/j.ymssp.2024.111750},
	language = {en},
	urldate = {2025-11-28},
	journal = {Mechanical Systems and Signal Processing},
	author = {Liao, Jing-Xiao and He, Chao and Li, Jipu and Sun, Jinwei and Zhang, Shiping and Zhang, Xiaoge},
	month = jan,
	year = {2025},
	pages = {111750},
}

@article{leakage9,
	title = {Optimizing {Bearing} {Fault} {Diagnosis} in {Rotating} {Electrical} {Machines} {Using} {Deep} {Learning} and {Frequency} {Domain} {Features}},
	volume = {15},
	issn = {2076-3417},
	url = {https://www.mdpi.com/2076-3417/15/6/3132},
	doi = {10.3390/app15063132},
	language = {en},
	number = {6},
	urldate = {2025-11-28},
	journal = {Applied Sciences},
	author = {Quiles-Cucarella, Eduardo and García-Bádenas, Alejandro and Agustí-Mercader, Ignacio and Escrivá-Escrivá, Guillermo},
	month = mar,
	year = {2025},
	pages = {3132},
}

@inproceedings{leakage10,
	address = {Nanchang, China},
	title = {Rolling bearing fault diagnosis model based on {CWT} algorithm and {CBAM}-{CNN}},
	isbn = {978-1-5106-8927-5 978-1-5106-8928-2},
	url = {https://www.spiedigitallibrary.org/conference-proceedings-of-spie/13560/3061739/Rolling-bearing-fault-diagnosis-model-based-on-CWT-algorithm-and/10.1117/12.3061739.full},
	doi = {10.1117/12.3061739},
	language = {en},
	urldate = {2025-11-28},
	booktitle = {International {Conference} on {Computer} {Graphics}, {Artificial} {Intelligence}, and {Data} {Processing} ({ICCAID} 2024)},
	publisher = {SPIE},
	author = {Xinrong, Li and Chao, Yun and Xin, Xu and Caiyun, Wang and Wei, Han and Xiong, Xiao},
	editor = {Xu, Xin and Mohd Zain, Azlan Bin},
	month = apr,
	year = {2025},
	pages = {152},
}

@article{leakage11,
	title = {Rolling {Bearing} {Fault} {Diagnosis} {Based} on {VMD}-{DWT} and {HADS}-{CNN}-{BiLSTM} {Hybrid} {Model}},
	volume = {13},
	issn = {2075-1702},
	url = {https://www.mdpi.com/2075-1702/13/5/423},
	doi = {10.3390/machines13050423},
	language = {en},
	number = {5},
	urldate = {2025-11-28},
	journal = {Machines},
	author = {Shao, Luchuan and Zhao, Bing and Kang, Xutao},
	month = may,
	year = {2025},
	pages = {423},
}

@inproceedings{leakage12,
	address = {Gwalior, India},
	title = {An {Improved} {Bearing} {Fault} {Investigation} {Scheme} {Using} {1D} {CNN} with {PCA} and {SVM}},
	copyright = {https://doi.org/10.15223/policy-029},
	isbn = {979-8-3315-2169-1},
	url = {https://ieeexplore.ieee.org/document/10985009/},
	doi = {10.1109/IATMSI64286.2025.10985009},
	language = {en},
	urldate = {2025-11-28},
	booktitle = {2025 {IEEE} {International} {Conference} on {Interdisciplinary} {Approaches} in {Technology} and {Management} for {Social} {Innovation} ({IATMSI})},
	publisher = {IEEE},
	author = {Kumar, Rajeev and Anand, R S and Akhtar, Muhammad Nameer and Khanna, Rintu},
	month = mar,
	year = {2025},
	pages = {1--5},
}

@article{leakage13,
	title = {Contrastive prototype guided federated learning for rotating machinery fault diagnosis under spatio-temporal domain shift},
	volume = {232},
	issn = {08883270},
	url = {https://linkinghub.elsevier.com/retrieve/pii/S088832702500408X},
	doi = {10.1016/j.ymssp.2025.112707},
	language = {en},
	urldate = {2025-11-28},
	journal = {Mechanical Systems and Signal Processing},
	author = {Fang, Luo and Shi, Jianhua and Qu, Hongyi and Safran, Mejdl and Tan, Jinbiao and Wan, Jiafu},
	month = jun,
	year = {2025},
	pages = {112707},
}

@article{leakage14,
	title = {Benchmarking {Machine} {Learning} {Algorithms} for {Bearing} {Fault} {Classification} {Using} {Vibration} {Data}: {A} {Deployment}-{Oriented} {Study}},
	volume = {13},
	copyright = {https://creativecommons.org/licenses/by/4.0/legalcode},
	issn = {2169-3536},
	shorttitle = {Benchmarking {Machine} {Learning} {Algorithms} for {Bearing} {Fault} {Classification} {Using} {Vibration} {Data}},
	url = {https://ieeexplore.ieee.org/document/11045386/},
	doi = {10.1109/ACCESS.2025.3581711},
	language = {en},
	urldate = {2025-11-28},
	journal = {IEEE Access},
	author = {Samal, Prasanta Kumar and Srinidhi, R. and Malik, Pramod Kumar and Manjunatha, H. J. and Jamadar, Imran M.},
	year = {2025},
	pages = {113984--114002},
}

@article{leakage15,
	title = {A {Rolling}-{Bearing}-{Fault} {Diagnosis} {Method} {Based} on a {Dual} {Multi}-{Scale} {Mechanism} {Applicable} to {Noisy}-{Variable} {Operating} {Conditions}},
	volume = {25},
	issn = {1424-8220},
	url = {https://www.mdpi.com/1424-8220/25/15/4649},
	doi = {10.3390/s25154649},
	language = {en},
	number = {15},
	urldate = {2025-11-28},
	journal = {Sensors},
	author = {Kang, Jing and Wang, Taiyong and Wei, Ye and Garba, Usman Haladu and Tian, Ying},
	month = jul,
	year = {2025},
	pages = {4649},
}

@article{leakage16,
	title = {{LSFConvformer}: {A} lightweight method for mechanical fault diagnosis under small samples and variable speeds with time-frequency fusion},
	volume = {236},
	issn = {08883270},
	shorttitle = {{LSFConvformer}},
	url = {https://linkinghub.elsevier.com/retrieve/pii/S0888327025007174},
	doi = {10.1016/j.ymssp.2025.113016},
	language = {en},
	urldate = {2025-11-28},
	journal = {Mechanical Systems and Signal Processing},
	author = {Shao, Haidong and Lai, Yanzuo and Liu, Haoran and Wang, Jie and Liu, Bin},
	month = aug,
	year = {2025},
	pages = {113016},
}

@article{leakage17,
	title = {A novel bearing faults diagnosis of rotor-bearing systems based on vibration responses and convolutional neural network},
	volume = {236},
	issn = {08883270},
	url = {https://linkinghub.elsevier.com/retrieve/pii/S0888327025007563},
	doi = {10.1016/j.ymssp.2025.113055},
	language = {en},
	urldate = {2025-11-28},
	journal = {Mechanical Systems and Signal Processing},
	author = {Patil, S.S. and Salunkhe, Vishal G. and Jadhav, Prashant S. and Khot, S.M. and Desavale, Shravani R. and Desavale, R.G.},
	month = aug,
	year = {2025},
	pages = {113055},
}

@article{leakage18,
	title = {Anovel fault diagnosis method for rolling bearing based on {SGMD} and improved {CNN}-{LSTM}},
	volume = {7},
	issn = {2631-8695},
	url = {https://iopscience.iop.org/article/10.1088/2631-8695/adf93b},
	doi = {10.1088/2631-8695/adf93b},
	language = {en},
	number = {3},
	urldate = {2025-11-28},
	journal = {Engineering Research Express},
	author = {Xu, Guohui and Cao, Jianbin and Liu, Wenyi and Song, Di and Zhong, Jiahao and Meng, Lei},
	month = sep,
	year = {2025},
	pages = {035567},
}

@article{zheng_deep_2021,
	title = {Deep {Domain} {Generalization} {Combining} {A} {Priori} {Diagnosis} {Knowledge} {Toward} {Cross}-{Domain} {Fault} {Diagnosis} of {Rolling} {Bearing}},
	volume = {70},
	copyright = {https://ieeexplore.ieee.org/Xplorehelp/downloads/license-information/IEEE.html},
	issn = {0018-9456, 1557-9662},
	url = {https://ieeexplore.ieee.org/document/9174912/},
	doi = {10.1109/TIM.2020.3016068},
	language = {en},
	urldate = {2025-12-02},
	journal = {IEEE Transactions on Instrumentation and Measurement},
	author = {Zheng, Huailiang and Yang, Yuantao and Yin, Jiancheng and Li, Yuqing and Wang, Rixin and Xu, Minqiang},
	year = {2021},
	pages = {1--11},
}

@article{matania_zero-fault-shot_2025,
	title = {Zero-fault-shot learning for bearing spall type classification by hybrid approach},
	volume = {224},
	issn = {08883270},
	url = {https://linkinghub.elsevier.com/retrieve/pii/S088832702401015X},
	doi = {10.1016/j.ymssp.2024.112117},
	language = {en},
	urldate = {2025-12-02},
	journal = {Mechanical Systems and Signal Processing},
	author = {Matania, Omri and Cohen, Roee and Bechhoefer, Eric and Bortman, Jacob},
	month = feb,
	year = {2025},
	pages = {112117},
}

@article{chen_deep_2023,
	title = {Deep {Transfer} {Learning} for {Bearing} {Fault} {Diagnosis}: {A} {Systematic} {Review} {Since} 2016},
	volume = {72},
	copyright = {https://ieeexplore.ieee.org/Xplorehelp/downloads/license-information/IEEE.html},
	issn = {0018-9456, 1557-9662},
	shorttitle = {Deep {Transfer} {Learning} for {Bearing} {Fault} {Diagnosis}},
	url = {https://ieeexplore.ieee.org/document/10042467/},
	doi = {10.1109/TIM.2023.3244237},
	language = {en},
	urldate = {2025-12-04},
	journal = {IEEE Transactions on Instrumentation and Measurement},
	author = {Chen, Xiaohan and Yang, Rui and Xue, Yihao and Huang, Mengjie and Ferrero, Roberto and Wang, Zidong},
	year = {2023},
	pages = {1--21},
}

@article{fawcett2006,
author = {Fawcett, Tom},
year = {2006},
month = {06},
pages = {861-874},
title = {Introduction to ROC analysis},
volume = {27},
journal = {Pattern Recognition Letters},
doi = {10.1016/j.patrec.2005.10.010}
}

@article{thuan_hust_2023,
	title = {{HUST} bearing: a practical dataset for ball bearing fault diagnosis},
	volume = {16},
	issn = {1756-0500},
	shorttitle = {{HUST} bearing},
	url = {https://bmcresnotes.biomedcentral.com/articles/10.1186/s13104-023-06400-4},
	doi = {10.1186/s13104-023-06400-4},
	language = {en},
	number = {1},
	urldate = {2026-04-22},
	journal = {BMC Research Notes},
	author = {Thuan, Nguyen Duc and Hong, Hoang Si},
	month = jul,
	year = {2023},
	pages = {138},
}

@article{soomro_insights_2024,
	title = {Insights into modern machine learning approaches for bearing fault classification: {A} systematic literature review},
	volume = {23},
	issn = {25901230},
	shorttitle = {Insights into modern machine learning approaches for bearing fault classification},
	url = {https://linkinghub.elsevier.com/retrieve/pii/S2590123024009551},
	doi = {10.1016/j.rineng.2024.102700},
	language = {en},
	urldate = {2026-05-01},
	journal = {Results in Engineering},
	author = {Soomro, Afzal Ahmed and Muhammad, Masdi B. and Mokhtar, Ainul Akmar and Md Saad, Mohamad Hanif and Lashari, Najeebullah and Hussain, Muhammad and Sarwar, Umair and Palli, Abdul Sattar},
	month = sep,
	year = {2024},
	pages = {102700},
}

@article{hoang_survey_2019,
	title = {A survey on {Deep} {Learning} based bearing fault diagnosis},
	volume = {335},
	issn = {09252312},
	url = {https://linkinghub.elsevier.com/retrieve/pii/S0925231218312657},
	doi = {10.1016/j.neucom.2018.06.078},
	language = {en},
	urldate = {2026-05-01},
	journal = {Neurocomputing},
	author = {Hoang, Duy-Tang and Kang, Hee-Jun},
	month = mar,
	year = {2019},
	pages = {327--335},
}

@article{barai_bearing_2022,
	title = {Bearing fault diagnosis using signal processing and machine learning techniques: {A} review},
	volume = {1259},
	issn = {1757-8981, 1757-899X},
	shorttitle = {Bearing fault diagnosis using signal processing and machine learning techniques},
	url = {https://iopscience.iop.org/article/10.1088/1757-899X/1259/1/012034},
	doi = {10.1088/1757-899X/1259/1/012034},
	language = {en},
	number = {1},
	urldate = {2026-05-01},
	journal = {IOP Conference Series: Materials Science and Engineering},
	author = {Barai, Viplav and Ramteke, Sangharatna M. and Dhanalkotwar, Vismay and Nagmote, Yatharth and Shende, Suyash and Deshmukh, Dheeraj},
	month = oct,
	year = {2022},
	pages = {012034},

}

\end{document}